\documentclass[10pt,twocolumn,letterpaper]{article}

\usepackage{calc}
\usepackage{graphicx}
\usepackage{csquotes}
\usepackage[dvipsnames]{xcolor}
\usepackage{iccv}
\usepackage{times}
\usepackage{epsfig}
\usepackage{etoolbox}
\usepackage{mathtools}
\usepackage{amssymb}

\makeatletter
\@namedef{ver@everyshi.sty}{}
\makeatother
\usepackage{tikz}

\usepackage{array,colortbl,multirow,multicol,booktabs,ctable}
\usepackage{bm}
\usepackage{subfig}
\usepackage{grffile} 
\usepackage{dblfloatfix}
\usepackage{placeins}
\usepackage{letltxmacro}
\usepackage{xifthen}
\usepackage{xparse}
\usepackage[pagebackref=true,breaklinks=true,letterpaper=true,colorlinks,bookmarks=false]{hyperref}

\iccvfinalcopy

\renewcommand{\paragraph}[1]{\noindent{\bf #1}}

\begin{document}
    \title{Perceptual Deep Depth Super-Resolution}
\author{\
Oleg Voynov\textsuperscript{1},
Alexey Artemov\textsuperscript{1},
Vage Egiazarian\textsuperscript{1},
Alexander Notchenko\textsuperscript{1},
\\\
Gleb Bobrovskikh\textsuperscript{1,2},
Denis Zorin\textsuperscript{3,1},
Evgeny Burnaev\textsuperscript{1}
\\\
\textsuperscript{1}Skolkovo Institute of Science and Technology,
\textsuperscript{2}Higher School of Economics,
\\\
\textsuperscript{3}New York University
\\\
{\tt\small \{oleg.voinov, a.artemov, vage.egiazarian, alexandr.notchenko\}@skoltech.ru,}
\\
{\tt\small bobrovskihg@gmail.com, dzorin@cs.nyu.edu, e.burnaev@skoltech.ru}
}
\def\projecturl{\href{http://adase.group/3ddl/projects/perceptual-depth-sr}{adase.group/3ddl/projects/perceptual-depth-sr}}

    %%% Graphics
\graphicspath{{img/}}
\DeclareGraphicsExtensions{.eps,.pdf,.png,.jpg}

%%% Formatting
\newcommand{\ra}[1]{\renewcommand{\arraystretch}{#1}}
\newcommand{\cm}[0]{\checkmark}

\ra{1.04}
\setlength{\tabcolsep}{5.5pt}
\def\vs.{vs.\spacefactor=\the\sfcode`\v}
\def\etc.{etc.\spacefactor=\the\sfcode`\c}

%%% Notations and aliases
\newcommand{\argmin}{\mathop{{\rm arg\,min}}}
\def\cnn{\mathrm{CNN}}
\def\depthmap{d}
\def\downsamplingoperator{\bm{\mathrm{D}}}
\def\featuremap{\bm{x}}
\def\lightdir{\bm{e}}
\def\loss#1{\mathcal{L}(#1)}
\def\metric_#1#2{\mathrm{#1}(#2)}
\def\normal{\bm{n}}
\def\ourmetric{\mathrm{RMSE_v}}
\def\ourmetricinloss{\mathrm{MSE_v}}
\def\rendering{I}
\def\rmseonimage{\mathrm{RMSE_v^1}}

    \maketitle

\begin{abstract}
	RGBD images, combining high-resolution color and lower-resolution depth from various types of depth sensors, are increasingly common.
	One can significantly improve the resolution of depth maps by taking advantage of color information;
	deep learning methods make combining color and depth information particularly easy.

	However, fusing these two sources of data may lead to a variety of artifacts.
	 If depth maps are used to reconstruct 3D shapes, \eg, for virtual reality applications, the visual quality of upsampled images is particularly important.

	The main idea of our approach is to measure the quality of depth map upsampling using renderings of resulting 3D surfaces.
	 We demonstrate that a simple visual appearance-based loss, when used with either a trained CNN or simply a deep prior, yields significantly improved 3D shapes, as measured by a number of existing perceptual metrics.
	 We compare this approach with a number of existing optimization and learning-based techniques.
\end{abstract}

\section{Introduction}
\label{sec:intro}

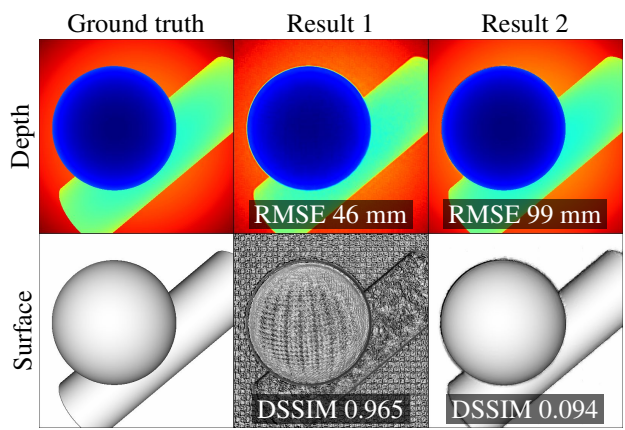
\begin{figure}[t]
	\centerline{\begin{tikzpicture}
        \node[inner sep=0] (image) {\includegraphics[width=.93\linewidth]{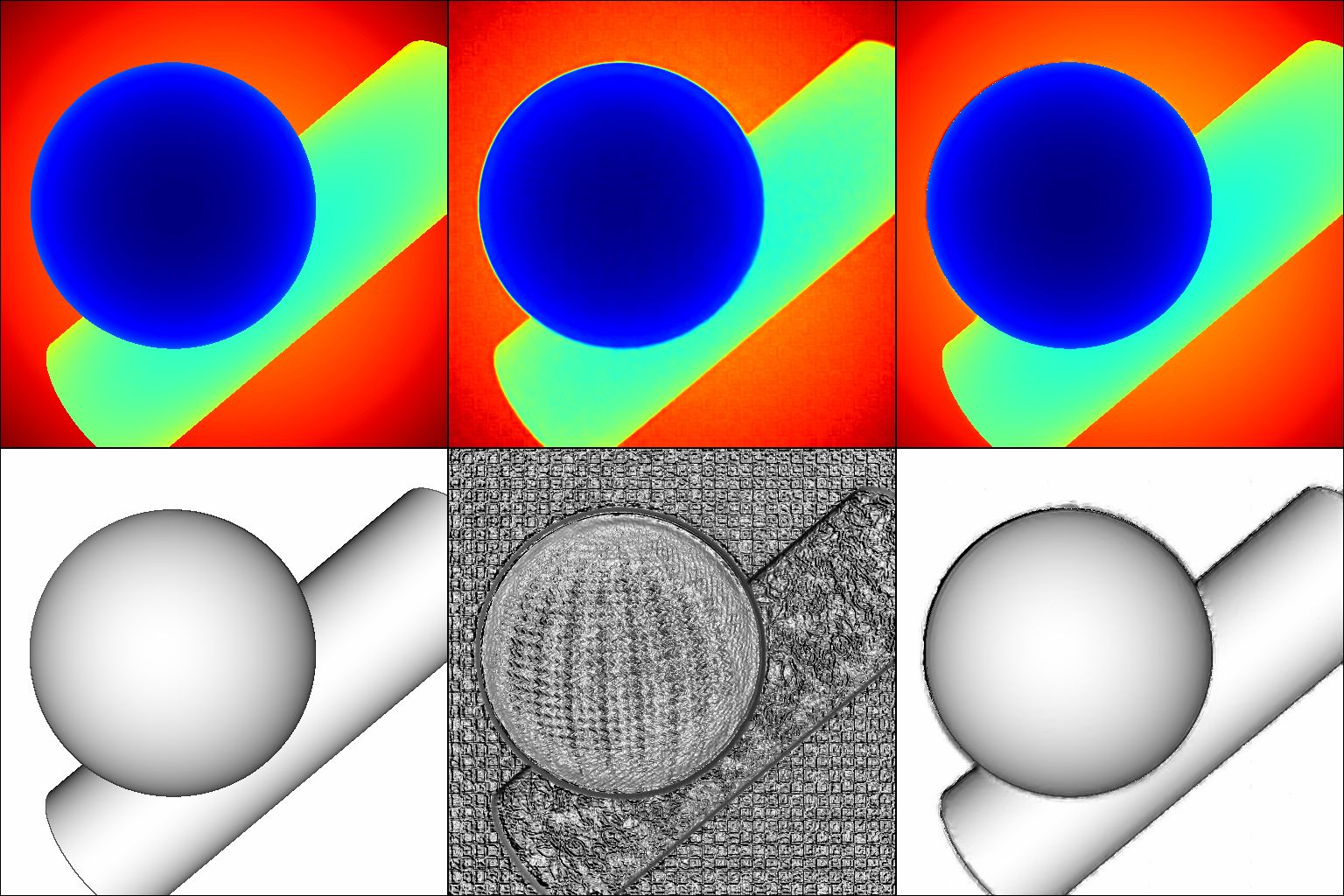}};
        \begin{scope}[shift=(image.north west), x={(image.north east)},y={(image.south west)}]
            \node at (.167, -.04) {Ground truth};
            \node at (.5, -.04) {Result 1};
            \node at (.833, -.04) {Result 2};
            \node[rotate=90] at (-.027, .25) {Depth};
            \node[rotate=90] at (-.027, .75) {Surface};
            \tikzstyle{label}=[fill=black, text=white, inner sep=2pt, fill opacity=0.6, text opacity=1]
            \node at (.5, 0.45) [label] {RMSE 46 mm};
            \node at (.833, 0.45) [label] {RMSE 99 mm};
            \node at (.5, 0.95) [label] {DSSIM 0.965};
            \node at (.833, 0.95) [label] {DSSIM 0.094};
        \end{scope}
	\end{tikzpicture}}
	\caption{Visually inferior super-resolution result in the middle gets higher score according to direct depth deviation but lower score according to perceptual deviation of the rendered image of the 3D surface.
	While the surfaces differ significantly, the corresponding depth maps do not capture this difference and look almost identical.}
	\label{fig:rmse_is_wrong}
\end{figure}

RGBD images are increasingly common as sensor technology becomes more widely available and affordable.
They can be used for reconstruction of the 3D shapes of objects and their surface appearance.
The better the quality of the depth component, the more reliable the reconstruction.

Unfortunately, for most methods of depth acquisition the resolution and quality of the depth component is insufficient for accurate surface reconstruction.
As the resolution of the RGB component is usually several times higher
and there is a high correlation between structural features of the color image and the depth map (\eg, object edges)
it is natural to use the color image for depth map super-resolution, \ie upsampling of the depth map.
Convolutional neural networks are a natural fit for this problem as they can easily fuse heterogeneous information.

A critical aspect of any upsampling method is the measure of quality it optimizes (\ie, the loss function), whether the technique is data-driven or not.
In this paper we focus on applications that require reconstruction of 3D geometry visible to the user,
like acquisition of realistic 3D scenes for virtual or augmented reality and computer graphics.
In these applications the \emph{visual} appearance of the resulting 3D shape,
\ie, how the surface looks when observed under various lighting conditions,
is of particular importance.

Most existing research on depth super-resolution is dominated by simple measures based on pointwise deviation of depth values.
However, direct pointwise difference of the depth maps do not capture the visual difference between the corresponding 3D shapes: for example, low-amplitude high-frequency variations of depth may correspond to significant difference in appearance,
while conversely, relatively large smooth changes in depth may be perceptually less relevant, as illustrated in~Figure~\ref{fig:rmse_is_wrong}.

Hence, we propose to compare the rendered images of the surface instead of the depth values directly.
In this paper we explore depth map super-resolution using a simple loss function based on visual differences.
Our loss function can be computed efficiently and is shown to be highly correlated with more elaborate perceptual metrics.
We demonstrate that this simple idea used with two deep learning-based RGBD super-resolution algorithms
results in a dramatic improvement of visual quality according to perceptual metrics and an informal perceptual study.
We compare our results with six state-of-the-art methods of depth super-resolution that are based on distinct principles and use several types of loss functions.

In summary, our contributions are as follows:
(1) we demonstrate that a simple and efficient visual difference-based metric for depth map comparison can be, on the one hand, easily combined with neural network-based whole-image upsampling techniques, and, on the other hand, is correlated with established proxies for human perception, validated with respect to experimental measurements;
(2) we demonstrate with extensive comparisons that with the use of this metric two methods of depth map super-resolution, one based on a trainable CNN and the other based on the deep prior, yield high-quality results as measured by multiple perceptual metrics.
To the best of our knowledge, our paper is the first to systematically study the performance of visual difference-based depth super-resolution across a variety of datasets, methods, and quality measures, including a basic human evaluation.

Throughout the paper we use the term \emph{depth map} to refer to the depth component of an RGBD image, and
the term \emph{normal map} to refer to the map of the same resolution with the 3D surface normal direction computed from the depth map at each pixel.
Finally, the \emph{rendering of a depth map} refers to the grayscale image obtained by constructing a 3D triangulation of the height field represented by the depth map, via computing the normal map from this triangulation, and rendering it using fixed material properties and a choice of lighting.
This is distinct from a commonly used depth map visualization with grayscale values obtained from the depth values by simple scaling.
We describe this in more detail in Section~\ref{sec:metrics}.

\section{Related work}
\label{sec:related}

\subsection{Image quality measures}
Quality measures play two important roles in image super-resolution: on the one hand, they are used to formulate an optimization functional or a loss function, on the other hand, they are used to evaluate the quality of the results.
Ideally, the same function should serve both purposes, however, in some instances it may be optimal to choose different functions for evaluation and optimization.
While in the former case the top priority is to capture the needs of the application,
in the latter case the efficiency of evaluation and differentiability are significant considerations.

In most works on depth map reconstruction and upsampling a limited number of simple metrics are used, both for optimization and final evaluation.
Typically these are scaled $L_2$ or $L_1$ norms of depth deviations (see \eg~\cite{eigen2014depth}).

Another set of measures introduced in~\cite{honauer2016dataset,honauer2015hci} and primarily used for evaluation, not optimization or learning,
consists of heuristic measures of various aspects of the depth map geometry: foreground flattening/thinning, fuzziness, bumpiness, etc.
Most of them require a very specific segmentation of the image for detection of flat areas and depth discontinuities.

Visual similarity measures, well-established in the area of photo-processing, aim to be consistent with human judgment, in the sense of similarity ordering (which of the two images is more similar to the ground truth?).
The examples include (1) the metrics based on simple vision models of \emph{structural similarity} SSIM~\cite{wang2004image}, FSIM~\cite{zhang2011fsim}, MSSIM~\cite{wang2003multiscale},
(2) based on a sophisticated model of low-level visual processing~\cite{mantiuk2011hdr},
or (3) on convolutional neural networks (see~\cite{Zhang_2018_CVPR} for a detailed overview).
The latter use a simple distance measure on deep features learned for an image understanding task,
\eg $L_2$ distance on the features learned for image classification,
and have been demonstrated to outperform statistical measures such as SSIM.

\subsection{Depth super-resolution}
\label{sec:depth-related}
Depth super-resolution is closely related to a number of depth processing tasks,
such as denoising, enhancement, inpainting, and densification (\eg,~\cite{chen2018estimating, cheng2018depth, chodosh2018deep, hua2018normalized, ma2018self, mal2018sparse, tsuchiya2017depth, uhrig2017sparsity, yan2018ddrnet}).
We directly focus on the problem of super-resolution, or more specifically, estimation of high-resolution depth map from a single low-resolution depth map and a high-resolution RGB image.

\paragraph{Convolutional neural networks} have achieved most impressive performance among learning-based methods in high-level computer vision tasks and recently have been applied to depth super-resolution~\cite{hui2016depth,li2016deep,riegler2016deep,song2016deep}.
One approach~\cite{hui2016depth} is to resolve ambiguity in the depth map upsampling by explicitly adding high-frequency features from high-resolution RGB data.
Another, hybrid approach~\cite{riegler2016deep, song2016deep} is to add a subsequent optimization stage to a CNN to produce sharper results.
Different approaches to CNN-based photo-guided depth super-resolution include linear filtering with CNN-derived kernels~\cite{kim2018deformable},
deep fusion of time-of-flight depth and stereo images~\cite{agresti2017deep},
and generative adversarial networks~\cite{zhao2017simultaneously}.

These techniques use either $L_2$ or $L_1$ norm of the depth differences as the basis of their loss functions, often combined with regularizers of different types.
The recent approach of~\cite{zhao2017simultaneously} is the closest to ours: it uses the difference of gradients as one of the loss terms to capture some of the visual information.
For evaluation, these works report root mean square error (RMSE), mean absolute error (MAE), peak signal-to-noise ratio (PSNR), all applied directly to depth maps,
and, rarely \cite{chen2018single, song2016deep, song2018deeply, zhao2017simultaneously}, perceptual SSIM \emph{also applied directly to depth maps}.
In contrast, we propose to measure the perceptual quality of depth map \emph{renderings}.

\paragraph{Dictionary learning} has also been investigated for depth super-resolution~\cite{ferstl2015variational,gu2017learning,li2018depth},
however, compared to CNNs, it is typically restricted to smaller dimensions and as a result to structurally simpler depth maps.

\paragraph{Variational approach} aims to combine RGB and depth information explicitly by carefully designing an optimization functional, without relying on learning.
Most relevant examples employ shape-from-shading problem statement for single-image~\cite{haefner2018fight} or multiple-image~\cite{peng2017depth} depth super-resolution.
These works include visual difference-related terms in the optimized functional and report normal deviation, capturing visual similarity.
While showing impressive results in many cases, they typically require prior segmentation of foreground objects and depend heavily on the quality of such segmentation.

Another strategy to tackle ambiguities in super-resolution is to design sophisticated regularizers to balance the data-fidelity terms against a structural image prior~\cite{ham2018robust,jiang2018depth,yang2014color}.
In contrast to this approach, which requires custom hand-crafted regularized objectives and optimization procedures, we focus on the standard training strategy (\ie, gradient-based optimization of a CNN) while using a loss function that captures visual similarity.

Yet another approach is to choose a carefully-designed model such as~\cite{zuo2018minimum}
featuring a sophisticated metric defined in a space of minimum spanning trees and including an explicit edge inconsistency model.
In contrast to ours, such model requires manual tuning of multiple hyperparameters.

\subsection{Perceptual photo super-resolution}
Perceptual metrics have been considered more broadly in the context of photo processing.
While convolutional neural networks for photo super-resolution trained with simple mean square or mean absolute color deviation keep demonstrating impressive results~\cite{han2018image,haris2018deep,zhang2018image,zhang2018residual},
it has been widely recognized that pixelwise difference of color image data is not well correlated with perceptual image difference.
For this reason, relying on a pixelwise color error may lead to suboptimal performance.

One solution is to instead use the loss function represented by the deviation of the features from a neural network trained for an image understanding task~\cite{johnson2016perceptual}.
This idea can be further combined with an adversarial training procedure to push the super-resolution result to the natural image manifold~\cite{ledig2016photo}.
Another extension to this idea is to train the neural network to generate images with natural distribution of statistical features~\cite{gondal2018unreasonable,mechrez2018learning,wang2018recovering,wang2018fully}.
To balance between the perceptual quality and pixelwise color deviation, generative adversarial networks can be used~\cite{cheon2019generative,luo2019bigan,vu2019perception}.

Another solution is to learn a quality measure from perceptual scores, collected from a human subject study, and use this quality measure as the loss function.
Such quality measure may capture similarity of two images~\cite{Zhang_2018_CVPR} or an absolute naturalness of the image~\cite{ma2017learning}.

\section{Metrics}
\label{sec:metrics}

\begin{figure}[t]
	\centerline{\includegraphics[width=\linewidth]{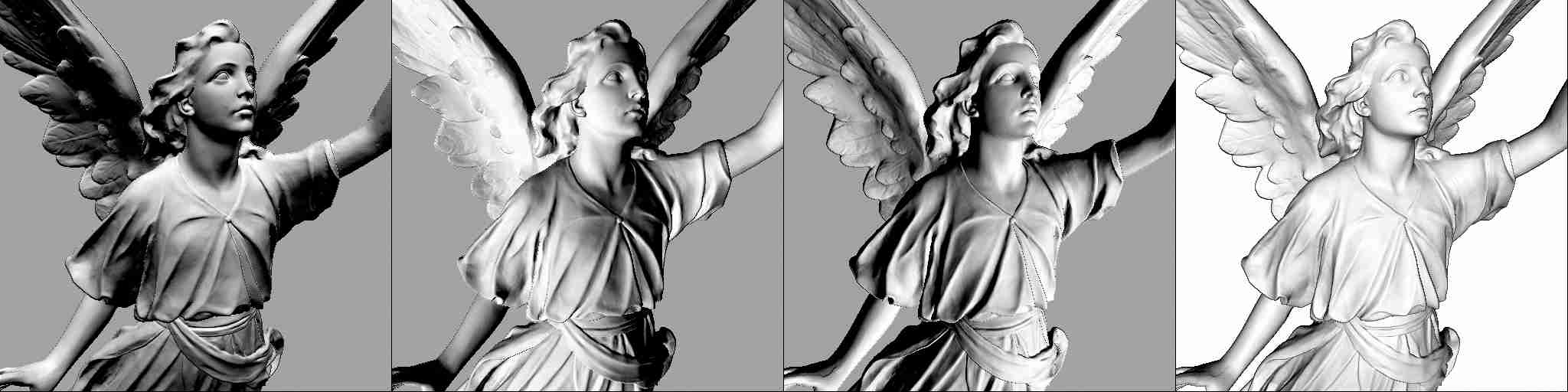}}
	\caption{Depth map renderings generated with four light directions that we use for metric calculation.}
	\label{fig:light_directions}
\end{figure}

In this section, we discuss visually-based metrics and how they can be used to evaluate the quality of depth map super-resolution and as loss functions.
The general principle we follow is to apply comparison metrics to \emph{renderings} of the depth maps to obtain a measure of their difference instead of considering depth maps directly.
The difficulty with this approach is that there are infinitely many
possible renderings depending on lighting conditions, material properties and camera position.
However, we demonstrate that even a very simple rendering procedure already yields substantially improved results.
We label visually-based metrics with subscript \textquote{v} and the metrics that compare the depth values directly with subscript \textquote{d}.

\paragraph{From depth map to visual representation.}
To approximate the appearance of a 3D scene depicted with a certain depth map we use a simple rendering procedure.
We illuminate the corresponding 3D surface with monochromatic directional light source and observe it with the same camera that the scene was originally acquired with.
We use the diffuse reflection model and do not take visibility into account.
For this model, the intensity of a pixel $(i, j)$ of the rendering  $\rendering$ is proportional to cosine of the angle between the normal at the point of the surface corresponding to the pixel $\normal_{ij}$ and direction to the light source $\lightdir$: $\rendering_{ij}=\lightdir\cdot\normal_{ij}$.
We calculate the normals from the depth maps using first-order finite-differences.
Any number of vectors $\lightdir$ can be used to generate a collection of renderings representing the depth map,
however, any rendering can be obtained as a linear combination of three basis ones corresponding to independent light directions.
Renderings for different light directions are presented in~Figure~\ref{fig:light_directions}.

\paragraph{Perceptual metrics.}
We briefly describe two representative metrics: a statistics-based DSSIM, and a neural network-based LPIPS. Either of these can be applied to three basis renderings (or a larger sample of renderings) and reduced to obtain the final value.
While, in principle, they can also be used as loss functions, the choice of a loss function needs to take stability and efficiency into account, so we opt for a more conservative choice described below.

\noindent\emph{Structural similarity index measure (SSIM)}~\cite{wang2004image} takes into account the changes in the local structure of an image, captured by statistical quantities computed on a small window around each pixel.
For each pair of pixels of the compared images $\rendering_k$, $k = 1,2$ the luminance term $\ell$, the contrast term $c$ and the structural term $s$, each normalized, are computed using the means $\mu_k$, standard deviations $\sigma_k$ and cross-covariance $\sigma_{12}$ of the pixels in the corresponding local windows.
The value of SSIM is then computed as pixelwise mean product of these terms
\begingroup
\setlength\abovedisplayskip{0.5em}
\setlength\belowdisplayskip{0.5em}
\begin{equation}
    \begin{gathered}
        \ell = \frac{2\mu_1 \mu_2}{\mu_1^2 + \mu_2^2},\;
        c = \frac{2\sigma_1 \sigma_2}{\sigma_1^2 + \sigma_2^2},\;
        s = \frac{\sigma_{12}}{\sigma_1\sigma_2}, \\
    \metric_{SSIM_v}{\rendering_1, \rendering_2} = \frac{1}{N}\sum_{ij}\ell_{ij}\cdot c_{ij}\cdot s_{ij},
    \end{gathered}
\end{equation}
\endgroup
where $N$ is the number of pixels.
Dissimilarity measure can be computed as $\metric_{DSSIM_v}{\rendering_1,\rendering_2} = 1-\metric_{SSIM_v}{\rendering_1,\rendering_2}$.

\noindent\emph{Neural net-based metrics} rely on the idea of measuring the distance between features extracted from a neural network.
Specifically, feature maps $\featuremap_{k\ell}$, $\ell = 1\ldots L$
with spatial dimensions $H_\ell \times W_\ell$
are extracted from $L$ layers of the network for each of the compared images.
In the simplest case, the metric value is then computed as pixelwise mean square difference of the feature maps, summed over the layers
\begingroup
\setlength\abovedisplayskip{0.5em}
\setlength\belowdisplayskip{0.5em}
\begin{equation}
	\metric_{NN_v}{\rendering_1, \rendering_2} = \sum_\ell \frac{1}{{H_\ell W_\ell}} \sum_{ij} \| \featuremap_{1\ell,ij} - \featuremap_{2\ell,ij}\|^2_2.
\end{equation}
\endgroup
\emph{Learned perceptual image patch similarity (LPIPS)}~\cite{Zhang_2018_CVPR} adds a learned channel-wise weighting to the above formula and uses 5 layers from Alexnet~\cite{krizhevsky2012imagenet} or VGG~\cite{Simonyan14c} or the first layer from Squezenet~\cite{i2016squeezenet} as the CNN of choice.

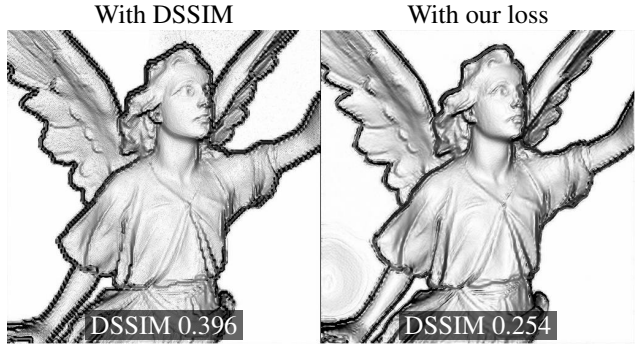
\begin{figure}
    \centerline{\begin{tikzpicture}
        \node[inner sep=0] (image) {\includegraphics[width=\linewidth]{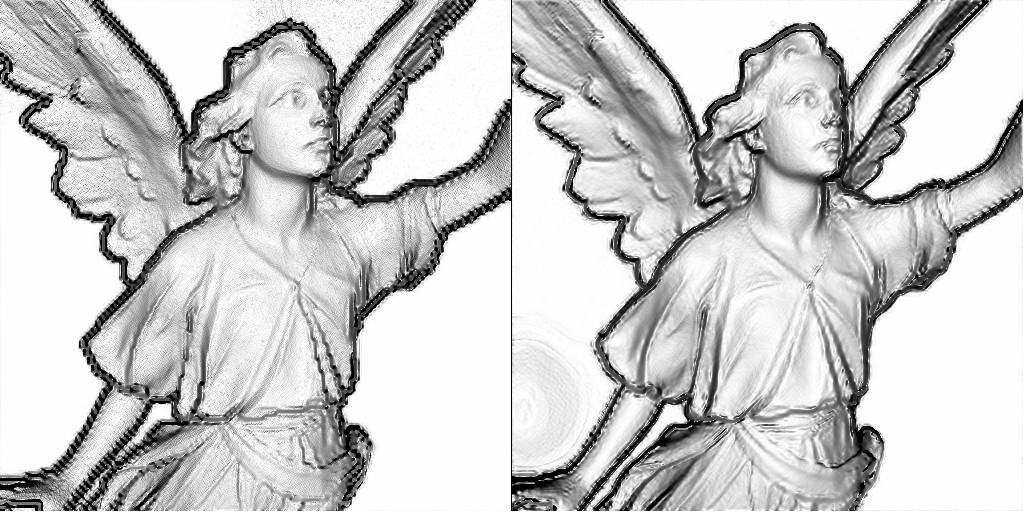}};
        \begin{scope}[shift=(image.north west), x={(image.north east)},y={(image.south west)}]
            \node at (0.25,-0.05) {With DSSIM};
            \node at (0.75,-0.05) {With our loss};
            \tikzstyle{label}=[fill=black, text=white, inner sep=2pt, fill opacity=0.6, text opacity=1]
            \node at (0.25,0.94) [label] {DSSIM 0.396};
            \node at (0.75,0.94) [label] {DSSIM 0.254};
        \end{scope}
    \end{tikzpicture}}
    \caption{Optimization with DSSIM as the loss function may produce the results inferior with respect to \emph{DSSIM itself} compared to our simpler loss function.}
    \label{fig:dssim-vs-ours}
\end{figure}

\paragraph{Our visual difference-based metric.}
While the metrics described above are good proxies for human evaluation of difference between depth map renderings, they are lacking as loss functions due to their complex landscapes.
Optimization with DSSIM as the loss function may produce the results actually inferior with respect to \emph{DSSIM itself} compared to a simpler loss function we define below, as illustrated in Figure~\ref{fig:dssim-vs-ours}.
LPIPS has a complex energy profile typical for neural networks, and having a neural network as the loss function for another may behave unpredictably~\cite{zhao2017loss}.

The simplest metric capturing the difference between all possible renderings of the depth maps $\depthmap_k$ can be computed as the average root mean square deviation of three basis renderings $\lightdir_m \cdot \normal_k$ in an orthogonal basis $\lightdir_1,\lightdir_2,\lightdir_3$
\begingroup
\small
\setlength\abovedisplayskip{0.5em}
\setlength\belowdisplayskip{0.5em}
\begin{equation}
    \begin{gathered}
        \metric_{\ourmetric}{\depthmap_1, \depthmap_2} = \sqrt{\metric_{\ourmetricinloss}{\depthmap_1, \depthmap_2}},\\
        \metric_{\ourmetricinloss}{\depthmap_1, \depthmap_2} = \frac{1}{3N}\sum_{ij,m} \| \lightdir_m \cdot \normal_{1,ij} - \lightdir_m \cdot \normal_{2,ij} \|_2^2,
    \end{gathered}
\end{equation}
\endgroup
similarly to RMS difference of the normal maps.

We found that this simple metric for depth map comparison is efficient and stable as the loss function
and at the same time, as we demonstrate in Section~\ref{sec:exper}, it is well correlated with DSSIM and LPIPS, \ie, situations when the value of one metric is high and the value of another is low are unlikely.
Our experiments confirm that optimization of this metric also improves both perceptual metrics.

\section{Methods}
\label{sec:methods}
We selected eight representative state-of-the-art depth processing methods based on different principles:
(1)~a purely variational method~\cite{haefner2018fight},
(2)~a bilateral filtering  method that uses a high-resolution edge map~\cite{xie2016edge},
(3)~a dictionary learning method~\cite{gu2017learning},
(4)~a hybrid CNN-variational method~\cite{riegler2016deep},
(5)~a pure CNN~\cite{hui2016depth},
(6)~a zero-shot CNN~\cite{Ulyanov_2018_CVPR},
(7)~a densification~\cite{mal2018sparse} and
(8)~an enhancement~\cite{yan2018ddrnet} CNNs.
Our goals were (a)~to modify the methods for using with the visual difference-based loss function,
and (b)~to compare the results of the modified methods with alternatives of different types.
In our experiments the last two methods did not perform well compared to others, so we did not consider them further.
We found that two neural network-based methods (5) and (6), that we refer to as MSG and DIP, can be easily modified for using with a visual difference-based loss function, as we explain now.

\paragraph{MSG}~\cite{hui2016depth} is a deep learning method that uses different strategies to upsample different spectral components of low-resolution depth map.
In the modified version of this method, that we denote by \textbf{MSG-V},
we replaced the original loss function with a combination of our visual difference-based metric and mean absolute deviation of Laplacian pyramid $\mathrm{Lap}_1$~\cite{pmlr-v80-bojanowski18a} as a regularizer
\begingroup
\setlength\abovedisplayskip{0.5em}
\setlength\belowdisplayskip{0.5em}
\begin{equation}
    \loss{\depthmap_1,\depthmap_2} = \metric_{Lap_1}{\depthmap_1,\depthmap_2} + w\cdot\metric_{\ourmetricinloss}{\depthmap_1,\depthmap_2}.
    \label{eq:ourloss}
\end{equation}
\endgroup

\paragraph{DIP}~\cite{Ulyanov_2018_CVPR} is a zero-shot deep learning approach, based on a remarkable observation that, even without any specialized training, the structure of CNN itself may be leveraged for solving inverse problems on images.
We note that this approach naturally allows simultaneous super-resolution and inpainting.
In this approach, the depth super-resolution problem would be formulated as
\begingroup
\setlength\abovedisplayskip{0.5em}
\setlength\belowdisplayskip{0.5em}
\begin{gather}
	\depthmap_{\theta^*}^\mathrm{SR}=\cnn_{\theta^*},\,\,
	\theta^* = \argmin_\theta\metric_{MSE_d}{\downsamplingoperator\depthmap_{\theta}^\mathrm{SR},\depthmap^\mathrm{LR}},
\end{gather}
\endgroup
where $\depthmap^\mathrm{LR}$ and $\depthmap_{\theta^*}^\mathrm{SR}$ are the low-resolution and super-resolved depth maps,
$\cnn_\theta$ is the output of the deep neural network parametrised by \(\theta\),
\(\downsamplingoperator\) is the downsampling operator,
and $\mathrm{MSE_d}$ is direct mean square difference of the depth maps.
To perform photo-guided super-resolution, we added a second output channel for intensity to the network
\begingroup
\small
\setlength\abovedisplayskip{0.5em}
\setlength\belowdisplayskip{0.5em}
\begin{equation}
    \begin{gathered}
        \depthmap_{\theta^*}^\mathrm{SR}=\cnn_{\theta^*}^{(1)},\quad
        \rendering_{\theta}=\cnn_{\theta}^{(2)},\\
        \theta^* = \argmin_\theta\metric_{MSE_d}{\downsamplingoperator\depthmap_{\theta}^\mathrm{SR},\depthmap^\mathrm{LR}} +w_I\cdot\metric_{Lap_1}{\rendering_{\theta},\rendering^\mathrm{HR}},
    \end{gathered}
\end{equation}
\endgroup
where $\rendering^\mathrm{HR}$ is the high-resolution photo guidance,
and for visually-based version \textbf{DIP-V} we further replaced the direct depth deviation $\mathrm{MSE_d}$ with the function from Equation~\ref{eq:ourloss}.

We used the remaining four methods (1)-(4) for comparison as-is, as modifying them for a different loss function would require substantial changes to the algorithms.

\paragraph{SRfS}~\cite{haefner2018fight} is a variational method relying on complimentarity of super-resolution and shape-from-shading problems.
It already includes a visual-difference based term (the remaining methods use depth difference metrics).
\paragraph{EG}~\cite{xie2016edge} approaches the problem via prediction of smooth high-resolution depth edges with Markov random field optimization.
It does not use a loss directly, therefore cannot be easily adapted.
\paragraph{DG}~\cite{gu2017learning} is a depth map enhancement method based on dictionary learning that uses depth difference-based fidelity term.
It makes a number of modeling choices which may not be suitable for a different loss function, and typically does not perform as well as neural network-based methods.
\paragraph{PDN}~\cite{riegler2016deep} is a hybrid method featuring two stages: the first is composed of fully-convolutional layers and predicts a rough super-resolved depth map, and the second performs an unrolled variational optimization, aiming to produce a sharp and noise-free result.

\section{Experiments}
\label{sec:exper}

\subsection{Data}
For evaluation we selected a representative and diverse set of 34 RGBD images featuring synthetic, high-quality real and low-quality real data with different levels of geometric and textural complexity.
We employed four datasets, most common in literature on depth super-resolution.
\emph{ICL-NUIM}~\cite{handa:etal:ICRA2014} includes photo-realistic RGB images along with synthetic depth, free from any acquisition noise.
\emph{Middlebury 2014}~\cite{scharstein2014high}, captured with a structured light system, provides high-quality ground truth for complex real-world scenes.
\emph{SUN RGBD}~\cite{song2015sun} contains images captured with four different consumer-level RGBD cameras: Intel RealSense, Asus Xtion, Microsoft Kinect v1 and v2.
\emph{ToFMark}~\cite{ferstl2013image} provides challenging real-world time-of-flight and intensity camera acquisitions together with an accurate ground truth from a structured light sensor.

In addition, we constructed a synthetic \emph{SimGeo} dataset, that consists of 6 geometrically simple scenes with low- and high-frequency texture, and without any, using Blender.
The purpose of SimGeo were to reveal artifacts that are not related to the noise or high-frequency geometry in the input data,
like false geometric detail caused by color variation on a smooth surface.

We resized and cropped each RGBD image to the resolution of $512\times512$ and generated low-resolution input depth maps with the scaling factors of 4 and 8, that are most common among the works on depth super-resolution.
We focused on two downsampling models: Box, \ie, each low-resolution pixel contains the mean value over the \textquote{box} neighbouring high-resolution pixels, and Nearest neighbour, \ie, each low-resolution pixel contains the value of the nearest high-resolution pixel.
For additional details on our evaluation data and the results for different downsampling models please refer to supplementary material.

\subsection{Evaluation details}
To quantify the performance of the methods, we measured direct RMS deviation of the depth maps (denoted by $\mathrm{RMSE_d}$) and deviation of their renderings with the metrics described in Section~\ref{sec:metrics}.
For visually-based metrics we calculated their values for three orthogonal light directions, corresponding to the three left-most images in Figure~\ref{fig:light_directions}, and the value for an additional light direction, corresponding to the right-most image.
We then took the worst of the four values.
With similar outcomes, we also explored different reducing strategies and a set of different metrics: BadPix and Bumpiness, applied directly to depth values, and BadPix and RMSE applied to separate depth map renderings.

Additionally, we conducted an informal perceptual study using the results on SimGeo, ICL-NUIM and Middlebury datasets, in which subjects were asked to choose the renderings of the upsampled depth maps that look most similar to the ground truth.

\subsection{Implementation details}
We evaluated publicly available trained models for EG, DG, and MSG and trained PDN using publicly available code;
we used the implementation of SRfS provided by the authors;
we adapted publicly available implementation of DIP for depth maps, as described in Section~\ref{sec:methods};
we reimplemented MSG-V in PyTorch~\cite{paszke2017automatic} and trained it according to the original paper using the patches from Middlebury and MPI Sintel~\cite{Butler:ECCV:2012}.
We selected the value of the weighting parameter $w$ in Equation~(\ref{eq:ourloss}) so that both terms of the loss contribute equally with respect to their magnitudes (see supplementary material for more details).

\subsection{Comparison of quality measures}
\begin{figure*}[p!]
    \centerline{\begin{tikzpicture}
        \node[anchor=south west,inner sep=0] (image) at (0,0) {%
            \includegraphics[width=\linewidth,height=2.6cm,keepaspectratio]{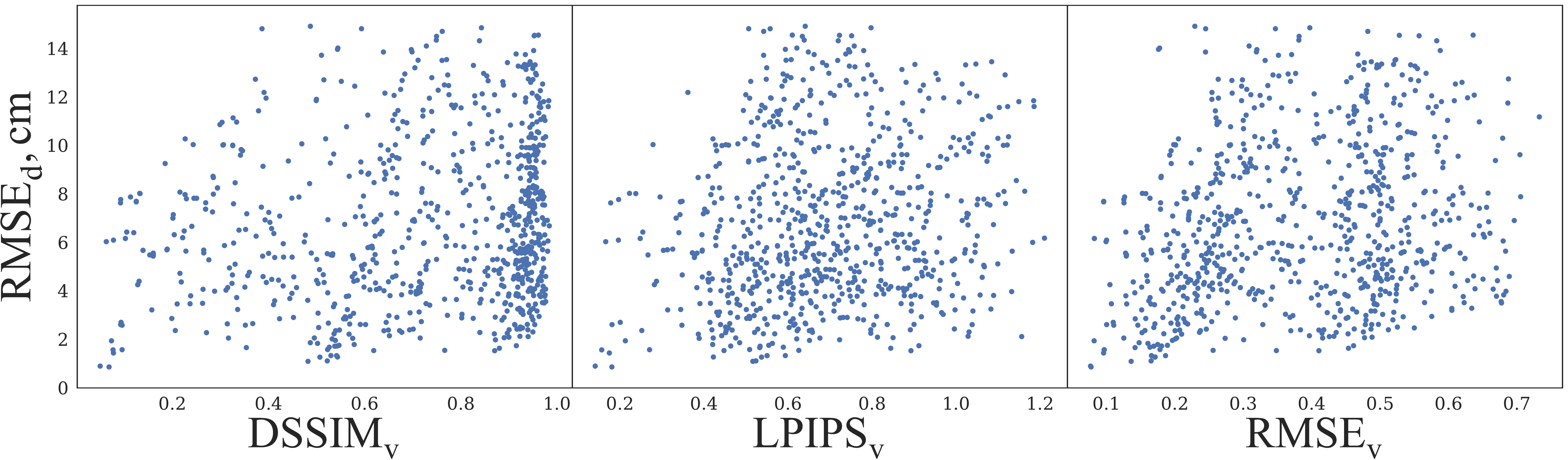}%
            \includegraphics[width=\linewidth,height=2.6cm,keepaspectratio]{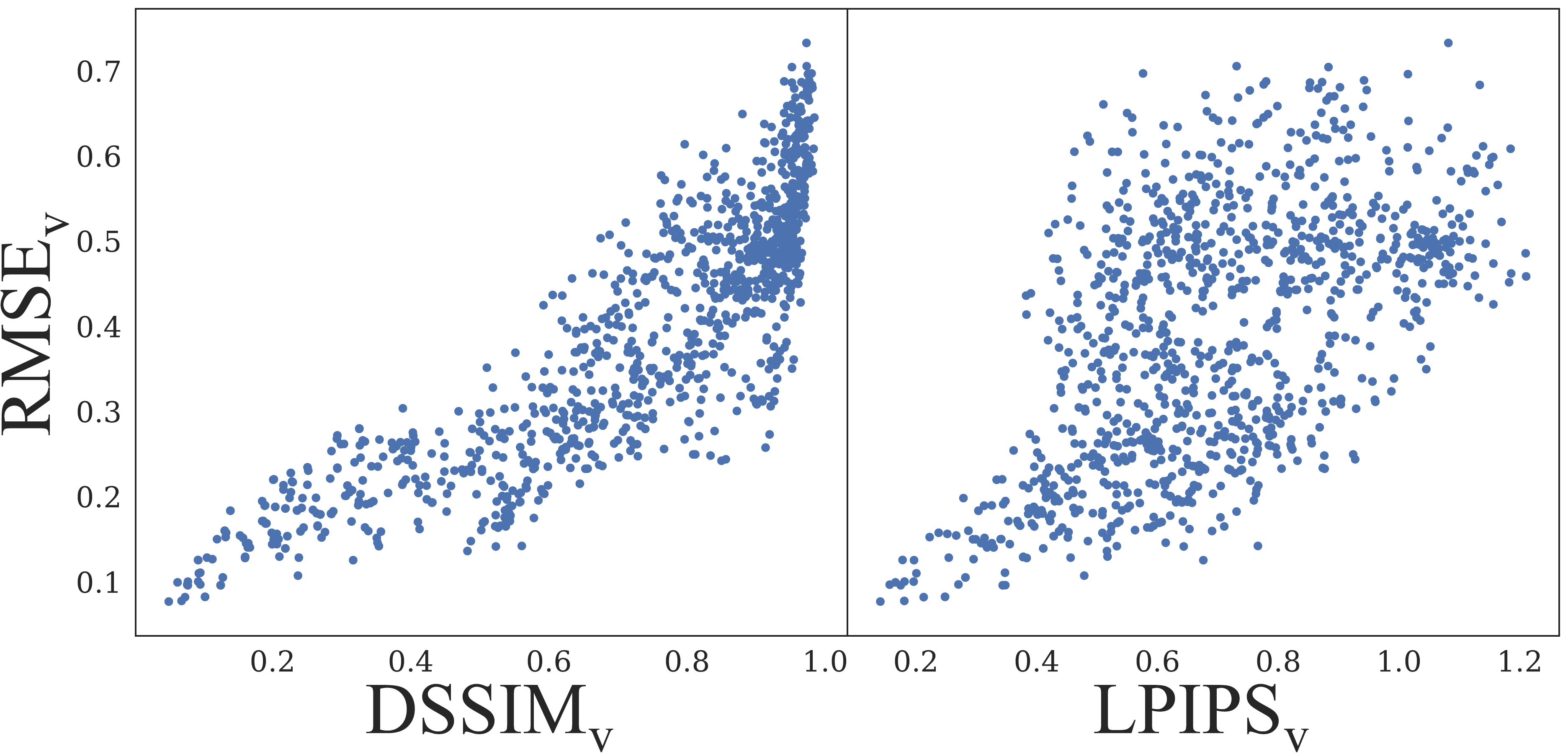}%
            \includegraphics[width=\linewidth,height=2.6cm,keepaspectratio]{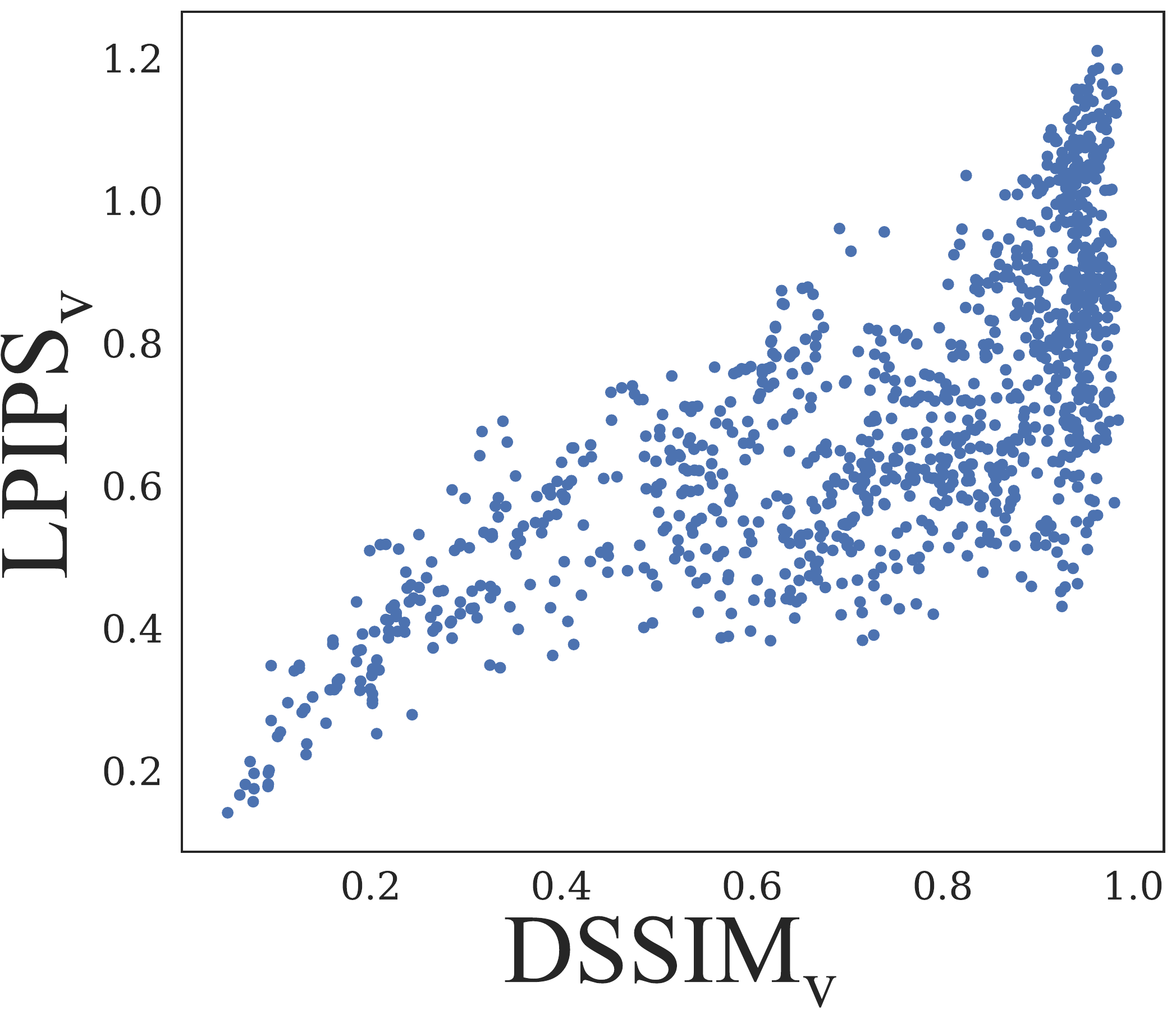}%
		};
        \begin{scope}[x={(image.south east)},y={(image.north west)}]
			\node[fill=white,inner sep=1pt] at (0.05,0.9) {0.09};
			\node[fill=white,inner sep=1pt] at (0.212,0.9) {0.05};
			\node[fill=white,inner sep=1pt] at (0.374,0.9) {0.12};
			\node[fill=white,inner sep=1pt] at (0.564,0.9) {0.85};
			\node[fill=white,inner sep=1pt] at (0.706,0.9) {0.57};
			\node[fill=white,inner sep=1pt] at (0.876,0.9) {0.75};
        \end{scope}
    \end{tikzpicture}}
	\caption{Scatter plots demonstrating correlation of quality measures, and the corresponding values of the Pearson correlation coefficient in the corner.
	Each point represents one super-resolution result.}
	\label{fig:scatterplots}
\end{figure*}
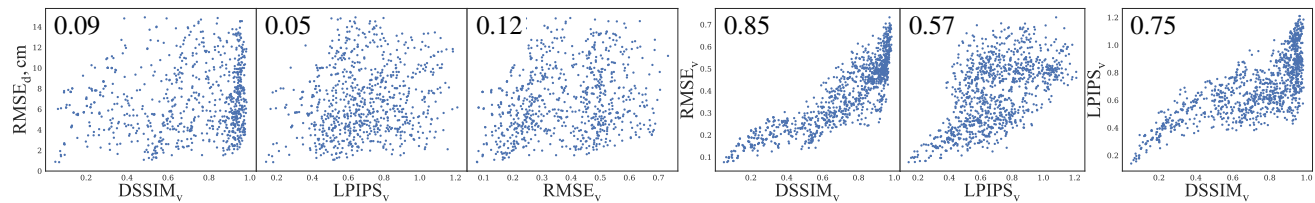

To quantify how well different metrics represent the visual quality of a super-resolved depth map,
we compared pairwise correlations of these metrics and calculated the corresponding values of Pearson correlation coefficient.
Since LPIPS as a neural network-based perceptual metric has been experimentally shown to represent human perception well, we used its value as the reference.
We found that the metrics based on direct depth deviation demonstrate weak correlation with perceptual metrics, as illustrated in Figure~\ref{fig:scatterplots} for $\mathrm{RMSE_d}$,
and hence are not suitable for measuring the depth map quality when the visual appearance plays an important role.
On the other hand, we found that our $\ourmetric$ correlates well with perceptual metrics, to the same extent they correlate with each other (see Figure~\ref{fig:scatterplots}).

\subsection{Comparison of super-resolution methods}
\begin{figure*}[p!]
	\centerline{\begin{tikzpicture}
		\node[anchor=south west,inner sep=0] (image) at (0,0) {\includegraphics[width=\linewidth]{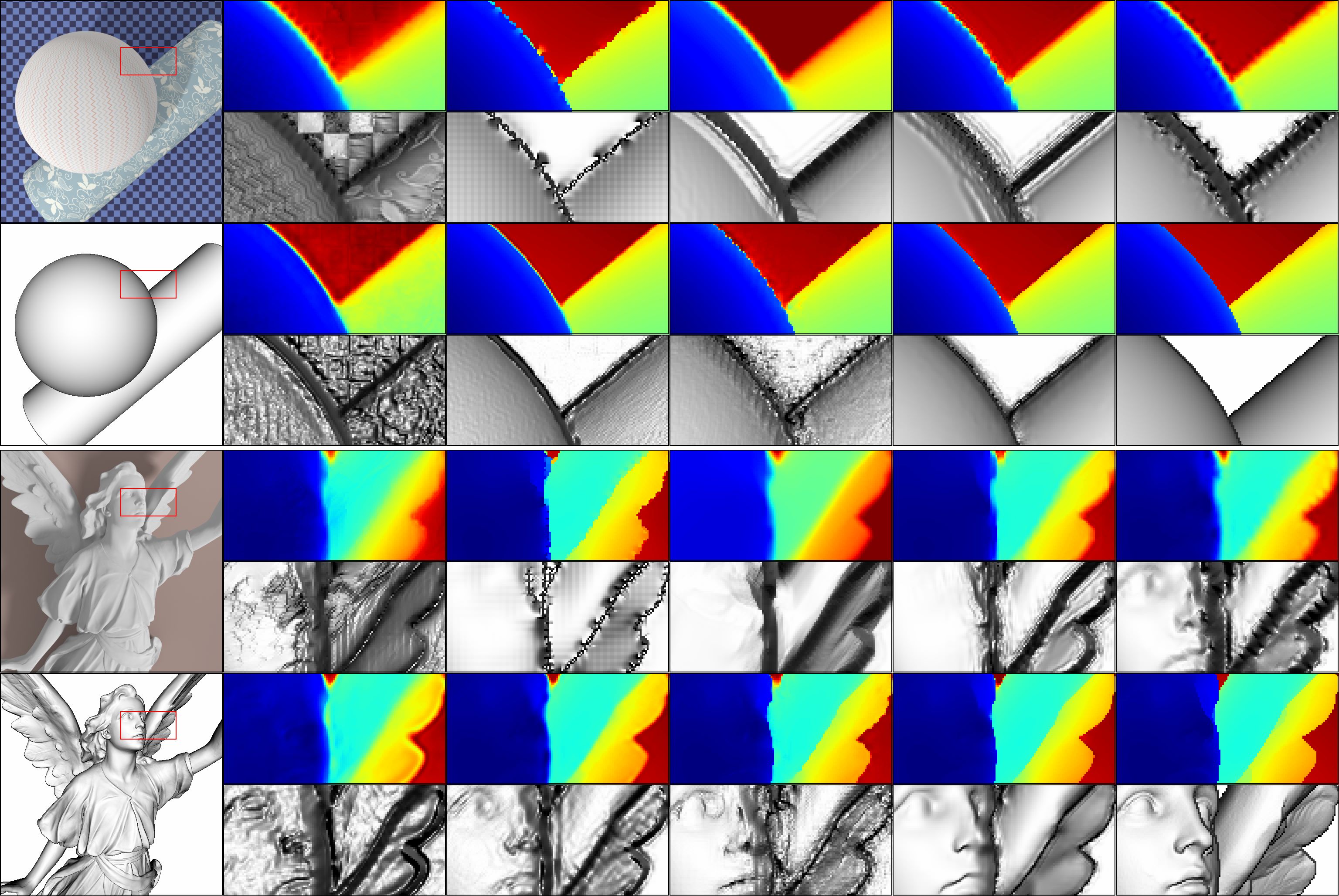}};
        \begin{scope}[x={(image.south east)},y={(image.north west)}]
            \node[fill=white,inner sep=1pt] at (0.211,0.769) {SRfS~\cite{haefner2018fight}};
			\node[fill=white,inner sep=1pt] at (0.37,0.769) {EG~\cite{xie2016edge}};
			\node[fill=white,inner sep=1pt] at (0.543,0.769) {PDN~\cite{riegler2016deep}};
			\node[fill=white,inner sep=1pt] at (0.703,0.769) {DG~\cite{gu2017learning}};
			\node[fill=white,inner sep=1pt] at (0.868,0.769) {Bicubic};
			\node[fill=white,inner sep=1pt] at (0.205,0.52) {DIP~\cite{Ulyanov_2018_CVPR}};
			\node[fill=white,inner sep=1pt] at (0.365,0.52) {DIP-V};
			\node[fill=white,inner sep=1pt] at (0.545,0.52) {MSG~\cite{hui2016depth}};
			\node[fill=white,inner sep=1pt] at (0.702,0.52) {MSG-V};
			\node[fill=white,inner sep=1pt] at (0.85,0.52) {GT};
			\node[fill=white,inner sep=1pt] at (0.211,0.266) {SRfS~\cite{haefner2018fight}};
			\node[fill=white,inner sep=1pt] at (0.37,0.266) {EG~\cite{xie2016edge}};
			\node[fill=white,inner sep=1pt] at (0.543,0.266) {PDN~\cite{riegler2016deep}};
			\node[fill=white,inner sep=1pt] at (0.703,0.266) {DG~\cite{gu2017learning}};
			\node[fill=white,inner sep=1pt] at (0.868,0.266) {Bicubic};
			\node[fill=white,inner sep=1pt] at (0.205,0.017) {DIP~\cite{Ulyanov_2018_CVPR}};
			\node[fill=white,inner sep=1pt] at (0.365,0.017) {DIP-V};
			\node[fill=white,inner sep=1pt] at (0.545,0.017) {MSG~\cite{hui2016depth}};
			\node[fill=white,inner sep=1pt] at (0.702,0.017) {MSG-V};
			\node[fill=white,inner sep=1pt] at (0.85,0.017) {GT};
        \end{scope}
    \end{tikzpicture}}
	\caption{Super-resolution results on \textquote{Sphere and cylinder} and \textquote{Lucy} from SimGeo with the scaling factor of 4.
	Depth maps are in pseudo-color and depth map renderings are in grayscale.
	Best viewed in color.}
	\label{fig:sphere_lucy}
\end{figure*}
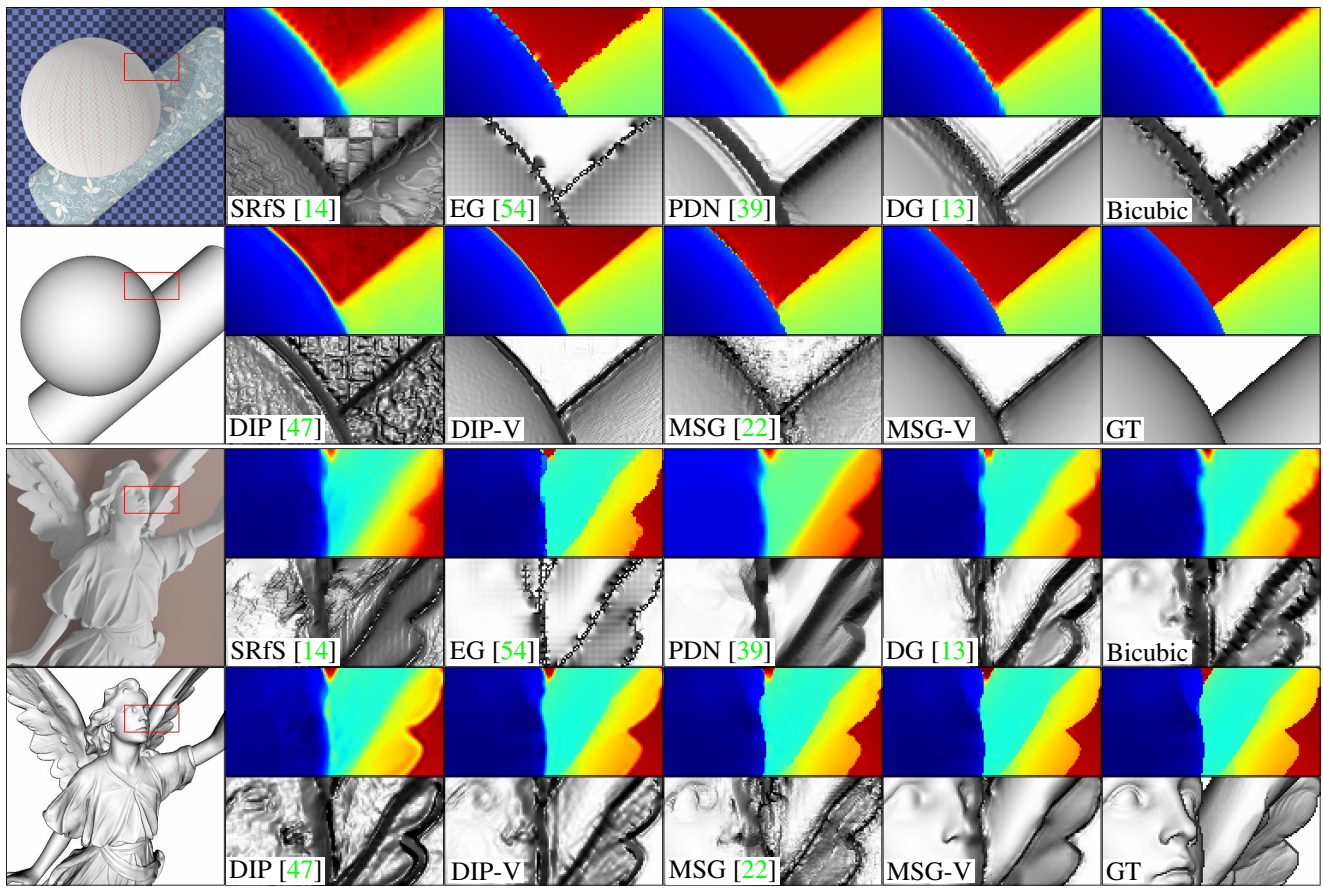

\begin{figure*}[p!]
	\centerline{\begin{tikzpicture}
		\node[anchor=south west,inner sep=0] (image) at (0,0) {\includegraphics[width=\linewidth]{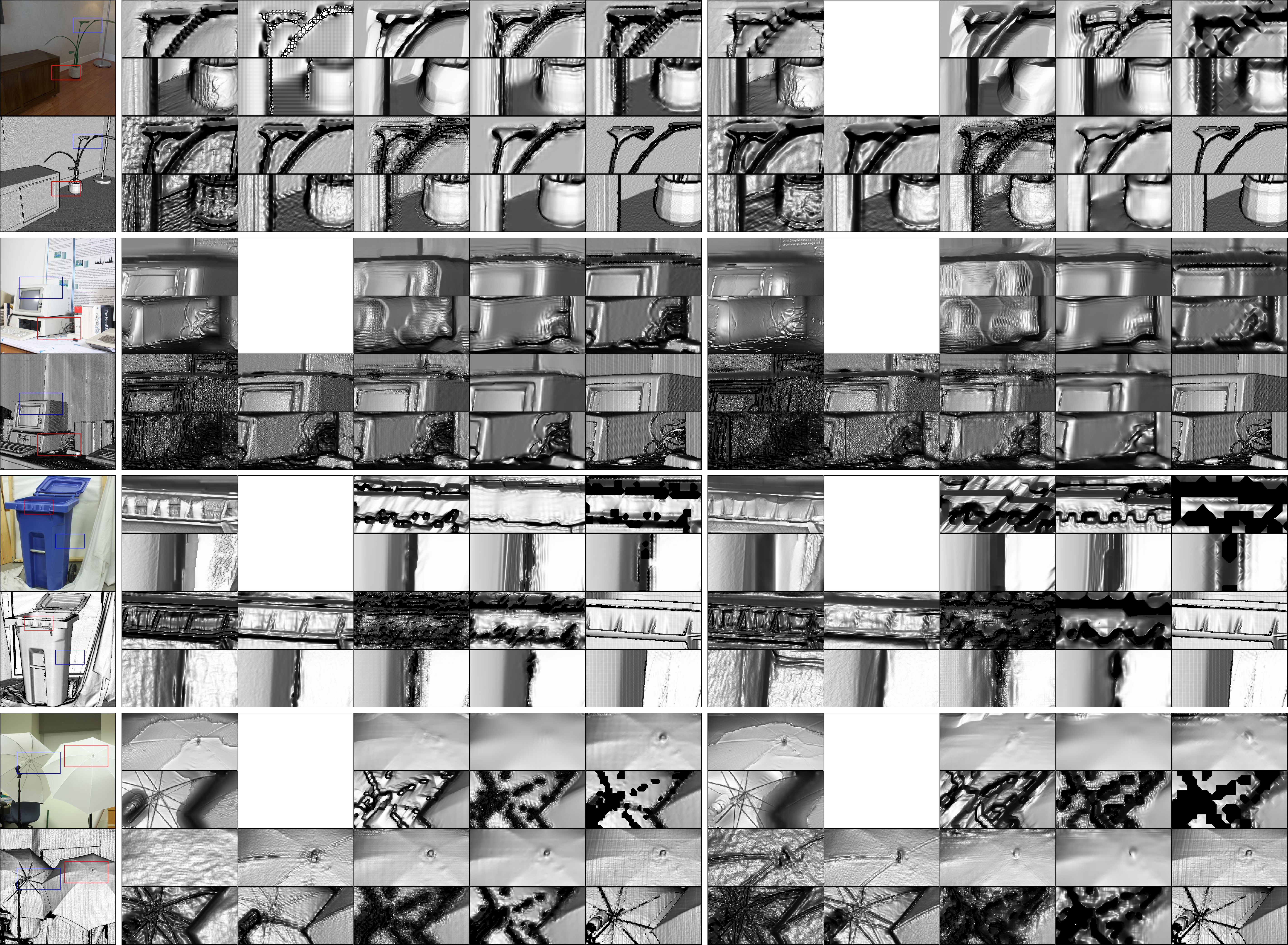}};
		\begin{scope}[x={(image.south east)},y={(image.north west)}]
			\begingroup
			\tiny
            % plant x4
			\node[fill=white,inner sep=0.5pt] at (0.1165,0.8845) {SRfS~\cite{haefner2018fight}};
			\node[fill=white,inner sep=0.5pt] at (0.203,0.8845) {EG~\cite{xie2016edge}};
			\node[fill=white,inner sep=0.5pt] at (0.296,0.8845) {PDN~\cite{riegler2016deep}};
			\node[fill=white,inner sep=0.5pt] at (0.3835,0.8845) {DG~\cite{gu2017learning}};
			\node[fill=white,inner sep=0.5pt] at (0.472,0.8845) {Bicubic};
			\node[fill=white,inner sep=0.5pt] at (0.114,0.762) {DIP~\cite{Ulyanov_2018_CVPR}};
			\node[fill=white,inner sep=0.5pt] at (0.2,0.762) {DIP-V};
			\node[fill=white,inner sep=0.5pt] at (0.297,0.762) {MSG~\cite{hui2016depth}};
			\node[fill=white,inner sep=0.5pt] at (0.3825,0.762) {MSG-V};
			\node[fill=white,inner sep=0.5pt] at (0.463,0.762) {GT};
			% plant x8
			\node[fill=white,inner sep=0.5pt] at (0.1165+.455,0.8845) {SRfS~\cite{haefner2018fight}};
			\node[fill=white,inner sep=0.5pt] at (0.296+.455,0.8845) {PDN~\cite{riegler2016deep}};
			\node[fill=white,inner sep=0.5pt] at (0.3835+.455,0.8845) {DG~\cite{gu2017learning}};
			\node[fill=white,inner sep=0.5pt] at (0.472+.455,0.8845) {Bicubic};
			\node[fill=white,inner sep=0.5pt] at (0.114+.455,0.762) {DIP~\cite{Ulyanov_2018_CVPR}};
			\node[fill=white,inner sep=0.5pt] at (0.2+.455,0.762) {DIP-V};
			\node[fill=white,inner sep=0.5pt] at (0.297+.455,0.762) {MSG~\cite{hui2016depth}};
			\node[fill=white,inner sep=0.5pt] at (0.3825+.455,0.762) {MSG-V};
			\node[fill=white,inner sep=0.5pt] at (0.463+.455,0.762) {GT};
			% vintage x4
			\node[fill=white,inner sep=0.5pt] at (0.1165,0.8845-0.251) {SRfS~\cite{haefner2018fight}};
			\node[fill=white,inner sep=0.5pt] at (0.219,0.8845-0.251) {EG~\cite{xie2016edge} -- failed};
			\node[fill=white,inner sep=0.5pt] at (0.296,0.8845-0.251) {PDN~\cite{riegler2016deep}};
			\node[fill=white,inner sep=0.5pt] at (0.3835,0.8845-0.251) {DG~\cite{gu2017learning}};
			\node[fill=white,inner sep=0.5pt] at (0.472,0.8845-0.251) {Bicubic};
			\node[fill=white,inner sep=0.5pt] at (0.114,0.762-0.251) {DIP~\cite{Ulyanov_2018_CVPR}};
			\node[fill=white,inner sep=0.5pt] at (0.2,0.762-0.251) {DIP-V};
			\node[fill=white,inner sep=0.5pt] at (0.297,0.762-0.251) {MSG~\cite{hui2016depth}};
			\node[fill=white,inner sep=0.5pt] at (0.3825,0.762-0.251) {MSG-V};
			\node[fill=white,inner sep=0.5pt] at (0.463,0.762-0.251) {GT};
			% vintage x8
			\node[fill=white,inner sep=0.5pt] at (0.1165+.455,0.8845-0.251) {SRfS~\cite{haefner2018fight}};
			\node[fill=white,inner sep=0.5pt] at (0.296+.455,0.8845-0.251) {PDN~\cite{riegler2016deep}};
			\node[fill=white,inner sep=0.5pt] at (0.3835+.455,0.8845-0.251) {DG~\cite{gu2017learning}};
			\node[fill=white,inner sep=0.5pt] at (0.472+.455,0.8845-0.251) {Bicubic};
			\node[fill=white,inner sep=0.5pt] at (0.114+.455,0.762-0.251) {DIP~\cite{Ulyanov_2018_CVPR}};
			\node[fill=white,inner sep=0.5pt] at (0.2+.455,0.762-0.251) {DIP-V};
			\node[fill=white,inner sep=0.5pt] at (0.297+.455,0.762-0.251) {MSG~\cite{hui2016depth}};
			\node[fill=white,inner sep=0.5pt] at (0.3825+.455,0.762-0.251) {MSG-V};
			\node[fill=white,inner sep=0.5pt] at (0.463+.455,0.762-0.251) {GT};
			% recycle x4
			\node[fill=white,inner sep=0.5pt] at (0.1165,0.8845-0.502) {SRfS~\cite{haefner2018fight}};
			\node[fill=white,inner sep=0.5pt] at (0.219,0.8845-0.502) {EG~\cite{xie2016edge} -- failed};
			\node[fill=white,inner sep=0.5pt] at (0.296,0.8845-0.502) {PDN~\cite{riegler2016deep}};
			\node[fill=white,inner sep=0.5pt] at (0.3835,0.8845-0.502) {DG~\cite{gu2017learning}};
			\node[fill=white,inner sep=0.5pt] at (0.472,0.8845-0.502) {Bicubic};
			\node[fill=white,inner sep=0.5pt] at (0.114,0.762-0.502) {DIP~\cite{Ulyanov_2018_CVPR}};
			\node[fill=white,inner sep=0.5pt] at (0.2,0.762-0.502) {DIP-V};
			\node[fill=white,inner sep=0.5pt] at (0.297,0.762-0.502) {MSG~\cite{hui2016depth}};
			\node[fill=white,inner sep=0.5pt] at (0.3825,0.762-0.502) {MSG-V};
			\node[fill=white,inner sep=0.5pt] at (0.463,0.762-0.502) {GT};
			% recycle x8
			\node[fill=white,inner sep=0.5pt] at (0.1165+.455,0.8845-0.502) {SRfS~\cite{haefner2018fight}};
			\node[fill=white,inner sep=0.5pt] at (0.296+.455,0.8845-0.502) {PDN~\cite{riegler2016deep}};
			\node[fill=white,inner sep=0.5pt] at (0.3835+.455,0.8845-0.502) {DG~\cite{gu2017learning}};
			\node[fill=white,inner sep=0.5pt] at (0.472+.455,0.8845-0.502) {Bicubic};
			\node[fill=white,inner sep=0.5pt] at (0.114+.455,0.762-0.502) {DIP~\cite{Ulyanov_2018_CVPR}};
			\node[fill=white,inner sep=0.5pt] at (0.2+.455,0.762-0.502) {DIP-V};
			\node[fill=white,inner sep=0.5pt] at (0.297+.455,0.762-0.502) {MSG~\cite{hui2016depth}};
			\node[fill=white,inner sep=0.5pt] at (0.3825+.455,0.762-0.502) {MSG-V};
			\node[fill=white,inner sep=0.5pt] at (0.463+.455,0.762-0.502) {GT};
			% umbrella x4
			\node[fill=white,inner sep=0.5pt] at (0.1165,0.8845-0.753) {SRfS~\cite{haefner2018fight}};
			\node[fill=white,inner sep=0.5pt] at (0.219,0.8845-0.753) {EG~\cite{xie2016edge} -- failed};
			\node[fill=white,inner sep=0.5pt] at (0.296,0.8845-0.753) {PDN~\cite{riegler2016deep}};
			\node[fill=white,inner sep=0.5pt] at (0.3835,0.8845-0.753) {DG~\cite{gu2017learning}};
			\node[fill=white,inner sep=0.5pt] at (0.472,0.8845-0.753) {Bicubic};
			\node[fill=white,inner sep=0.5pt] at (0.114,0.762-0.753) {DIP~\cite{Ulyanov_2018_CVPR}};
			\node[fill=white,inner sep=0.5pt] at (0.2,0.762-0.753) {DIP-V};
			\node[fill=white,inner sep=0.5pt] at (0.297,0.762-0.753) {MSG~\cite{hui2016depth}};
			\node[fill=white,inner sep=0.5pt] at (0.3825,0.762-0.753) {MSG-V};
			\node[fill=white,inner sep=0.5pt] at (0.463,0.762-0.753) {GT};
			% umbrella x8
			\node[fill=white,inner sep=0.5pt] at (0.1165+.455,0.8845-0.753) {SRfS~\cite{haefner2018fight}};
			\node[fill=white,inner sep=0.5pt] at (0.296+.455,0.8845-0.753) {PDN~\cite{riegler2016deep}};
			\node[fill=white,inner sep=0.5pt] at (0.3835+.455,0.8845-0.753) {DG~\cite{gu2017learning}};
			\node[fill=white,inner sep=0.5pt] at (0.472+.455,0.8845-0.753) {Bicubic};
			\node[fill=white,inner sep=0.5pt] at (0.114+.455,0.762-0.753) {DIP~\cite{Ulyanov_2018_CVPR}};
			\node[fill=white,inner sep=0.5pt] at (0.2+.455,0.762-0.753) {DIP-V};
			\node[fill=white,inner sep=0.5pt] at (0.297+.455,0.762-0.753) {MSG~\cite{hui2016depth}};
			\node[fill=white,inner sep=0.5pt] at (0.3825+.455,0.762-0.753) {MSG-V};
			\node[fill=white,inner sep=0.5pt] at (0.463+.455,0.762-0.753) {GT};
			\endgroup
			% decorations
			\draw (0.0945,-.005) -- (0.5445,-.005);
			\node at (0.3195,-.02) {$\times4$};
			\draw (0.0945+.455,-.005) -- (0.5445+.455,-.005);
			\node at (0.3195+.455,-.02) {$\times8$};
		\end{scope}
	\end{tikzpicture}}
	\caption{Depth map renderings corresponding to super-resolution results on \textquote{Plant} from \textquote{ICL-NUIM} and \textquote{Vintage}, \textquote{Recycle} and \textquote{Umbrella} from Middlebury datasets
	with the scaling factor of 4 on the left and the scaling factor of 8 on the right.
	Best viewed in large scale.}
	\label{fig:plant_vintage_recycle}
\end{figure*}

In Table~\ref{tab:simgeo} and Figure~\ref{fig:sphere_lucy} we present the super-resolution results on our SimGeo dataset with the scaling factor of 4;
in Table~\ref{tab:icl_middlebury} and Figure~\ref{fig:plant_vintage_recycle} we present the results on ICL-NUIM and Middlebury datasets with the scaling factors of 4 and 8.
We use Box downsampling model in both cases.
Please find the additional results in supplementary material or online\footnote{\href{https://mega.nz/\#F!yvRXBABI!pucRoBvtnthzHI1oqsxEvA!y6JmCajS}{mega.nz/\#F!yvRXBABI!pucRoBvtnthzHI1oqsxEvA!y6JmCajS}}.

In general, we found that the methods EG, PDN and DG do not recover fine details of the surface,
typically oversmoothing the result in comparison to, \eg, Bicubic upsampling,
the methods SRfS and original DIP suffer from false geometry artifacts in case of a smooth textured surface,
and original MSG introduces severe noise around the depth edges.
As illustrated in Figure~\ref{fig:plant_vintage_recycle} and Table~\ref{tab:icl_middlebury},
all the methods from prior works perform relatively poorly on the images with regions of missing depth measurements (rendered in black),
including the ones that inpaint these regions explicitly (SRfS, DG) or implicitly (DIP).
The method EG failed to converge on some images.

In contrast, we observed that integration of our visual difference-based loss into DIP and MSG significantly improved the results of both methods qualitatively and quantitatively.
The visual difference-based version DIP-V do not suffer from false geometry artifacts as much as the original version.
On the challenging images from Middlebury dataset, where it performed simultaneous super-resolution and inpainting,
DIP-V mostly outperformed other methods as measured by the perceptual metrics and was preferred by more than 80\% of subjects in the perceptual study.
The visual difference-based version MSG-V produces significantly less noisy results in comparison to the original version, in some cases almost without any noticeable artifacts.
On the data without missing measurements, including hole-filled \textquote{Vintage} from Middlebury,
MSG-V mostly outperformed other methods as measured by the perceptual metrics and was preferred by more than 80\% of subjects.
On SimGeo, ICL-NUIM and Middlebury combined, one of our modified versions, DIP-V or MSG-V, was preferred over the other methods by more than 85\% of subjects.

For reference, in Figure~\ref{fig:sphere_lucy} we include pseudo-color visualizations of the depth maps.
Notice that while the upsampled depth maps obtained with different methods are almost indistinguishable in this form of visualization, commonly used in the literature on depth processing for qualitative evaluation,
the corresponding renderings and, consequently, the underlying geometry varies dramatically.

\begin{table*}[p!]
\footnotesize
\setlength\tabcolsep{\widthof{0}*\real{.4}}
\setlength\aboverulesep{0pt}
\setlength\belowrulesep{0pt}
\centering
\begin{tabular}{l|cccc|cccc|cccc|cccc}
	\midrule
	\multicolumn{1}{c}{}         & \multicolumn{4}{c|}{Sphere and cylinder, x4}                          & \multicolumn{4}{c|}{Lucy, x4}                                         & \multicolumn{4}{c|}{Cube, x4}                                         & \multicolumn{4}{c}{SimGeo average, x4}                                \\
	           \cmidrule{2-17}
	\multicolumn{1}{c}{}         &            \fontsize{6}{7}$\mathrm{RMSE_d}$ &           \fontsize{6}{7}$\mathrm{DSSIM_v}$ &           \fontsize{6}{7}$\mathrm{LPIPS_v}$ &            \fontsize{6}{7}$\ourmetric$ &            \fontsize{6}{7}$\mathrm{RMSE_d}$ &           \fontsize{6}{7}$\mathrm{DSSIM_v}$ &           \fontsize{6}{7}$\mathrm{LPIPS_v}$ &            \fontsize{6}{7}$\ourmetric$ &            \fontsize{6}{7}$\mathrm{RMSE_d}$ &           \fontsize{6}{7}$\mathrm{DSSIM_v}$ &           \fontsize{6}{7}$\mathrm{LPIPS_v}$ &            \fontsize{6}{7}$\ourmetric$ &            \fontsize{6}{7}$\mathrm{RMSE_d}$ &           \fontsize{6}{7}$\mathrm{DSSIM_v}$ &           \fontsize{6}{7}$\mathrm{LPIPS_v}$ &            \fontsize{6}{7}$\ourmetric$ \\
	\midrule                                                                                                                                                                                                                                                                                                                     
	SRfS~\cite{haefner2018fight} &              70 &             887 &            1025 &             417 &              82 &             811 &             781 &             367 &              52 &             934 &            1036 &             361 &              61 &             711 &             869 &             311 \\
	EG~\cite{xie2016edge}        &              55 & \underline{143} &             326 & \underline{130} &              69 &             357 &             426 & \underline{220} &              43 & \underline{113} & \underline{214} & \underline{105} &              53 & \underline{168} &             306 & \underline{136} \\
	PDN~\cite{riegler2016deep}   &             157 &             198 & \underline{295} &             150 &             173 &             456 & \underline{368} &             251 &             164 &             156 &             250 &             145 &             162 &             224 & \underline{278} &             165 \\
	DG~\cite{gu2017learning}     &              56 &             265 &             372 &             166 &              69 &             523 &             558 &             249 &              44 &             218 &             411 &             139 &              54 &             293 &             420 &             171 \\
	Bicubic                      &              57 &             189 &             313 &             189 &              72 & \underline{355} &             398 &             267 &              44 &             131 &             287 &             160 &              55 &             197 &             320 &             193 \\
	DIP~\cite{Ulyanov_2018_CVPR} &              46 &             965 &            1062 &             548 &  \underline{53} &             827 &             615 &             344 &              45 &             963 &             906 &             530 &              52 &             887 &             893 &             395 \\
	MSG~\cite{hui2016depth}      &  \underline{41} &             626 &             859 &             229 &              54 &             444 &             480 &             259 &  \underline{29} &             445 &             687 &             176 &  \underline{39} &             374 &             569 &             194 \\
	\midrule
	DIP-V                        &     \textbf{28} &             560 &             766 &             142 &     \textbf{44} &             421 &             446 &             223 &     \textbf{26} &             352 &             613 &             146 &     \textbf{33} &             313 &             524 &             147 \\
	MSG-V                        &              99 &     \textbf{94} &    \textbf{267} &     \textbf{96} &              74 &    \textbf{205} &    \textbf{251} &    \textbf{156} &             102 &     \textbf{70} &    \textbf{179} &     \textbf{77} &              96 &     \textbf{95} &    \textbf{194} &     \textbf{99} \\
	\midrule
\end{tabular}
\caption{Quantitative evaluation on SimGeo dataset. $\mathrm{RMSE_d}$ is in millimeters, other metrics are in thousandths.
Lower values correspond to better results.
The best result is in bold, the second best is underlined.}
\label{tab:simgeo}
\end{table*}

\begin{table*}[p!]
\footnotesize
\setlength\tabcolsep{\widthof{0}*\real{.85}}
\setlength\aboverulesep{0pt}
\setlength\belowrulesep{0pt}
\centering
\begin{tabular}{l|cc|cc|cc|cc|cc|cc|cc|cc|cc|cc|cc|cc}
	\midrule
	\multicolumn{1}{c}{}         & \multicolumn{6}{c|}{Plant}                                                                                & \multicolumn{6}{c|}{Vintage}                                                                              & \multicolumn{6}{c|}{Recycle}                                                                              & \multicolumn{6}{c}{Umbrella}                                                                              \\
	           \cmidrule{2-25}                                                                                                                                                                                                                                                                                                   
	\multicolumn{1}{c}{}         & \multicolumn{2}{c}{$\mathrm{DSSIM_v}$}         & \multicolumn{2}{c}{$\mathrm{LPIPS_v}$}         & \multicolumn{2}{c|}{$\ourmetric$}         & \multicolumn{2}{c}{$\mathrm{DSSIM_v}$}         & \multicolumn{2}{c}{$\mathrm{LPIPS_v}$}         & \multicolumn{2}{c|}{$\ourmetric$}         & \multicolumn{2}{c}{$\mathrm{DSSIM_v}$}         & \multicolumn{2}{c}{$\mathrm{LPIPS_v}$}         & \multicolumn{2}{c|}{$\ourmetric$}         & \multicolumn{2}{c}{$\mathrm{DSSIM_v}$}         & \multicolumn{2}{c}{$\mathrm{LPIPS_v}$}         & \multicolumn{2}{c}{$\ourmetric$}          \\
  \multicolumn{1}{c}{}         &            x4   &            x8   &            x4   &              x8 &            x4   &            x8   &            x4   &              x8 &            x4   &            x8   &            x4   &              x8 &            x4   &            x8   &            x4   &              x8 &            x4   &              x8 &            x4   &            x8   &            x4   &              x8 &            x4   &              x8 \\
	\midrule                                                                                                                                                                                                                                                                                                                     
	SRfS~\cite{haefner2018fight} &             658 &             692 &             632 &             649 &             280 &             309 &             721 &             749 &             631 & \underline{634} &             346 &             382 &             715 &             772 &             610 &             623 &             376 &             410 &             843 &             853 &             797 &             831 &             397 &             443 \\
	EG~\cite{xie2016edge}        &             568 &                 &             677 &                 &             255 &                 &                 &                 &                 &                 &                 &                 &                 &                 &                 &                 &                 &                 &                 &                 &                 &                 &                 &                 \\
	PDN~\cite{riegler2016deep}   &             574 &             612 &             659 &             699 &             269 &             305 &             663 &             714 &             706 &             700 &             319 &             350 &             635 &    \textbf{701} &             523 &             589 &             364 &             457 &             799 &    \textbf{828} &             847 &             882 &             367 &             452 \\
	DG~\cite{gu2017learning}     &             611 &             622 &             745 &             785 &             268 &             291 &             666 &             669 &             796 &             840 &             290 & \underline{300} &             696 & \underline{719} &             602 &             617 & \underline{328} & \underline{383} &             846 &             878 &             781 &             856 &             399 &             457 \\
	Bicubic                      & \underline{562} & \underline{610} &             688 &             763 &             249 &             290 & \underline{558} & \underline{649} &             602 &             729 & \underline{258} &             302 &    \textbf{575} &             721 & \underline{474} &             576 &             329 &             398 &    \textbf{749} & \underline{837} &             747 &             886 & \underline{323} & \underline{380} \\
	DIP~\cite{Ulyanov_2018_CVPR} &             919 &             880 &             764 &             723 &             490 &             437 &             953 &             965 &             910 &             872 &             656 &             687 &             871 &             923 &             576 &             605 &             434 &             500 &             915 &             953 &             737 & \underline{722} &             467 &             528 \\
	MSG~\cite{hui2016depth}      &             571 &             645 & \underline{582} &    \textbf{495} & \underline{234} &             285 &             708 &             785 &    \textbf{510} &    \textbf{610} &             292 &             364 &             741 &             869 &             624 &             661 &             485 &             550 &             834 &             896 & \underline{678} &             787 &             442 &             496 \\
	\midrule
	DIP-V                        &             694 &             707 &    \textbf{463} & \underline{555} &             262 & \underline{276} &             804 &             884 & \underline{579} &             674 &             343 &             435 &    \textbf{575} &             735 &    \textbf{388} &    \textbf{485} &    \textbf{273} &    \textbf{332} &             796 &             854 &    \textbf{604} &    \textbf{598} &    \textbf{318} &    \textbf{352} \\
	MSG-V                        &    \textbf{524} &    \textbf{575} &             639 &             720 &    \textbf{194} &    \textbf{236} &    \textbf{536} &    \textbf{643} &             670 &             702 &    \textbf{211} &    \textbf{268} & \underline{603} &             737 &             520 & \underline{564} &             368 &             473 & \underline{778} &             842 &             800 &             890 &             348 &             427 \\
	\midrule
\end{tabular}
\caption{Quantitative evaluation on ICL-NUIM and Middlebury datasets.
All metrics are in thousandths.
Lower values correspond to better results.
The best result is in bold, the second best is underlined.}
\label{tab:icl_middlebury}
\end{table*}

\section{Conclusion}
\label{sec:concl}
We have explored depth map super-resolution with a simple visual difference-based metric as the loss function.
Via comparison of this metric with a variety of perceptual quality measures, we have demonstrated that it can be considered a reasonable proxy for human perception in the problem of depth super-resolution with the focus on visual quality of the 3D surface.
Via an extensive evaluation of several depth-processing methods on a range of synthetic and real data, we have demonstrated that using this metric as the loss function yields significantly improved results in comparison to the common direct pixel-wise deviation of depth values.
We have combined our metric with relatively simple and non-specific deep learning architectures and expect that this approach will be beneficial for other related problems.

We have focused on the case of single regularly sampled RGBD images, but a lot of geometric data has less regular form.
The future work would be to adapt the developed methodology to a more general sampling of the depth values for the cases of multiple RGBD images or point clouds annotated with a collection of RGB images.

\section*{Acknowledgements}
The work was supported by The Ministry of Education and Science of Russian Federation, grant No. 14.615.21.0004, grant code: RFMEFI61518X0004.

The authors acknowledge the usage of the Skoltech CDISE HPC cluster Zhores for obtaining the results presented in this paper.

    \clearpage\FloatBarrier\appendix
    \section*{Supplementary material}
\section{Additional evaluation details}
\label{sec:sup_evaluation_details}

In the literature on range image processing, the term \emph{depth} is used to denote three different types of range data:
\begin{itemize}
	\item \emph{disparity}, presented in, \eg, Middlebury dataset, \ie, the difference in image location of a feature within two stereo images;
	\item \emph{orthogonal depth}, presented in, \eg, SUN-RGBD dataset, \ie, the distance from a point in the 3D-space to the image plane;
	\item \emph{perspective depth}, presented in, \eg, low-resolution scans in ToFMark dataset, \ie, the distance from a point in the 3D-space to the camera.
\end{itemize}
We use the term \emph{depth map} to denote any data of this kind,
however, in our experiments we evaluated each super-resolution method on the range type that it was designed for.
For evaluation of the disparity processing methods on the datasets that do not provide disparity maps, we calculated virtual disparity images with the baseline of 20 cm.

Here we describe the quality measures that we considered in addition to the ones discussed in the main text.
We recall that we label the metrics that compare the depth values directly with subscript \textquote{d}, and the visually-based metrics with subscript \textquote{v}.

\emph{BadPix} is the fraction of measurements with absolute deviation larger than a certain threshold $\tau$
\begin{equation*}
	\mathrm{BadPix_d}(\tau|\depthmap_1,\depthmap_2) = \frac{1}{N}\big|\big\{ij: \big| \depthmap_{1,ij}- \depthmap_{2,ij} \big| > \tau\big\}\big|,
\end{equation*}
or the fraction of measurements with relative deviation larger than a threshold
\begin{equation*}
	\mathrm{BadPix_d}(\tau{\scriptstyle\%}|\depthmap_1,\depthmap_2) = \frac{1}{N}\big|\big\{ij: \big| \frac{\depthmap_{1,ij}-\depthmap_{2,ij}}{\depthmap_{2,ij}}\big| > \frac{\tau}{100}\big\}\big|,
\end{equation*}
where $\depthmap_1$ and $\depthmap_2$ are the compared depth maps,
$ij$ represents individual pixels,
and $N$ is the number of pixels.
We considered BadPix for depth map comparison with absolute thresholds of 1, 5, and 10 cm and relative thresholds of 1, 5, and 10\%.
We also considered this metric for comparison of depth map renderings with the absolute thresholds of 1, 5, and 10 each divided by 255 (which correspond to deviations by the respective numbers of shades of gray in 8-bit grayscale images).

\emph{Bumpiness}, introduced in~\cite{honauer2015hci} for piece-wise planar regions and generalized in~\cite{honauer2016dataset} for arbitrary smooth surfaces, is a measure of surface smoothness with respect to a reference.
It is calculated as
\begin{multline*}
	\mathrm{Bumpiness_d}(\depthmap_1,\depthmap_2) =\\
	\frac{1}{N} \sum_{ij}\min(0.05,\lVert\mathrm{H}_{\depthmap_1 - \depthmap_2}(i,j)\rVert_\mathrm{F})\cdot 100,
\end{multline*}
where $\lVert\cdot\rVert_\mathrm{F}$ is Frobenius norm and $\mathrm{H}_f(i,j)$ is the Hessian matrix of the function $f$, calculated at point $(i,j)$.
We used the original implementation of this metric.
Since this metric includes some parameter values, presumably, specific for the original evaluation dataset, we converted the depth maps to disparity using the camera intrinsics of this dataset.

We used the implementation of $\mathrm{SSIM}$ from \emph{scikit-image}~\cite{van2014scikit} and the original implementation of $\mathrm{LPIPS}$ from~\cite{Zhang_2018_CVPR}.

In addition to our $\ourmetric$ we considered RMS difference of two rendered images without averaging over the basis renderings, \ie, calculated for a single lighting condition.
We denote this metric as $\rmseonimage$: for a light direction $\lightdir$ and a pair of normal maps $\normal_1,\normal_2$ it is calculated as
\begin{equation*}
	\rmseonimage(\depthmap_1,\depthmap_2) = \sqrt{\frac{1}{N}\sum_{i,j} \| \lightdir\cdot \normal_{1,ij} - \lightdir \cdot \normal_{2,ij} \|_2^2}.
\end{equation*}

\section{Comparison of quality measures}
\label{sec:sup_metrics_evaluation}

\begin{figure*}
	\centerline{\includegraphics[width=\linewidth]{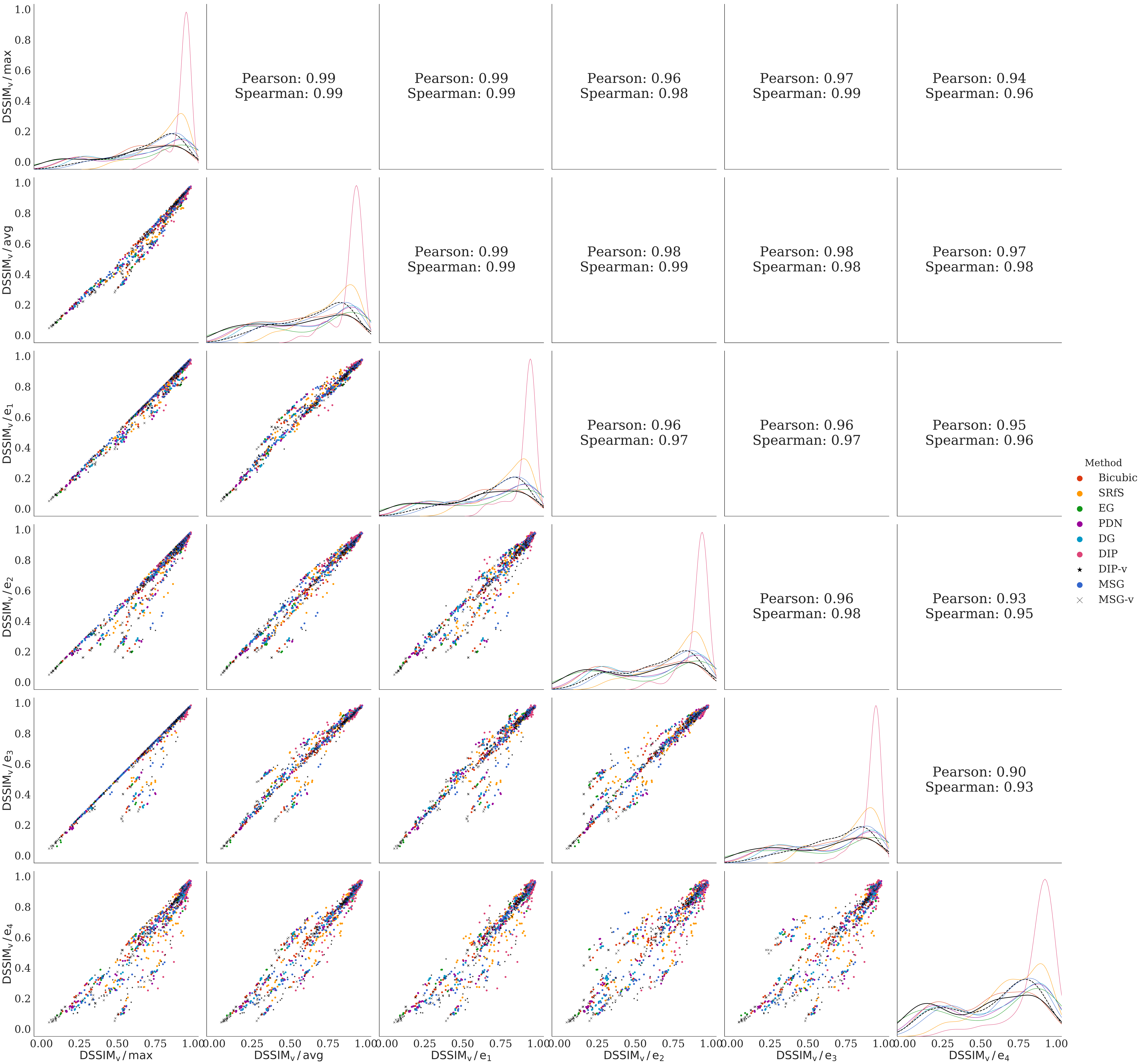}}
	\caption{Comparison of different versions of $\mathrm{DSSIM_v}$ metric.
	Best viewed in large scale and in color.}
	\label{fig:dssim_v}
\end{figure*}

\begin{figure*}
	\centerline{\includegraphics[width=\linewidth]{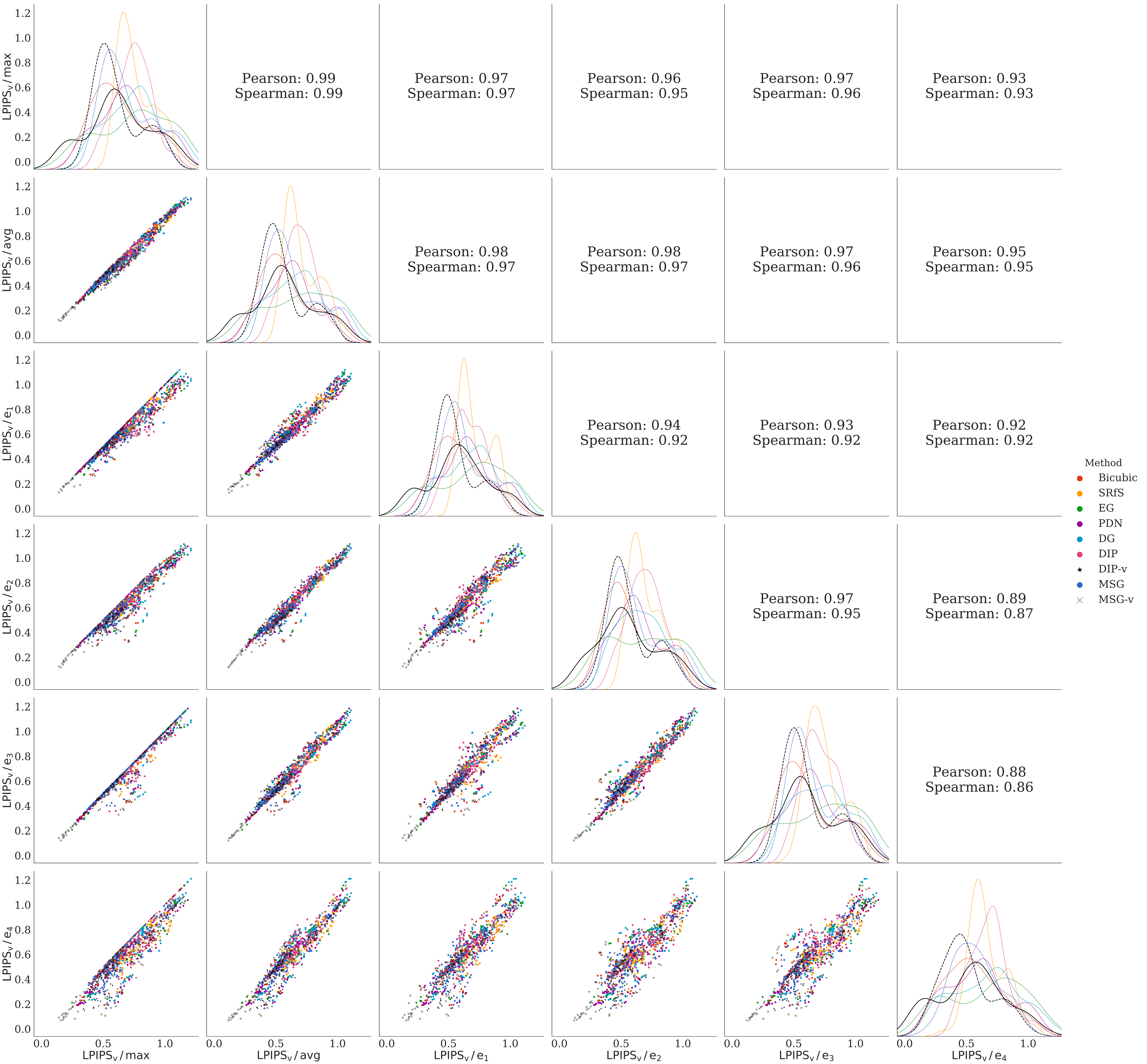}}
	\caption{Comparison of different versions of $\mathrm{LPIPS_v}$ metric.
	Best viewed in large scale and in color.}
	\label{fig:lpips_v}
\end{figure*}

\begin{figure*}
	\centerline{\includegraphics[width=\linewidth]{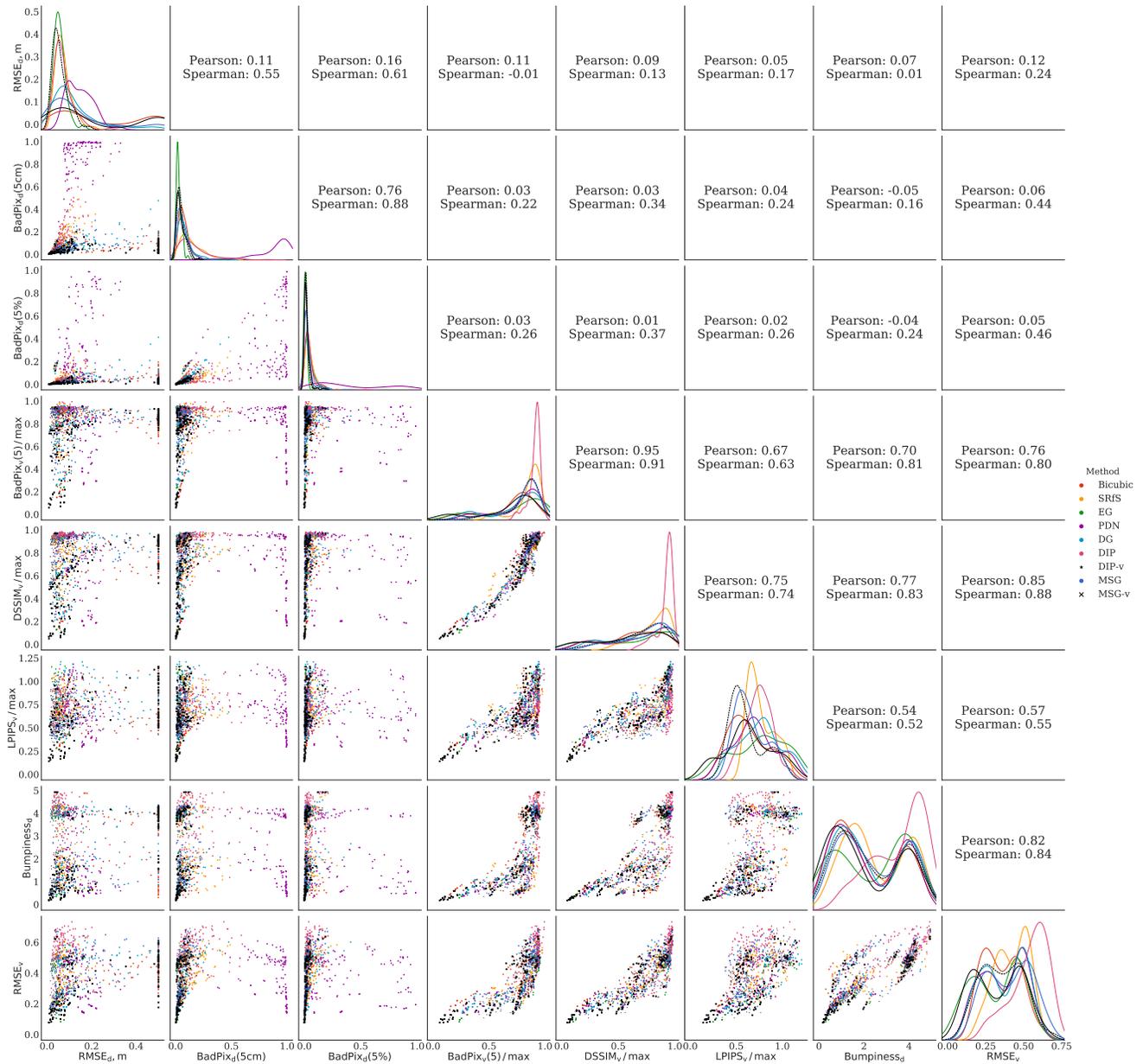}}
	\caption{Comparison of metrics of different types.
	Best viewed in large scale and in color.}
	\label{fig:scatter_different}
\end{figure*}

\begin{figure*}
	\centerline{\includegraphics[width=\linewidth]{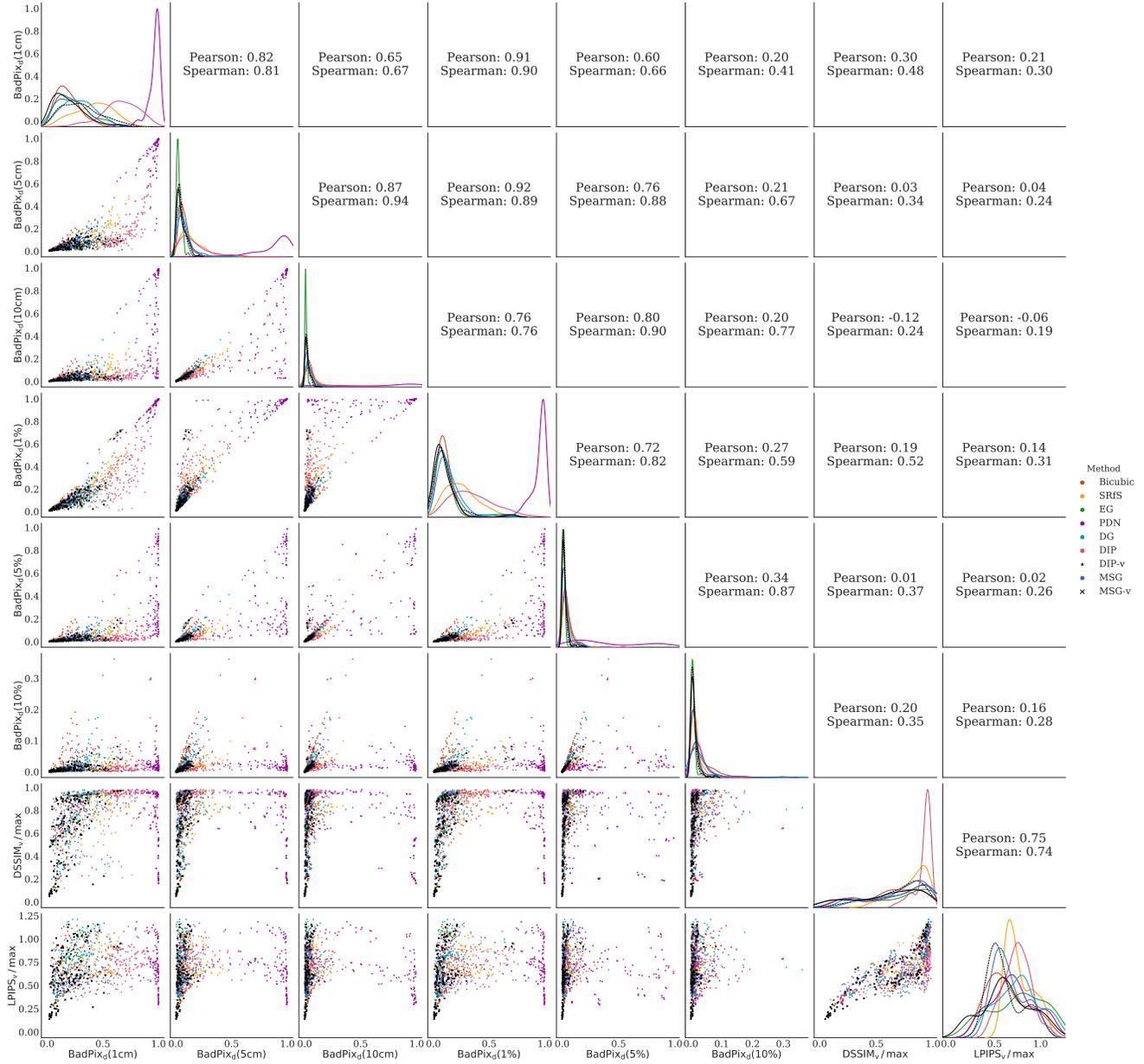}}
	\caption{Comparison of different pixel-wise metrics applied to depth directly and perceptual metrics.
	Best viewed in large scale and in color.}
	\label{fig:pixelwise_d}
\end{figure*}

\begin{figure*}
	\centerline{\includegraphics[width=\linewidth]{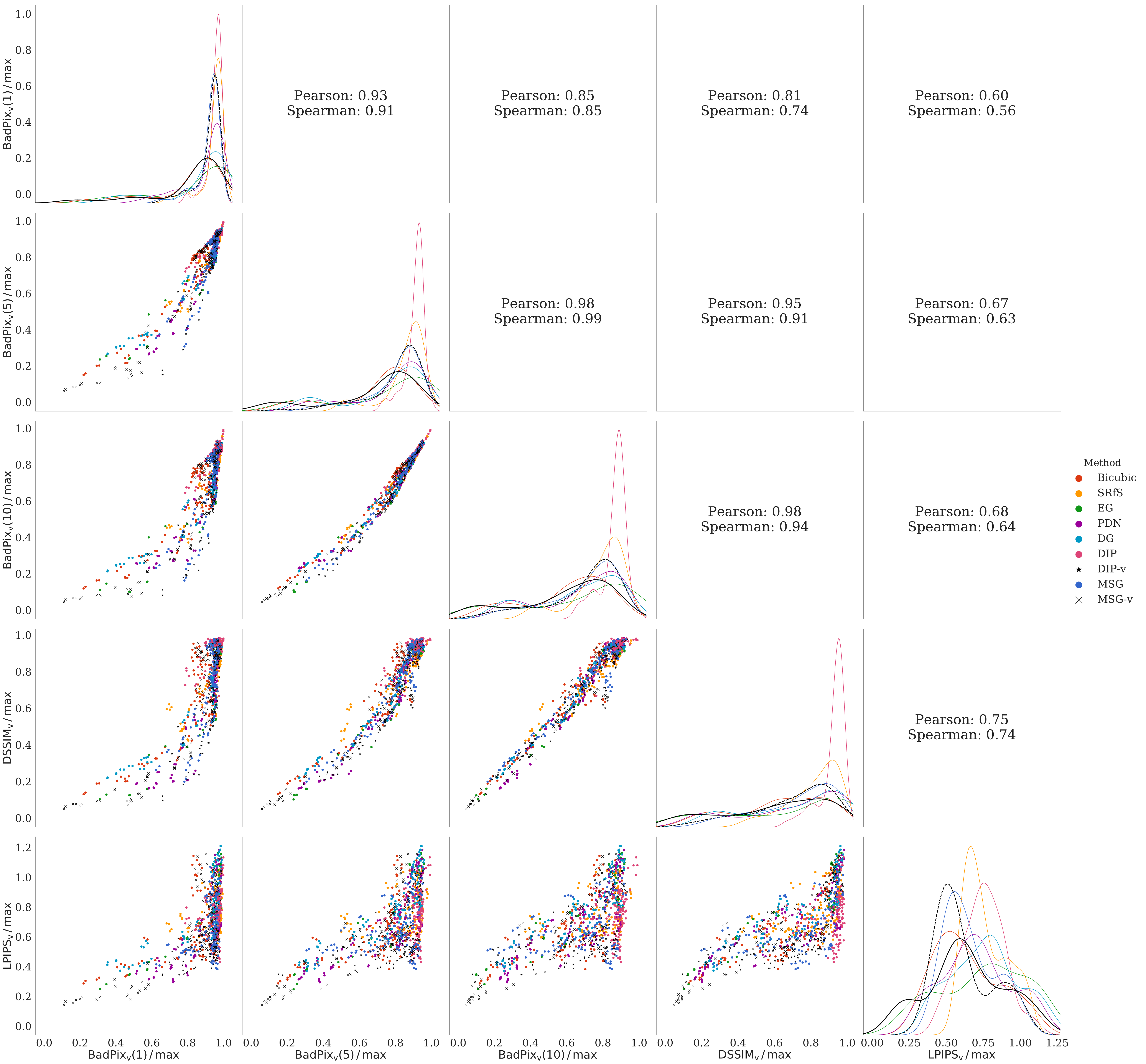}}
	\caption{Comparison of different pixel-wise metrics applied to rendered images and perceptual metrics.
	Best viewed in large scale and in color.}
	\label{fig:pixelwise_v}
\end{figure*}

In {Figures~\ref{fig:rmse_v1}-\ref{fig:pixelwise_v}} we compare the relations between different subsets of quality measures.
We present pair-wise correlations of the metrics in the form of scatter plots in the lower half of the figure and Pearson and Spearman correlation coefficients in the upper half of the figure.
For reference, on the diagonal of the figure we also include kernel density estimates of metric value distributions for each super-resolution method.
The distributions for the modified methods DIP-v and MSG-v are represented with the dashed black and solid black curves respectively.

On the depth maps with missing measurements, the methods that do not inpaint the regions with the missing measurements (including MSG-v) sometimes produced severe outliers around these regions.
To minimize the influence of such outliers on the results of the metric comparison, we limited the value of $\mathrm{RMSE_d}$ to a maximum of 0.5 meters.
Among the collected super-resolved images, 8\% exceeded this threshold.

For each metric, applied to rendered images, we gathered the values of this metric for four different light directions, as described in Section~{5.2} of the main text.
We then calculated two additional values, the worst and the average of these four.
We label the respective versions of the metric with suffixes $\lightdir_1$, $\lightdir_2$, $\lightdir_3$, $\lightdir_4$, $\mathrm{max}$ and $\mathrm{avg}$.
For each metric, we found that these six versions are strongly correlated, as illustrated in Figures~{\ref{fig:rmse_v1}-\ref{fig:lpips_v}}, so we further focused on the worst value of each metric.

We also found that different versions of $\rmseonimage$ produce very similar results to our $\ourmetric$, as illustrated in Figure~\ref{fig:rmse_v1}.
It is consistent with the observation that if $\ourmetric$ is bounded by a constant $C$, then for \emph{any} choice of the light direction $\lightdir$, $\rmseonimage$ is bounded by $C$, which can be easily seen from the fact that $\ourmetric$ does not depend on the choice of the basis, so we can choose one of the basis light directions to be equal to $\lightdir$.

In Figure~\ref{fig:scatter_different} we compare the metrics of different types: pixel-wise $\mathrm{RMSE_d}$, $\mathrm{BadPix_d(5cm)}$ and $\mathrm{BadPix_d(5\%)}$ applied to depth directly;
\textquote{worst} versions of pixel-wise $\mathrm{BadPix_v(5)}$ and perceptual $\mathrm{DSSIM_v}$ and $\mathrm{LPIPS_v}$, applied to rendered images;
geometrical $\mathrm{Bumpiness_d}$ and our $\ourmetric$.
We found that all three pixel-wise metrics applied to depth directly demonstrate weak correlation with visual and geometrical metrics.
Pixel-wise $\mathrm{BadPix_v(5)}$ applied to rendered images, although strongly correlated with perceptual metrics, is inappropriate for gradient-based optimization.
Additional comparison of pixel-wise $\mathrm{BadPix_d}$ and $\mathrm{BadPix_v}$ with different thresholds to perceptual $\mathrm{DSSIM_v}$ and $\mathrm{LPIPS_v}$ ({Figures~\ref{fig:pixelwise_d}~and~\ref{fig:pixelwise_v}}) leads to the same conclusions.
$\mathrm{Bumpiness_d}$ is also strongly correlated with perceptual metrics but only measures local curvature deviation, while the visual appearance of 3D surface is determined by its local orientation.

\section{Comparison of super-resolution methods}
\label{sec:sup_methods_evaluation}
In {Tables~\ref{tab:SimGeo_Cube}-\ref{tab:Middlebury_Complex}} we present the results of quantitative evaluation of super-resolution methods on the datasets SimGeo, ICL-NUIM and Middlebury for Box downsampling model and scaling factors of 4 and 8.
In Table~\ref{tab:average} we present the average values. $\mathrm{RMSE_d}$ is in millimeters, $\mathrm{BadPix}$ is in percents, $\mathrm{DSSIM_v}$, $\mathrm{LPIPS_v}$ and $\ourmetric$ are in thousandths.
For all visual metrics except $\ourmetric$ the presented value is of the \textquote{worst} version.
For all metrics the lower value corresponds to the better result.
The best results are highlighted in bold and the second best results are underlined.

In addition to metric values, the last three columns of the tables contain the results of the informal perceptual study collected over approximately 250 subjects.
In this study, for each scene from SimGeo, ICL-NUIM and Middlebury datasets subjects were shown the renderings of super-resolved depth maps, shuffled randomly, and were asked to choose the renderings, the most and second most similar to ground truth.
The renderings calculated with the fourth light direction were used.
The values in the columns \textquote{User, 1st}, \textquote{User, 2nd}, and \textquote{Top 2} represent the percentages of the subjects who chose the rendering of the super-resolved depth map, produced by the method in the corresponding method, as the most similar, second most similar, or one of the two most similar to the ground truth respectively.
We found that our $\ourmetric$ is mostly consistent with human judgements.

\begin{table*}
\footnotesize
\setlength\tabcolsep{\widthof{0}*\real{.8}}
\setlength\aboverulesep{0pt}
\setlength\belowrulesep{0pt}
\centering
\begin{tabular}{l|cc|cc|cc|cc|cc|cc|cc|c|c|c}
    \toprule
    \multicolumn{1}{c}{} & \multicolumn{17}{c}{Average performance on SimGeo dataset} \\
               \cmidrule{2-18}
    \multicolumn{1}{c}{} & \multicolumn{2}{c}{$\mathrm{RMSE_d}$} & \multicolumn{2}{c}{$\mathrm{BadPix_d(5cm)}$} & \multicolumn{2}{c}{$\mathrm{BadPix_v(5)}$} & \multicolumn{2}{c}{$\mathrm{DSSIM_v}$} & \multicolumn{2}{c}{$\mathrm{LPIPS_v}$} & \multicolumn{2}{c}{$\mathrm{Bumpiness_d}$} & \multicolumn{2}{c}{$\ourmetric$} & \multicolumn{1}{c}{User, 1st} & \multicolumn{1}{c}{User, 2nd} & \multicolumn{1}{c}{Top 2} \\
    \multicolumn{1}{c}{} & x4 & x8 & x4 & x8 & x4 & x8 & x4 & x8 & x4 & x8 & x4 & x8 & x4 & x8 & x4 & x4 & x4 \\
    \midrule
    Bicubic & 55 & 79 & 4.1 & 7.9 & \underline{23.2} & \underline{38.3} & 197 & 301 & 320 & 427 & 0.70 & 0.98 & 193 & 234 & 0.5 & 7.8 & 8.3 \\
SRfS~\cite{haefner2018fight} & 61 & 88 & 7.5 & 14.3 & 74.2 & 77.1 & 711 & 729 & 869 & 865 & 1.48 & 1.69 & 311 & 328 & 0.0 & 0.0 & 0.0 \\
EG~\cite{xie2016edge} & 53 &   & 2.2 &   & 33.1 &   & \underline{168} &   & 306 &   & \underline{0.54} &   & \underline{136} &   & 0.2 & 3.9 & 4.2 \\
PDN~\cite{riegler2016deep} & 162 & 211 & 99.4 & 99.1 & 39.2 & 45.1 & 224 & \underline{264} & \underline{278} & \underline{407} & 0.63 & \underline{0.79} & 165 & 201 & 1.6 & \underline{12.3} & 13.8 \\
DG~\cite{gu2017learning} & 54 & 84 & 3.0 & 6.4 & 35.2 & 39.1 & 293 & 316 & 420 & 437 & 0.69 & 0.82 & 171 & 190 & 0.2 & 2.9 & 3.2 \\
DIP~\cite{Ulyanov_2018_CVPR} & 52 & 59 & 8.5 & 12.5 & 90.5 & 92.0 & 887 & 880 & 893 & 915 & 2.21 & 2.77 & 395 & 475 & 0.6 & 0.9 & 1.5 \\
MSG~\cite{hui2016depth} & \underline{39} & \underline{39} & \underline{1.5} & 3.3 & 51.9 & 69.3 & 374 & 544 & 569 & 713 & 0.79 & 0.97 & 194 & 242 & 0.4 & 3.7 & 4.0 \\
	\midrule
DIP-v & \textbf{33} & 41 & 1.7 & \underline{2.3} & 49.7 & 67.1 & 313 & 491 & 524 & 598 & 0.60 & 0.88 & 147 & \underline{174} & \underline{8.3} & \textbf{59.4} & \underline{67.8} \\
MSG-v & 96 & \textbf{29} & \textbf{0.7} & \textbf{1.5} & \textbf{14.2} & \textbf{34.6} & \textbf{95} & \textbf{206} & \textbf{194} & \textbf{367} & \textbf{0.34} & \textbf{0.46} & \textbf{99} & \textbf{129} & \textbf{88.1} & 9.1 & \textbf{97.2} \\
    \midrule
		\midrule
    \multicolumn{1}{c}{} & \multicolumn{17}{c}{Average performance on ICL-NUIM dataset} \\
    \midrule
    Bicubic & 34 & 54 & 2.8 & 5.5 & \underline{59.3} & \underline{64.2} & \underline{431} & \underline{490} & 558 & 668 & 1.15 & 1.32 & 210 & 252 & 5.0 & \underline{28.3} & 33.3 \\
SRfS~\cite{haefner2018fight} & 42 & 62 & 5.5 & 11.0 & 73.5 & 76.1 & 641 & 664 & 636 & 660 & 1.72 & 1.83 & 287 & 314 & 0.0 & 0.0 & 0.0 \\
PDN~\cite{riegler2016deep} & 135 & 165 & 93.8 & 82.9 & 66.2 & 70.2 & 480 & 509 & 623 & 650 & \underline{1.14} & \underline{1.24} & 237 & 264 & 2.6 & 10.5 & 13.1 \\
DG~\cite{gu2017learning} & 36 & 58 & 4.3 & 6.4 & 64.4 & 65.5 & 497 & 505 & 663 & 689 & 1.28 & 1.32 & 234 & 259 & 0.6 & 5.5 & 6.1 \\
DIP~\cite{Ulyanov_2018_CVPR} & 43 & 56 & 10.6 & 14.2 & 83.6 & 83.4 & 812 & 806 & 690 & 690 & 2.73 & 2.58 & 394 & 389 & 1.1 & 0.9 & 2.0 \\
MSG~\cite{hui2016depth} & \underline{25} & \textbf{36} & \underline{1.6} & \underline{3.5} & 64.1 & 69.0 & 489 & 557 & \underline{510} & \underline{534} & 1.27 & 1.46 & 210 & 255 & 1.1 & 7.2 & 8.3 \\
	\midrule
DIP-v & 28 & \underline{40} & 2.6 & 3.9 & 67.8 & 69.6 & 516 & 548 & \textbf{407} & \textbf{503} & 1.45 & 1.56 & \underline{209} & \underline{236} & \underline{9.6} & \textbf{31.9} & \underline{41.4} \\
MSG-v & \textbf{24} & 41 & \textbf{1.3} & \textbf{3.1} & \textbf{56.3} & \textbf{61.1} & \textbf{387} & \textbf{437} & 527 & 602 & \textbf{0.94} & \textbf{1.06} & \textbf{157} & \textbf{192} & \textbf{79.9} & 11.8 & \textbf{91.7} \\
		\midrule
		\midrule
    \multicolumn{1}{c}{} & \multicolumn{17}{c}{Average performance on Middlebury dataset} \\
    \midrule
    Bicubic & 843 & 1139 & 10.8 & 13.9 & \textbf{71.5} & \textbf{76.7} & \textbf{648} & \textbf{748} & \underline{575} & 720 & \textbf{0.87} & \textbf{0.76} & \textbf{344} & \textbf{386} & 4.1 & \underline{25.3} & 29.4 \\
SRfS~\cite{haefner2018fight} & 100 & 145 & 21.4 & 33.6 & 86.4 & 89.5 & 780 & 810 & 669 & 704 & 1.32 & \underline{1.28} & 428 & 461 & 0.0 & 0.0 & 0.0 \\
PDN~\cite{riegler2016deep} & 173 & 225 & 85.3 & 76.5 & 83.4 & 86.4 & 744 & 790 & 653 & 711 & 1.38 & 1.67 & 405 & 467 & 9.9 & \textbf{28.1} & \underline{37.9} \\
DG~\cite{gu2017learning} & 266 & 330 & 15.0 & 24.5 & 81.8 & 84.1 & 765 & 784 & 728 & 740 & 1.54 & 1.73 & 421 & 442 & 0.7 & 10.6 & 11.3 \\
DIP~\cite{Ulyanov_2018_CVPR} & \underline{72} & \underline{104} & 19.6 & 24.4 & 92.4 & 93.4 & 927 & 947 & 737 & 717 & 2.82 & 2.90 & 565 & 592 & 1.2 & 5.6 & 6.8 \\
MSG~\cite{hui2016depth} & 228 & 426 & 10.8 & 13.1 & 81.8 & 87.2 & 774 & 858 & 649 & 696 & 1.96 & 2.19 & 477 & 525 & 0.2 & 1.6 & 1.8 \\
	\midrule
DIP-v & \textbf{56} & \textbf{87} & \textbf{6.4} & \underline{10.6} & 83.1 & 87.4 & 728 & 821 & \textbf{506} & \textbf{568} & 1.34 & 1.56 & \underline{353} & \underline{409} & \textbf{72.3} & 18.2 & \textbf{90.5} \\
MSG-v & 96 & 133 & \underline{7.3} & \textbf{9.2} & \underline{73.3} & \underline{79.0} & \underline{667} & \underline{757} & 639 & \underline{690} & \underline{1.20} & 1.35 & 376 & 431 & \underline{10.8} & 9.9 & 20.7 \\
    \midrule
    \midrule
		\multicolumn{1}{c}{} & \multicolumn{17}{c}{Average performance on the scenes without missing measurements (SimGeo, ICL-NUIM, Vintage)} \\
               \cmidrule{2-18}
    \multicolumn{1}{c}{} & \multicolumn{2}{c}{$\mathrm{RMSE_d}$} & \multicolumn{2}{c}{$\mathrm{BadPix_d(5cm)}$} & \multicolumn{2}{c}{$\mathrm{BadPix_v(5)}$} & \multicolumn{2}{c}{$\mathrm{DSSIM_v}$} & \multicolumn{2}{c}{$\mathrm{LPIPS_v}$} & \multicolumn{2}{c}{$\mathrm{Bumpiness_d}$} & \multicolumn{2}{c}{$\ourmetric$} & \multicolumn{1}{c}{User, 1st} & \multicolumn{1}{c}{User, 2nd} & \multicolumn{1}{c}{Top 2} \\
    \multicolumn{1}{c}{} & x4 & x8 & x4 & x8 & x4 & x8 & x4 & x8 & x4 & x8 & x4 & x8 & x4 & x8 & x4 & x4 & x4 \\
    \midrule
    Bicubic & 46 & 69 & 3.5 & 6.9 & \underline{43.7} & \underline{53.3} & \underline{333} & 415 & \underline{452} & 561 & 0.97 & 1.19 & 206 & 248 & 3.0 & \underline{18.9} & 21.9 \\
SRfS~\cite{haefner2018fight} & 55 & 81 & 7.3 & 14.1 & 74.6 & 77.4 & 680 & 701 & 743 & 753 & 1.60 & 1.75 & 303 & 326 & 0.0 & 0.0 & 0.0 \\
PDN~\cite{riegler2016deep} & 148 & 187 & 94.4 & 90.1 & 55.0 & 59.8 & 378 & \underline{412} & 482 & \underline{542} & \underline{0.93} & \underline{1.06} & 210 & 241 & 1.9 & 10.5 & 12.4 \\
DG~\cite{gu2017learning} & 47 & 73 & 3.9 & 6.7 & 52.1 & 54.4 & 416 & 430 & 561 & 584 & 1.02 & 1.10 & 209 & 230 & 0.4 & 4.0 & 4.4 \\
DIP~\cite{Ulyanov_2018_CVPR} & 50 & 62 & 10.7 & 15.9 & 87.6 & 88.2 & 857 & 853 & 801 & 808 & 2.59 & 2.79 & 414 & 452 & 0.8 & 0.8 & 1.7 \\
MSG~\cite{hui2016depth} & \underline{33} & \underline{39} & \underline{1.7} & 3.6 & 59.8 & 70.3 & 454 & 569 & 547 & 622 & 1.08 & 1.26 & 210 & 258 & 0.7 & 5.8 & 6.4 \\
	\midrule
DIP-v & \textbf{31} & 43 & 2.2 & \underline{3.3} & 60.8 & 69.9 & 444 & 548 & 474 & 560 & 1.09 & 1.32 & \underline{191} & \underline{223} & \underline{10.2} & \textbf{45.5} & \underline{55.8} \\
MSG-v & 38 & \textbf{38} & \textbf{1.1} & \textbf{2.6} & \textbf{38.0} & \textbf{50.1} & \textbf{264} & \textbf{346} & \textbf{385} & \textbf{501} & \textbf{0.69} & \textbf{0.81} & \textbf{135} & \textbf{169} & \textbf{82.7} & 10.9 & \textbf{93.6} \\
    \midrule
    \midrule
		\multicolumn{1}{c}{} & \multicolumn{17}{c}{Average performance on the scenes with missing measurements (Middlebury excluding Vintage)} \\
    \midrule
    Bicubic & 972 & 1313 & 11.8 & 14.7 & \textbf{71.3} & \textbf{76.6} & \textbf{663} & \textbf{765} & \underline{570} & 718 & \textbf{0.77} & \textbf{0.61} & \underline{358} & \textbf{400} & 3.8 & \underline{24.7} & 28.5 \\
SRfS~\cite{haefner2018fight} & 100 & 145 & 22.2 & 33.8 & 86.9 & 89.8 & 790 & 820 & 676 & 716 & 1.26 & \underline{1.21} & 441 & 474 & 0.0 & 0.0 & 0.0 \\
PDN~\cite{riegler2016deep} & 178 & 234 & 88.3 & 76.1 & 83.5 & 86.6 & 757 & 803 & 644 & 713 & 1.36 & 1.69 & 419 & 487 & \underline{11.5} & \textbf{32.8} & \underline{44.3} \\
DG~\cite{gu2017learning} & 298 & 367 & 16.3 & 26.9 & 82.2 & 84.7 & 781 & 803 & 716 & 724 & 1.55 & 1.76 & 442 & 465 & 0.9 & 12.2 & 13.1 \\
DIP~\cite{Ulyanov_2018_CVPR} & \underline{72} & \underline{102} & 18.8 & 20.7 & 92.2 & 93.3 & 923 & 943 & 708 & 691 & 2.62 & 2.68 & 549 & 577 & 1.2 & 6.4 & 7.7 \\
MSG~\cite{hui2016depth} & 259 & 488 & 12.1 & 14.1 & 82.0 & 87.7 & 785 & 870 & 673 & 711 & 2.02 & 2.24 & 507 & 552 & 0.2 & 0.2 & 0.5 \\
	\midrule
DIP-v & \textbf{58} & \textbf{91} & \textbf{7.1} & \underline{11.4} & 82.7 & 87.2 & 716 & 811 & \textbf{494} & \textbf{550} & 1.23 & 1.41 & \textbf{354} & \underline{405} & \textbf{80.1} & 13.8 & \textbf{93.9} \\
MSG-v & 107 & 145 & \underline{8.1} & \textbf{9.8} & \underline{73.6} & \underline{79.2} & \underline{688} & \underline{776} & 634 & \underline{688} & \underline{1.19} & 1.34 & 404 & 458 & 1.3 & 8.8 & 10.1 \\
    \midrule
    \midrule
    \multicolumn{1}{c}{} & \multicolumn{17}{c}{Average performance on SimGeo, ICL-NUIM, Middlebury} \\
    \midrule
    Bicubic & 339 & 462 & 6.1 & 9.3 & \underline{52.4} & \underline{60.6} & \underline{437} & \underline{525} & 489 & 611 & \underline{0.91} & \underline{1.00} & 254 & 296 & 3.3 & \underline{20.7} & 24.0 \\
SRfS~\cite{haefner2018fight} & 69 & 101 & 12.0 & 20.3 & 78.5 & 81.3 & 715 & 738 & 722 & 741 & 1.50 & 1.58 & 347 & 372 & 0.0 & 0.0 & 0.0 \\
PDN~\cite{riegler2016deep} & 157 & 202 & 92.4 & 85.7 & 64.0 & 68.2 & 498 & 536 & 533 & 596 & 1.07 & 1.26 & 276 & 319 & 5.0 & 17.5 & 22.5 \\
DG~\cite{gu2017learning} & 126 & 166 & 7.8 & 13.1 & 61.6 & 64.0 & 531 & 548 & 610 & 628 & 1.19 & 1.31 & 283 & 305 & 0.5 & 6.6 & 7.1 \\
DIP~\cite{Ulyanov_2018_CVPR} & \underline{57} & 75 & 13.3 & 17.4 & 89.1 & 89.8 & 878 & 881 & 771 & 771 & 2.60 & 2.76 & 457 & 491 & 1.0 & 2.6 & 3.6 \\
MSG~\cite{hui2016depth} & 104 & 181 & 5.0 & 6.9 & 66.8 & 75.8 & 559 & 664 & 587 & 650 & 1.37 & 1.57 & 304 & 350 & 0.5 & 4.0 & 4.6 \\
	\midrule
DIP-v & \textbf{40} & \textbf{58} & \underline{3.8} & \underline{5.9} & 67.7 & 75.4 & 530 & 631 & \underline{481} & \textbf{557} & 1.14 & 1.34 & \underline{242} & \underline{280} & \underline{32.3} & \textbf{35.5} & \textbf{67.8} \\
MSG-v & 60 & \underline{72} & \textbf{3.3} & \textbf{4.8} & \textbf{49.2} & \textbf{59.3} & \textbf{398} & \textbf{482} & \textbf{464} & \underline{560} & \textbf{0.85} & \textbf{0.98} & \textbf{220} & \textbf{260} & \textbf{57.0} & 10.2 & \underline{67.3} \\
		\bottomrule
\end{tabular}
\caption{Quantitative evaluation summary.
The best result is in bold, the second best in underlined.}
\label{tab:average}
\end{table*}

\section{Training with $\ourmetricinloss$}
\label{sec:training}
Since optimization of $\ourmetricinloss$ alone is an ill-posed problem, we used a regularization term that penalizes absolute depth deviation.
We found that among different regularizers, including $\mathrm{MSE_d}$, $\mathrm{Lap_1}$ produces the best results.
In general, we found that optimization leads to the best results if the terms are weighted in such way that geometrically corresponding depth error and angular normal error result in the same magnitudes of terms.
The corresponding value of the weighing parameter $w$ in Equation~4 of the main text is determined by the properties of the training data, such as depth map scaling or field of view of the camera.

\section{Noisy depth measurements in the input}
\label{sec:noisy_data}
\begin{figure}
	\centerline{\includegraphics[width=\linewidth]{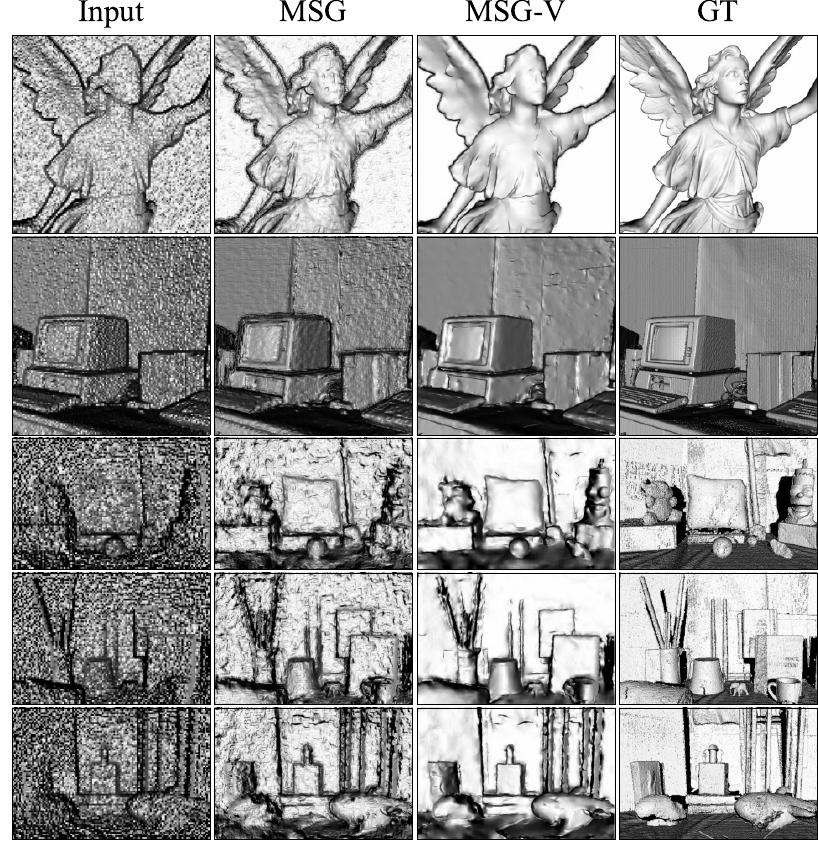}}
	\caption{$\times 4$ super-resolution results produced by the original MSG and MSG-V with our loss, both trained on noisy data.
	The upper two samples contain synthetic noise, while the lower three from ToFMark dataset represent real noisy ToF measurements.
	Best viewed in large scale.}
	\label{fig:noisy}
\end{figure}

\begin{figure*}[p!]
    \centerline{\includegraphics[width=\linewidth]{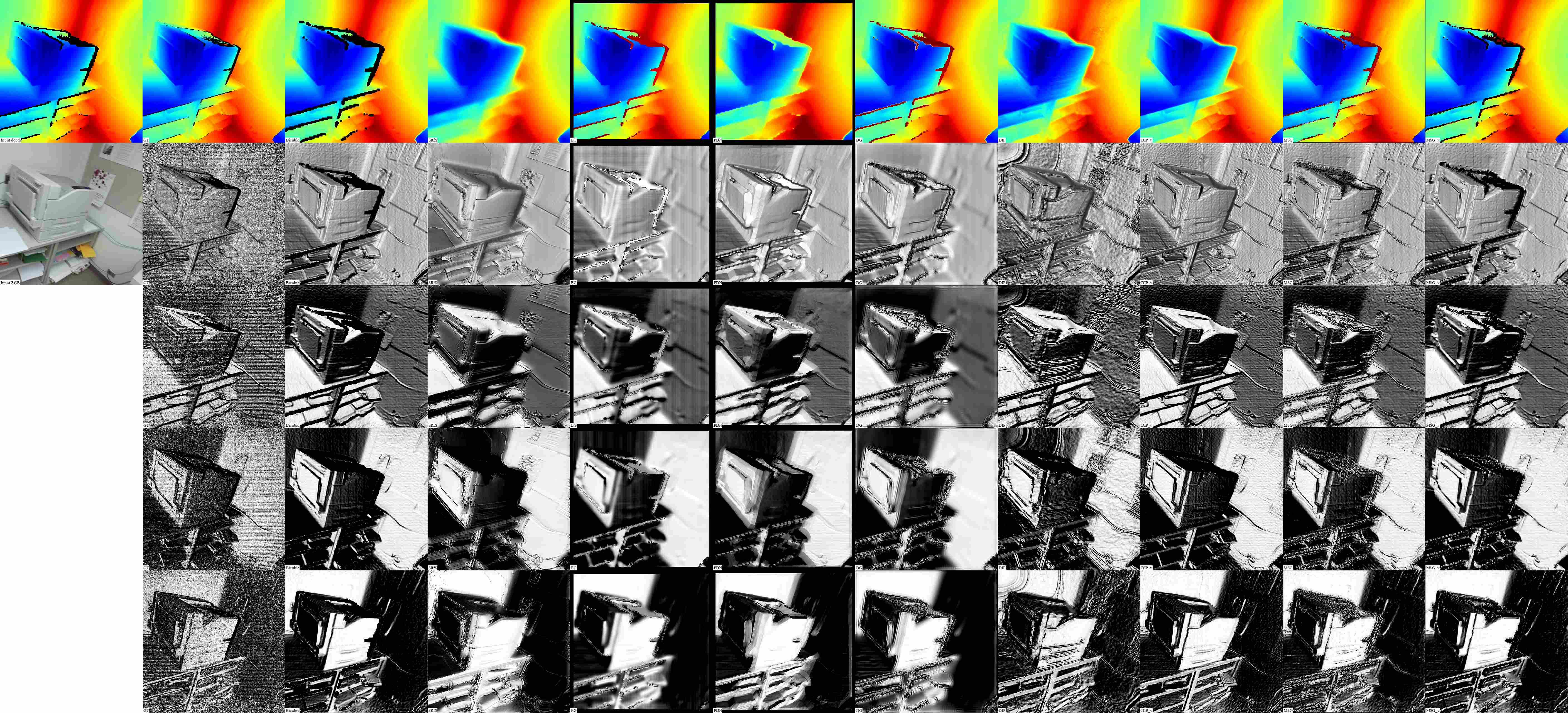}}
		\caption{x4 super-resolution results on a Kinect v2 RGBD scan from SUN RGBD dataset.
		Each visualization is labeled in the bottom left corner.
		Ground truth is in the 2nd column, DIP-v is in the third from the right, MSG-v in the last one.}
    \label{fig:kinnect}
\end{figure*}

\begin{figure*}[p!]
    \centerline{\includegraphics[width=\linewidth]{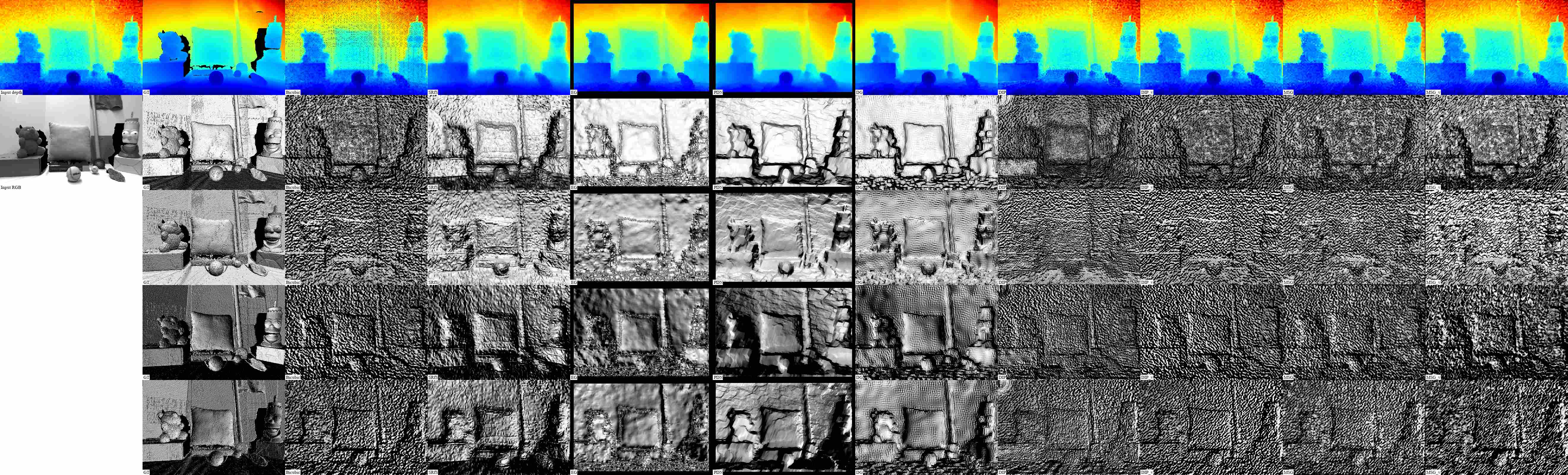}}
		\caption{x4 super-resolution results on \textquote{Devil} from ToFMark dataset.
		Each visualization is labeled in the bottom left corner.
		Ground truth is in the 2nd column, DIP-v is in the third from the right, MSG-v in the last one.}
    \label{fig:tof}
\end{figure*}

SimGeo, ICL-NUIM and Middlebury datasets were our primary evaluation sets, yielding the most pronounced outcomes,
however, these datasets contain only noise-free scenes.
As we were interested in evaluation of our approach on a diverse set of RGBD images, we included twelve scenes from SUN RGBD dataset and three scenes from ToFMark dataset that feature real-world noise patterns in our evaluation data.
We observed that increased levels of noise are extremely harmful to all non over-smoothing methods, including those modified with our loss, as they fail to produce reasonable super-resolution results, as illustrated in Figures~\ref{fig:kinnect}-\ref{fig:tof}.
To demonstrate that this is not a limitation of our approach, in Figure~\ref{fig:noisy} we present the super-resolution results produced by modified and unmodified versions of MSG, trained on the data with synthetic multiplicative gaussian noise.

\section{Different downsampling models}
\label{sec:downsampling_models}
\begin{figure}
	\centerline{\includegraphics[width=\linewidth]{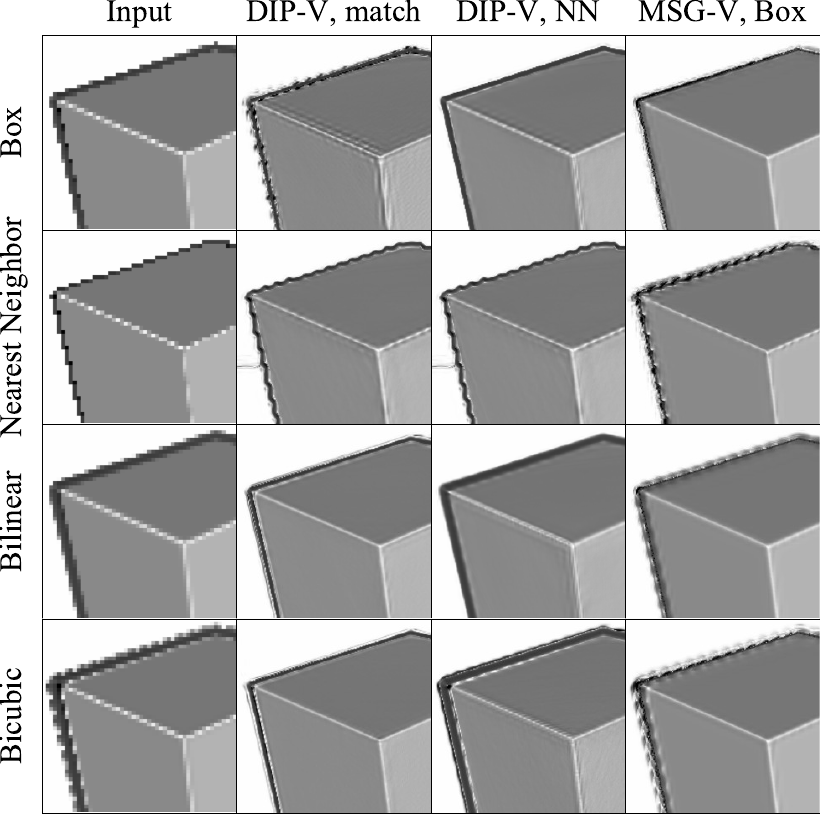}}
	\caption{$\times 4$ super-resolution results for different input downsampling models produced by DIP-V with a matching downsampling model, DIP-V with Nearest Neighbor downsampling model and MSG-V with Box downsampling model.
	Best viewed in large scale.}
	\label{fig:downsampling}
\end{figure}

In Figure~\ref{fig:downsampling} we present the results for different downsampling models, used for calculation of low-resolution input.
We found that the visual quality remains high when the downsampling model used during training and that of the input match;
if this is not the case, the quality deteriorates, as expected.


    \begin{figure*}
	\centerline{\includegraphics[width=\linewidth]{scatter_plots/RMSE_v1}}
	\caption{Comparison of different versions of $\rmseonimage$ metric and $\ourmetric$ metric.
	Best viewed in large scale and in color.}
	\label{fig:rmse_v1}
\end{figure*}


\begingroup
\FloatBarrier
\setlength\tabcolsep{\widthof{0}*\real{.8}}
\setlength\aboverulesep{0pt}
\setlength\belowrulesep{0pt}

\begin{table*}
\footnotesize
\centering
\begin{tabular}{l|cc|cc|cc|cc|cc|cc|cc|c|c|c}
    \toprule
		\multicolumn{1}{c}{} & \multicolumn{17}{c}{Cube, high-frequency texture} \\
               \cmidrule{2-18}
    \multicolumn{1}{c}{} & \multicolumn{2}{c}{$\mathrm{RMSE_d}$} & \multicolumn{2}{c}{$\mathrm{BadPix_d(5cm)}$} & \multicolumn{2}{c}{$\mathrm{BadPix_v(5)}$} & \multicolumn{2}{c}{$\mathrm{DSSIM_v}$} & \multicolumn{2}{c}{$\mathrm{LPIPS_v}$} & \multicolumn{2}{c}{$\mathrm{Bumpiness_d}$} & \multicolumn{2}{c}{$\ourmetric$} & \multicolumn{1}{c}{User, 1st} & \multicolumn{1}{c}{User, 2nd} & \multicolumn{1}{c}{Top 2} \\
    \multicolumn{1}{c}{} & x4 & x8 & x4 & x8 & x4 & x8 & x4 & x8 & x4 & x8 & x4 & x8 & x4 & x8 & x4 & x4 & x4 \\
    \midrule
    Bicubic & 44 & 63 & 2.7 & 5.2 & \underline{15.0} & \textbf{27.3} & 131 & \underline{204} & 287 & \underline{395} & 0.43 & 0.61 & 160 & 188 & 0.7 & \underline{13.2} & 14.0 \\
SRfS~\cite{haefner2018fight} & 52 & 75 & 6.2 & 12.1 & 89.2 & 80.3 & 934 & 818 & 1036 & 938 & 1.73 & 1.67 & 361 & 339 & 0.0 & 0.0 & 0.0 \\
EG~\cite{xie2016edge} & 43 &   & 1.2 &   & 25.4 &   & \underline{113} &   & \underline{214} &   & \underline{0.35} &   & \underline{105} &   & 0.7 & 9.6 & 10.3 \\
PDN~\cite{riegler2016deep} & 164 & 219 & 99.6 & 99.4 & 27.5 & \underline{29.5} & 156 & \textbf{186} & 250 & \textbf{368} & 0.39 & \underline{0.49} & 145 & 171 & 0.7 & 4.4 & 5.1 \\
DG~\cite{gu2017learning} & 44 & 67 & 1.9 & 4.2 & 26.4 & 30.1 & 218 & 240 & 411 & 437 & 0.44 & 0.55 & 139 & \underline{159} & 0.7 & 7.4 & 8.1 \\
DIP~\cite{Ulyanov_2018_CVPR} & 45 & 48 & 6.4 & 8.5 & 93.5 & 92.5 & 963 & 947 & 906 & 918 & 2.98 & 2.50 & 530 & 494 & 0.0 & 0.7 & 0.7 \\
MSG~\cite{hui2016depth} & \underline{29} & 38 & 1.0 & 2.6 & 60.1 & 77.9 & 445 & 653 & 687 & 877 & 0.79 & 0.98 & 176 & 233 & 0.0 & 0.0 & 0.0 \\
	\midrule
DIP-v & \textbf{26} & \underline{36} & \underline{0.8} & \underline{1.6} & 56.2 & 60.8 & 352 & 413 & 613 & 653 & 0.64 & 0.89 & 146 & 162 & \underline{5.9} & \textbf{58.1} & \underline{64.0} \\
MSG-v & 102 & \textbf{20} & \textbf{0.3} & \textbf{0.7} & \textbf{9.3} & 51.0 & \textbf{70} & 316 & \textbf{179} & 676 & \textbf{0.20} & \textbf{0.39} & \textbf{77} & \textbf{125} & \textbf{91.2} & 6.6 & \textbf{97.8} \\
    \midrule
    \midrule
    \multicolumn{1}{c}{} & \multicolumn{17}{c}{Cube, no texture} \\
    \midrule
    Bicubic & 44 & 63 & 2.7 & 5.2 & \underline{15.0} & \underline{27.3} & 131 & 204 & 287 & 395 & 0.43 & 0.61 & 160 & 188 & 0.0 & 5.9 & 5.9 \\
SRfS~\cite{haefner2018fight} & 43 & 63 & 2.1 & 4.5 & 53.4 & 51.7 & 516 & 476 & 754 & 728 & 0.67 & 0.89 & 219 & 228 & 0.0 & 0.0 & 0.0 \\
EG~\cite{xie2016edge} & 43 &   & 1.2 &   & 25.4 &   & 128 &   & \underline{282} &   & 0.35 &   & \underline{105} &   & 0.0 & 2.9 & 2.9 \\
PDN~\cite{riegler2016deep} & 164 & 219 & 99.6 & 99.4 & 26.3 & 29.3 & 162 & \underline{185} & 314 & \underline{353} & 0.38 & 0.49 & 145 & 171 & 0.7 & 4.4 & 5.1 \\
DG~\cite{gu2017learning} & 44 & 67 & 1.9 & 4.2 & 26.4 & 30.1 & 218 & 240 & 411 & 437 & 0.44 & 0.55 & 139 & 159 & 0.0 & 2.2 & 2.2 \\
DIP~\cite{Ulyanov_2018_CVPR} & 72 & 56 & 23.2 & 17.3 & 94.3 & 99.1 & 912 & 980 & 1026 & 1133 & 2.05 & 4.22 & 434 & 683 & 0.0 & 0.7 & 0.7 \\
MSG~\cite{hui2016depth} & 29 & \underline{26} & 1.0 & 1.7 & 30.7 & 49.8 & 199 & 314 & 509 & 642 & 0.42 & 0.47 & 157 & 171 & 0.0 & \underline{6.6} & 6.6 \\
	\midrule
DIP-v & \underline{26} & 35 & \underline{0.8} & \underline{1.4} & 15.1 & 45.4 & \underline{95} & 237 & 347 & 478 & \underline{0.28} & \underline{0.35} & 111 & \underline{107} & \underline{1.5} & \textbf{76.5} & \underline{77.9} \\
MSG-v & \textbf{9} & \textbf{19} & \textbf{0.3} & \textbf{0.4} & \textbf{6.0} & \textbf{13.0} & \textbf{50} & \textbf{73} & \textbf{141} & \textbf{213} & \textbf{0.17} & \textbf{0.21} & \textbf{77} & \textbf{82} & \textbf{97.8} & 0.7 & \textbf{98.5} \\
    \bottomrule
\end{tabular}
\caption{Quantitative evaluation on \textquote{Cube} with different RGBs from SimGeo dataset.
The best result is in bold, the second best is underlined.}
\label{tab:SimGeo_Cube}
\end{table*}

\begin{table*}
\footnotesize
\centering
\begin{tabular}{l|cc|cc|cc|cc|cc|cc|cc|c|c|c}
    \toprule
		\multicolumn{1}{c}{} & \multicolumn{17}{c}{Sphere and cylinder, high-frequency texture} \\
               \cmidrule{2-18}
    \multicolumn{1}{c}{} & \multicolumn{2}{c}{$\mathrm{RMSE_d}$} & \multicolumn{2}{c}{$\mathrm{BadPix_d(5cm)}$} & \multicolumn{2}{c}{$\mathrm{BadPix_v(5)}$} & \multicolumn{2}{c}{$\mathrm{DSSIM_v}$} & \multicolumn{2}{c}{$\mathrm{LPIPS_v}$} & \multicolumn{2}{c}{$\mathrm{Bumpiness_d}$} & \multicolumn{2}{c}{$\ourmetric$} & \multicolumn{1}{c}{User, 1st} & \multicolumn{1}{c}{User, 2nd} & \multicolumn{1}{c}{Top 2} \\
    \multicolumn{1}{c}{} & x4 & x8 & x4 & x8 & x4 & x8 & x4 & x8 & x4 & x8 & x4 & x8 & x4 & x8 & x4 & x4 & x4 \\
    \midrule
    Bicubic & 57 & 82 & 4.1 & 8.1 & \underline{20.1} & \underline{36.7} & 189 & 294 & 313 & \underline{420} & 0.67 & 0.98 & 189 & 234 & 0.0 & 0.7 & 0.7 \\
SRfS~\cite{haefner2018fight} & 70 & 102 & 12.1 & 24.6 & 91.9 & 91.8 & 887 & 865 & 1025 & 1008 & 2.43 & 2.41 & 417 & 403 & 0.0 & 0.0 & 0.0 \\
EG~\cite{xie2016edge} & 55 &   & 2.4 &   & 30.4 &   & \underline{143} &   & 326 &   & \underline{0.50} &   & \underline{130} &   & 0.0 & 1.5 & 1.5 \\
PDN~\cite{riegler2016deep} & 157 & 197 & 99.3 & 98.9 & 40.7 & 54.1 & 198 & \textbf{242} & \underline{295} & 461 & 0.60 & \underline{0.77} & 150 & 187 & 1.5 & 9.6 & 11.0 \\
DG~\cite{gu2017learning} & 56 & 87 & 3.2 & 6.3 & 30.9 & \textbf{35.2} & 265 & \underline{285} & 372 & \textbf{386} & 0.66 & \underline{0.77} & 166 & \underline{180} & 0.0 & 1.5 & 1.5 \\
DIP~\cite{Ulyanov_2018_CVPR} & 46 & 69 & 3.9 & 27.2 & 97.0 & 99.2 & 965 & 975 & 1062 & 1014 & 4.01 & 4.80 & 548 & 696 & 1.5 & 2.9 & 4.4 \\
MSG~\cite{hui2016depth} & \underline{41} & \underline{41} & 1.4 & 3.6 & 72.6 & 85.6 & 626 & 820 & 859 & 960 & 0.98 & 1.43 & 229 & 314 & 0.0 & 0.0 & 0.0 \\
	\midrule
DIP-v & \textbf{28} & 43 & \underline{1.2} & \underline{2.4} & 69.2 & 86.0 & 560 & 850 & 766 & 832 & 0.56 & 1.45 & 142 & 242 & \underline{32.4} & \textbf{52.9} & \underline{85.3} \\
MSG-v & 99 & \textbf{37} & \textbf{0.6} & \textbf{2.0} & \textbf{14.3} & 53.0 & \textbf{94} & 334 & \textbf{267} & 583 & \textbf{0.29} & \textbf{0.55} & \textbf{96} & \textbf{164} & \textbf{64.7} & \underline{30.9} & \textbf{95.6} \\
    \midrule
    \midrule
    \multicolumn{1}{c}{} & \multicolumn{17}{c}{Sphere and cylinder, no texture} \\
    \midrule
    Bicubic & 57 & 82 & 4.1 & 8.1 & \underline{20.2} & 36.8 & 189 & 294 & 325 & 437 & 0.67 & 0.98 & 190 & 233 & 0.0 & 0.7 & 0.7 \\
SRfS~\cite{haefner2018fight} & 59 & 85 & 4.6 & 8.6 & 51.4 & 70.8 & 430 & 619 & 657 & 766 & 0.77 & 1.25 & 193 & 256 & 0.0 & 0.0 & 0.0 \\
EG~\cite{xie2016edge} & 56 &   & 2.4 &   & 30.9 &   & \underline{160} &   & 383 &   & 0.50 &   & \underline{128} &   & 0.0 & 1.5 & 1.5 \\
PDN~\cite{riegler2016deep} & 157 & 197 & 99.3 & 98.9 & 38.0 & 44.1 & 202 & \underline{218} & \underline{294} & \underline{386} & 0.58 & 0.76 & 150 & 186 & 5.9 & \underline{17.6} & 23.5 \\
DG~\cite{gu2017learning} & 57 & 87 & 3.2 & 6.4 & 31.0 & \underline{35.3} & 265 & 284 & 396 & 409 & 0.66 & 0.78 & 165 & 180 & 0.7 & 2.2 & 2.9 \\
DIP~\cite{Ulyanov_2018_CVPR} & 49 & 56 & 5.0 & 5.5 & 85.6 & 81.6 & 856 & 662 & 927 & 723 & 1.01 & 0.96 & 244 & 249 & 1.5 & 0.0 & 1.5 \\
MSG~\cite{hui2016depth} & 40 & \underline{37} & \underline{1.4} & 3.1 & 45.6 & 64.5 & 288 & 444 & 509 & 610 & 0.65 & 0.76 & 183 & 218 & 0.7 & 0.7 & 1.5 \\
	\midrule
DIP-v & \underline{35} & 39 & \underline{1.4} & \underline{1.8} & 41.0 & 72.6 & 210 & 523 & 517 & 643 & \underline{0.47} & \underline{0.70} & 130 & \underline{141} & \underline{9.6} & \textbf{64.7} & \underline{74.3} \\
MSG-v & \textbf{14} & \textbf{27} & \textbf{0.7} & \textbf{1.3} & \textbf{8.5} & \textbf{18.0} & \textbf{77} & \textbf{93} & \textbf{174} & \textbf{200} & \textbf{0.27} & \textbf{0.32} & \textbf{96} & \textbf{110} & \textbf{81.6} & 12.5 & \textbf{94.1} \\
    \midrule
    \midrule
		\multicolumn{1}{c}{} & \multicolumn{17}{c}{Sphere and cylinder, low-frequency texture} \\
    \midrule
    Bicubic & 57 & 82 & 4.1 & 8.1 & \underline{20.1} & 36.7 & 189 & 294 & 313 & 420 & 0.67 & 0.98 & 189 & 234 & 0.0 & 2.2 & 2.2 \\
SRfS~\cite{haefner2018fight} & 62 & 91 & 6.8 & 14.9 & 74.9 & 81.0 & 691 & 738 & 961 & 956 & 1.38 & 1.65 & 311 & 335 & 0.0 & 0.0 & 0.0 \\
EG~\cite{xie2016edge} & 54 &   & 2.4 &   & 30.4 &   & \underline{160} &   & 377 &   & \underline{0.50} &   & 129 &   & \underline{0.7} & 7.4 & 8.1 \\
PDN~\cite{riegler2016deep} & 157 & 197 & 99.3 & 98.9 & 37.9 & 44.5 & 202 & \underline{219} & \underline{299} & 397 & 0.58 & 0.76 & 150 & 186 & \underline{0.7} & \underline{36.0} & 36.8 \\
DG~\cite{gu2017learning} & 56 & 87 & 3.2 & 6.3 & 30.9 & \underline{35.2} & 265 & 285 & 372 & \underline{386} & 0.66 & 0.77 & 166 & 180 & 0.0 & 3.7 & 3.7 \\
DIP~\cite{Ulyanov_2018_CVPR} & 49 & 52 & 8.0 & 4.9 & 85.5 & 84.7 & 796 & 812 & 821 & 924 & 1.19 & 1.18 & 267 & 250 & 0.0 & 0.0 & 0.0 \\
MSG~\cite{hui2016depth} & 41 & \underline{41} & \underline{1.3} & 3.0 & 39.6 & 66.2 & 264 & 458 & 493 & 612 & 0.64 & 0.74 & 181 & 213 & 0.0 & 1.5 & 1.5 \\
	\midrule
DIP-v & \underline{38} & 42 & 1.7 & \underline{2.2} & 48.0 & 60.4 & 238 & 351 & 456 & 516 & \underline{0.50} & \underline{0.61} & \underline{128} & \underline{152} & \underline{0.7} & \textbf{47.8} & \underline{48.5} \\
MSG-v & \textbf{16} & \textbf{26} & \textbf{0.7} & \textbf{1.2} & \textbf{8.5} & \textbf{17.5} & \textbf{76} & \textbf{92} & \textbf{156} & \textbf{181} & \textbf{0.27} & \textbf{0.31} & \textbf{97} & \textbf{100} & \textbf{97.8} & 1.5 & \textbf{99.3} \\
    \bottomrule
\end{tabular}
\caption{Quantitative evaluation on \textquote{Sphere and cylinder} with different RGBs from SimGeo dataset.
The best result is in bold, the second best is underlined.}
\label{tab:SimGeo_Sphere_and_cylinder}
\end{table*}

\begin{table*}
\footnotesize
\centering
\begin{tabular}{l|cc|cc|cc|cc|cc|cc|cc|c|c|c}
    \toprule
    \multicolumn{1}{c}{} & \multicolumn{17}{c}{Lucy} \\
               \cmidrule{2-18}
    \multicolumn{1}{c}{} & \multicolumn{2}{c}{$\mathrm{RMSE_d}$} & \multicolumn{2}{c}{$\mathrm{BadPix_d(5cm)}$} & \multicolumn{2}{c}{$\mathrm{BadPix_v(5)}$} & \multicolumn{2}{c}{$\mathrm{DSSIM_v}$} & \multicolumn{2}{c}{$\mathrm{LPIPS_v}$} & \multicolumn{2}{c}{$\mathrm{Bumpiness_d}$} & \multicolumn{2}{c}{$\ourmetric$} & \multicolumn{1}{c}{User, 1st} & \multicolumn{1}{c}{User, 2nd} & \multicolumn{1}{c}{Top 2} \\
    \multicolumn{1}{c}{} & x4 & x8 & x4 & x8 & x4 & x8 & x4 & x8 & x4 & x8 & x4 & x8 & x4 & x8 & x4 & x4 & x4 \\
    \midrule
    Bicubic & 72 & 103 & 6.8 & 13.0 & \underline{48.8} & \underline{65.0} & \underline{355} & \underline{519} & 398 & 497 & 1.37 & 1.74 & 267 & 328 & \underline{2.2} & \underline{24.3} & 26.5 \\
SRfS~\cite{haefner2018fight} & 82 & 113 & 13.2 & 20.8 & 84.6 & 87.1 & 811 & 857 & 781 & 792 & 1.90 & 2.28 & 367 & 407 & 0.0 & 0.0 & 0.0 \\
EG~\cite{xie2016edge} & 69 &   & 3.5 &   & 56.2 &   & 357 &   & 426 &   & \underline{1.05} &   & \underline{220} &   & 0.0 & 0.7 & 0.7 \\
PDN~\cite{riegler2016deep} & 173 & 234 & 99.0 & 98.8 & 64.9 & 68.9 & 456 & 535 & \underline{368} & 480 & 1.24 & 1.47 & 251 & 303 & 0.0 & 1.5 & 1.5 \\
DG~\cite{gu2017learning} & 69 & 108 & 4.9 & 11.0 & 65.5 & 68.6 & 523 & 562 & 558 & 565 & 1.28 & 1.50 & 249 & 281 & 0.0 & 0.7 & 0.7 \\
DIP~\cite{Ulyanov_2018_CVPR} & \underline{53} & 75 & 4.7 & 11.4 & 87.4 & 95.2 & 827 & 908 & 615 & 778 & 2.02 & 2.93 & 344 & 478 & 0.7 & 0.7 & 1.5 \\
MSG~\cite{hui2016depth} & 54 & \underline{53} & \underline{2.7} & 5.4 & 62.9 & 71.7 & 444 & 577 & 480 & 578 & 1.30 & 1.42 & 259 & 306 & 1.5 & 13.2 & 14.7 \\
	\midrule
DIP-v & \textbf{44} & 55 & 4.6 & \underline{4.4} & 69.0 & 77.5 & 421 & 574 & 446 & \underline{468} & 1.15 & \underline{1.27} & 223 & \underline{239} & 0.0 & \textbf{56.6} & \underline{56.6} \\
MSG-v & 74 & \textbf{47} & \textbf{1.6} & \textbf{3.7} & \textbf{38.8} & \textbf{55.0} & \textbf{205} & \textbf{325} & \textbf{251} & \textbf{348} & \textbf{0.82} & \textbf{0.96} & \textbf{156} & \textbf{195} & \textbf{95.6} & 2.2 & \textbf{97.8} \\
    \bottomrule
\end{tabular}
\caption{Quantitative evaluation on \textquote{Lucy} from SimGeo dataset.
The best result is in bold, the second best is underlined.}
\label{tab:SimGeo_Lucy}
\end{table*}

\begin{table*}
\footnotesize
\centering
\begin{tabular}{l|cc|cc|cc|cc|cc|cc|cc|c|c|c}
    \toprule
    \multicolumn{1}{c}{} & \multicolumn{17}{c}{Painting} \\
               \cmidrule{2-18}
    \multicolumn{1}{c}{} & \multicolumn{2}{c}{$\mathrm{RMSE_d}$} & \multicolumn{2}{c}{$\mathrm{BadPix_d(5cm)}$} & \multicolumn{2}{c}{$\mathrm{BadPix_v(5)}$} & \multicolumn{2}{c}{$\mathrm{DSSIM_v}$} & \multicolumn{2}{c}{$\mathrm{LPIPS_v}$} & \multicolumn{2}{c}{$\mathrm{Bumpiness_d}$} & \multicolumn{2}{c}{$\ourmetric$} & \multicolumn{1}{c}{User, 1st} & \multicolumn{1}{c}{User, 2nd} & \multicolumn{1}{c}{Top 2} \\
    \multicolumn{1}{c}{} & x4 & x8 & x4 & x8 & x4 & x8 & x4 & x8 & x4 & x8 & x4 & x8 & x4 & x8 & x4 & x4 & x4 \\
    \midrule
    Bicubic & 28 & 47 & 2.5 & 5.6 & \underline{57.1} & 64.1 & \underline{423} & 514 & 544 & 649 & 0.95 & 1.15 & 213 & 265 & \underline{4.4} & \textbf{47.8} & \underline{52.2} \\
SRfS~\cite{haefner2018fight} & 39 & 60 & 6.5 & 15.9 & 78.4 & 81.2 & 707 & 722 & 612 & 661 & 1.47 & 1.55 & 308 & 337 & 0.0 & 0.0 & 0.0 \\
EG~\cite{xie2016edge} & 36 &   & 3.1 &   & 61.9 &   & 481 &   & 720 &   & 0.94 &   & 231 &   & 0.0 & 3.7 & 3.7 \\
PDN~\cite{riegler2016deep} & 151 & 215 & 99.3 & 99.2 & 65.2 & 70.2 & 488 & 532 & 669 & 709 & \underline{0.89} & \underline{1.01} & 237 & 275 & \underline{4.4} & 10.3 & 14.7 \\
DG~\cite{gu2017learning} & 31 & 49 & 2.4 & 5.5 & 61.9 & \underline{63.9} & 503 & \underline{506} & 678 & 700 & 1.08 & 1.13 & 232 & 272 & 0.7 & 3.7 & 4.4 \\
DIP~\cite{Ulyanov_2018_CVPR} & 30 & 37 & 4.0 & 4.7 & 80.4 & 79.5 & 802 & 766 & 630 & 612 & 2.18 & 1.82 & 362 & 341 & 0.0 & 0.0 & 0.0 \\
MSG~\cite{hui2016depth} & \underline{21} & \textbf{29} & \underline{1.2} & \underline{2.2} & 63.7 & 67.9 & 495 & 570 & \underline{475} & \underline{507} & 0.97 & 1.12 & \underline{203} & 243 & 2.2 & 5.1 & 7.4 \\
	\midrule
DIP-v & 22 & \underline{32} & 2.3 & 3.2 & 70.1 & 70.3 & 567 & 564 & \textbf{386} & \textbf{501} & 1.07 & 1.12 & 210 & \underline{239} & 2.9 & \underline{21.3} & 24.3 \\
MSG-v & \textbf{17} & 34 & \textbf{0.9} & \textbf{1.8} & \textbf{51.4} & \textbf{58.0} & \textbf{354} & \textbf{410} & 532 & 607 & \textbf{0.67} & \textbf{0.77} & \textbf{142} & \textbf{170} & \textbf{85.3} & 8.1 & \textbf{93.4} \\
    \midrule
    \midrule
    \multicolumn{1}{c}{} & \multicolumn{17}{c}{Sofa} \\
    \midrule
    Bicubic & 38 & 58 & 1.8 & 3.6 & \underline{75.4} & \underline{77.0} & \underline{566} & \underline{616} & 704 & 764 & \underline{2.12} & \underline{2.33} & \underline{212} & \underline{250} & 3.7 & 15.4 & 19.1 \\
SRfS~\cite{haefner2018fight} & 39 & 58 & 2.0 & 3.5 & 82.3 & 88.1 & 715 & 832 & 631 & 743 & 2.97 & 3.45 & 310 & 405 & 0.0 & 0.0 & 0.0 \\
EG~\cite{xie2016edge} & 42 &   & 2.5 &   & 79.0 &   & 598 &   & 767 &   & 2.28 &   & 213 &   & 0.0 & 8.8 & 8.8 \\
PDN~\cite{riegler2016deep} & 86 & 91 & 71.0 & 70.8 & 83.3 & 83.0 & 641 & 658 & 784 & 763 & 2.40 & 2.50 & 260 & 264 & 0.7 & 3.7 & 4.4 \\
DG~\cite{gu2017learning} & 41 & 63 & 3.2 & 4.4 & 77.7 & 77.9 & 624 & 632 & 823 & 855 & 2.30 & \underline{2.33} & 255 & 263 & 0.0 & 5.1 & 5.1 \\
DIP~\cite{Ulyanov_2018_CVPR} & 45 & 57 & 7.1 & 12.7 & 93.1 & 94.0 & 928 & 946 & 758 & 738 & 3.91 & 3.99 & 518 & 560 & 0.0 & 0.0 & 0.0 \\
MSG~\cite{hui2016depth} & \textbf{27} & \textbf{36} & 1.2 & 2.3 & 80.6 & 85.7 & 718 & 791 & \underline{606} & \underline{610} & 2.71 & 3.22 & 254 & 316 & 0.0 & 0.7 & 0.7 \\
	\midrule
DIP-v & \textbf{27} & \underline{43} & \underline{0.9} & \underline{2.0} & 79.1 & 82.5 & 645 & 718 & \textbf{414} & \textbf{585} & 2.67 & 3.07 & 215 & 266 & \underline{19.1} & \textbf{47.8} & \underline{66.9} \\
MSG-v & \underline{35} & 44 & \textbf{0.7} & \textbf{1.6} & \textbf{74.0} & \textbf{75.7} & \textbf{537} & \textbf{585} & 710 & 759 & \textbf{1.96} & \textbf{2.10} & \textbf{165} & \textbf{196} & \textbf{76.5} & \underline{18.4} & \textbf{94.9} \\
    \midrule
    \midrule
    \multicolumn{1}{c}{} & \multicolumn{17}{c}{Plant} \\
    \midrule
    Bicubic & 38 & 58 & 3.7 & 6.4 & \underline{75.9} & \underline{79.9} & \underline{562} & \underline{610} & 688 & 763 & 1.58 & 1.79 & 249 & 290 & 1.5 & \underline{22.1} & 23.5 \\
SRfS~\cite{haefner2018fight} & 46 & 65 & 5.8 & 9.5 & 82.9 & 85.0 & 658 & 692 & 632 & 649 & 1.96 & 2.13 & 280 & 309 & 0.0 & 0.0 & 0.0 \\
EG~\cite{xie2016edge} & 43 &   & 4.5 &   & 82.2 &   & 568 &   & 677 &   & 1.64 &   & 255 &   & 0.0 & 0.7 & 0.7 \\
PDN~\cite{riegler2016deep} & 88 & 89 & 94.5 & 37.8 & 79.5 & 82.5 & 574 & 612 & 659 & 699 & \underline{1.46} & \underline{1.60} & 269 & 305 & 4.4 & 7.4 & 11.8 \\
DG~\cite{gu2017learning} & 40 & 63 & 3.9 & 6.7 & 79.5 & 81.1 & 611 & 622 & 745 & 785 & 1.67 & 1.70 & 268 & 291 & 2.2 & 11.0 & 13.2 \\
DIP~\cite{Ulyanov_2018_CVPR} & 38 & 47 & 6.9 & 6.1 & 93.9 & 92.8 & 919 & 880 & 764 & 723 & 4.33 & 3.95 & 490 & 437 & 0.0 & 0.7 & 0.7 \\
MSG~\cite{hui2016depth} & \underline{31} & \underline{44} & \underline{2.3} & \textbf{3.7} & 78.0 & 81.8 & 571 & 645 & \underline{582} & \textbf{495} & 1.62 & 1.84 & \underline{234} & 285 & 0.0 & 11.8 & 11.8 \\
	\midrule
DIP-v & \underline{31} & \textbf{40} & 4.7 & 4.8 & 83.5 & 84.1 & 694 & 707 & \textbf{463} & \underline{555} & 2.25 & 2.21 & 262 & \underline{276} & \underline{11.0} & \textbf{33.1} & \underline{44.1} \\
MSG-v & \textbf{27} & \underline{44} & \textbf{1.8} & \underline{3.9} & \textbf{74.3} & \textbf{77.8} & \textbf{524} & \textbf{575} & 639 & 720 & \textbf{1.31} & \textbf{1.47} & \textbf{194} & \textbf{236} & \textbf{80.9} & 13.2 & \textbf{94.1} \\
    \bottomrule
\end{tabular}
\caption{Quantitative evaluation on RGBD frames from ICL-NUIM \textquote{Living Room} sequence.
The best result is in bold, the second best is underlined.}
\label{tab:ICL-NUIM_Livingroom}
\end{table*}

\begin{table*}
\footnotesize
\centering
\begin{tabular}{l|cc|cc|cc|cc|cc|cc|cc|c|c|c}
    \toprule
    \multicolumn{1}{c}{} & \multicolumn{17}{c}{Office} \\
               \cmidrule{2-18}
    \multicolumn{1}{c}{} & \multicolumn{2}{c}{$\mathrm{RMSE_d}$} & \multicolumn{2}{c}{$\mathrm{BadPix_d(5cm)}$} & \multicolumn{2}{c}{$\mathrm{BadPix_v(5)}$} & \multicolumn{2}{c}{$\mathrm{DSSIM_v}$} & \multicolumn{2}{c}{$\mathrm{LPIPS_v}$} & \multicolumn{2}{c}{$\mathrm{Bumpiness_d}$} & \multicolumn{2}{c}{$\ourmetric$} & \multicolumn{1}{c}{User, 1st} & \multicolumn{1}{c}{User, 2nd} & \multicolumn{1}{c}{Top 2} \\
    \multicolumn{1}{c}{} & x4 & x8 & x4 & x8 & x4 & x8 & x4 & x8 & x4 & x8 & x4 & x8 & x4 & x8 & x4 & x4 & x4 \\
    \midrule
    Bicubic & 47 & 80 & 4.0 & 7.8 & \underline{24.4} & \underline{34.1} & \underline{216} & \underline{285} & \underline{412} & 594 & 0.81 & 0.95 & 208 & 254 & \underline{19.9} & \textbf{44.1} & \underline{64.0} \\
SRfS~\cite{haefner2018fight} & 49 & 89 & 5.8 & 14.4 & 53.4 & 54.4 & 595 & 593 & 690 & 636 & 1.71 & 1.66 & 298 & 302 & 0.0 & 0.0 & 0.0 \\
PDN~\cite{riegler2016deep} & 185 & 185 & 99.3 & 90.5 & 36.5 & 50.2 & 250 & 294 & 457 & 518 & \underline{0.76} & \underline{0.92} & 234 & 272 & 0.7 & 3.7 & 4.4 \\
DG~\cite{gu2017learning} & 49 & 85 & 9.0 & 11.9 & 36.5 & 37.6 & 319 & 330 & 534 & 571 & 1.03 & 1.05 & 240 & 266 & 0.0 & 0.7 & 0.7 \\
DIP~\cite{Ulyanov_2018_CVPR} & 76 & 109 & 30.3 & 48.2 & 72.1 & 73.9 & 726 & 819 & 690 & 797 & 2.45 & 2.70 & 372 & 408 & 1.5 & 1.5 & 2.9 \\
MSG~\cite{hui2016depth} & \underline{35} & \textbf{48} & \underline{2.4} & \underline{6.8} & 35.4 & 44.5 & 263 & 360 & 415 & 543 & 0.83 & 0.95 & \underline{199} & 247 & 2.2 & 3.7 & 5.9 \\
	\midrule
DIP-v & 40 & \underline{65} & 3.8 & 7.4 & 45.4 & 47.9 & 311 & 352 & 414 & \underline{504} & 1.08 & 1.18 & 205 & \underline{235} & 17.6 & \underline{25.0} & 42.6 \\
MSG-v & \textbf{32} & \underline{65} & \textbf{1.9} & \textbf{5.3} & \textbf{19.3} & \textbf{29.6} & \textbf{157} & \textbf{224} & \textbf{313} & \textbf{432} & \textbf{0.59} & \textbf{0.72} & \textbf{151} & \textbf{198} & \textbf{58.1} & 21.3 & \textbf{79.4} \\
    \midrule
    \midrule
    \multicolumn{1}{c}{} & \multicolumn{17}{c}{Coat rack} \\
    \midrule
    Bicubic & \underline{13} & 20 & 1.5 & 3.0 & \underline{73.1} & \underline{75.3} & \underline{507} & 539 & 537 & 651 & 0.54 & 0.60 & 171 & 196 & 0.0 & \underline{19.1} & 19.1 \\
SRfS~\cite{haefner2018fight} & 24 & 28 & 3.8 & 5.4 & 82.3 & 80.5 & 672 & 556 & 650 & 612 & 0.83 & 0.57 & 237 & 203 & 0.0 & 0.0 & 0.0 \\
EG~\cite{xie2016edge} & \underline{13} &   & 1.2 &   & 77.8 &   & 541 &   & 550 &   & 0.55 &   & 186 &   & 0.7 & 7.4 & 8.1 \\
PDN~\cite{riegler2016deep} & 140 & 191 & 99.6 & 99.9 & 77.1 & 78.0 & 544 & 557 & 621 & 631 & \underline{0.48} & \underline{0.50} & 178 & 193 & \underline{5.1} & \textbf{28.7} & \underline{33.8} \\
DG~\cite{gu2017learning} & \underline{13} & 20 & 1.4 & 3.2 & 74.4 & 75.6 & 530 & \underline{532} & 593 & 621 & 0.54 & 0.58 & 166 & 201 & 0.0 & 9.6 & 9.6 \\
DIP~\cite{Ulyanov_2018_CVPR} & 15 & 24 & 1.9 & 3.5 & 85.5 & 85.4 & 766 & 701 & 625 & 624 & 1.15 & 0.97 & 256 & 246 & 4.4 & 2.2 & 6.6 \\
MSG~\cite{hui2016depth} & \textbf{11} & \textbf{17} & \underline{0.9} & \textbf{1.6} & 73.3 & 76.1 & 522 & 546 & 523 & \underline{554} & 0.51 & 0.55 & \underline{165} & 189 & 2.2 & 16.2 & 18.4 \\
	\midrule
DIP-v & \underline{13} & \textbf{17} & 1.8 & \underline{2.0} & 75.2 & 75.5 & 543 & 542 & \textbf{422} & \textbf{463} & 0.62 & 0.60 & 171 & \underline{181} & 0.7 & 12.5 & 13.2 \\
MSG-v & \textbf{11} & \underline{18} & \textbf{0.8} & 2.2 & \textbf{71.4} & \textbf{74.2} & \textbf{482} & \textbf{502} & \underline{516} & 563 & \textbf{0.42} & \textbf{0.48} & \textbf{136} & \textbf{161} & \textbf{86.8} & 4.4 & \textbf{91.2} \\
    \midrule
    \midrule
    \multicolumn{1}{c}{} & \multicolumn{17}{c}{Displays} \\
    \midrule
    Bicubic & 41 & 63 & 3.2 & 6.4 & \underline{49.9} & \underline{54.9} & \underline{315} & \underline{374} & 460 & 585 & 0.92 & 1.08 & 208 & 256 & 0.7 & \underline{21.3} & 22.1 \\
SRfS~\cite{haefner2018fight} & 53 & 75 & 9.0 & 17.3 & 61.9 & 67.3 & 500 & 591 & 599 & 659 & 1.35 & 1.60 & 288 & 328 & 0.0 & 0.0 & 0.0 \\
EG~\cite{xie2016edge} & 46 &   & 5.9 &   & 66.7 &   & 388 &   & 587 &   & 0.94 &   & 216 &   & 0.0 & 2.9 & 2.9 \\
PDN~\cite{riegler2016deep} & 159 & 220 & 99.2 & 99.0 & 55.4 & 57.2 & 381 & 403 & 547 & 580 & \underline{0.85} & \underline{0.95} & 242 & 275 & 0.0 & 9.6 & 9.6 \\
DG~\cite{gu2017learning} & 43 & 66 & 5.8 & 6.7 & 56.5 & 56.7 & 395 & 406 & 606 & 601 & 1.06 & 1.10 & 243 & 265 & 0.7 & 2.9 & 3.7 \\
DIP~\cite{Ulyanov_2018_CVPR} & 52 & 60 & 13.4 & 9.7 & 76.9 & 74.6 & 732 & 724 & 672 & 645 & 2.36 & 2.06 & 365 & 344 & 0.7 & 0.7 & 1.5 \\
MSG~\cite{hui2016depth} & \underline{26} & \textbf{42} & \underline{1.7} & 4.4 & 53.9 & 58.0 & 367 & 430 & 461 & \underline{493} & 0.97 & 1.08 & 204 & 251 & 0.0 & 5.9 & 5.9 \\
	\midrule
DIP-v & 32 & 45 & 2.4 & \underline{4.0} & 53.7 & 57.6 & 336 & 407 & \textbf{344} & \textbf{409} & 1.00 & 1.18 & \underline{191} & \underline{221} & \underline{5.9} & \textbf{51.5} & \underline{57.4} \\
MSG-v & \textbf{23} & \underline{43} & \textbf{1.4} & \textbf{3.5} & \textbf{47.2} & \textbf{51.0} & \textbf{271} & \textbf{324} & \underline{451} & 531 & \textbf{0.69} & \textbf{0.80} & \textbf{152} & \textbf{190} & \textbf{91.9} & 5.1 & \textbf{97.1} \\
    \bottomrule
\end{tabular}
\caption{Quantitative evaluation on RGBD frames from ICL-NUIM \textquote{Office Room} sequence.
The best result is in bold, the second best is underlined.}
\label{tab:ICL-NUIM_Officeroom}
\end{table*}

\begin{table*}
\footnotesize
\centering
\begin{tabular}{l|cc|cc|cc|cc|cc|cc|cc|c|c|c}
    \toprule
    \multicolumn{1}{c}{} & \multicolumn{17}{c}{Vintage} \\
               \cmidrule{2-18}
    \multicolumn{1}{c}{} & \multicolumn{2}{c}{$\mathrm{RMSE_d}$} & \multicolumn{2}{c}{$\mathrm{BadPix_d(5cm)}$} & \multicolumn{2}{c}{$\mathrm{BadPix_v(5)}$} & \multicolumn{2}{c}{$\mathrm{DSSIM_v}$} & \multicolumn{2}{c}{$\mathrm{LPIPS_v}$} & \multicolumn{2}{c}{$\mathrm{Bumpiness_d}$} & \multicolumn{2}{c}{$\ourmetric$} & \multicolumn{1}{c}{User, 1st} & \multicolumn{1}{c}{User, 2nd} & \multicolumn{1}{c}{Top 2} \\
    \multicolumn{1}{c}{} & x4 & x8 & x4 & x8 & x4 & x8 & x4 & x8 & x4 & x8 & x4 & x8 & x4 & x8 & x4 & x4 & x4 \\
    \midrule
    Bicubic & 67 & 98 & 4.6 & 9.0 & \underline{72.8} & \textbf{77.3} & \underline{558} & \underline{649} & 602 & 729 & 1.51 & 1.64 & \underline{258} & 302 & 5.9 & \underline{28.7} & 34.6 \\
SRfS~\cite{haefner2018fight} & 101 & 145 & 16.8 & 32.3 & 83.7 & 87.2 & 721 & 749 & 631 & \underline{634} & 1.64 & 1.68 & 346 & 382 & 0.0 & 0.0 & 0.0 \\
PDN~\cite{riegler2016deep} & 140 & 174 & 67.6 & 79.0 & 82.3 & 85.7 & 663 & 714 & 706 & 700 & 1.51 & 1.57 & 319 & 350 & 0.0 & 0.0 & 0.0 \\
DG~\cite{gu2017learning} & 72 & 107 & 7.1 & 10.4 & 79.4 & 80.1 & 666 & 669 & 796 & 840 & \underline{1.50} & \underline{1.52} & 290 & \underline{300} & 0.0 & 0.7 & 0.7 \\
DIP~\cite{Ulyanov_2018_CVPR} & 74 & 117 & 24.8 & 46.9 & 93.6 & 94.2 & 953 & 965 & 910 & 872 & 4.01 & 4.16 & 656 & 687 & 0.7 & 0.7 & 1.5 \\
MSG~\cite{hui2016depth} & \underline{41} & \textbf{59} & 3.2 & \underline{6.8} & 80.6 & 84.6 & 708 & 785 & \textbf{510} & \textbf{610} & 1.62 & 1.85 & 292 & 364 & 0.0 & 9.6 & 9.6 \\
	\midrule
DIP-v & 42 & 67 & \underline{2.7} & \textbf{5.9} & 85.2 & 88.8 & 804 & 884 & \underline{579} & 674 & 1.94 & 2.48 & 343 & 435 & \underline{25.7} & \textbf{44.1} & \underline{69.9} \\
MSG-v & \textbf{33} & \underline{65} & \textbf{2.5} & \textbf{5.9} & \textbf{71.4} & \underline{77.6} & \textbf{536} & \textbf{643} & 670 & 702 & \textbf{1.29} & \textbf{1.43} & \textbf{211} & \textbf{268} & \textbf{67.6} & 16.2 & \textbf{83.8} \\
    \midrule
    \midrule
    \multicolumn{1}{c}{} & \multicolumn{17}{c}{Recycle} \\
    \midrule
    Bicubic & 587 & 880 & 9.2 & 16.6 & \textbf{70.6} & \textbf{78.6} & \textbf{575} & 721 & \underline{474} & 576 & \textbf{1.23} & \textbf{1.17} & 329 & 398 & 0.0 & \underline{11.0} & 11.0 \\
SRfS~\cite{haefner2018fight} & 47 & 72 & 10.2 & 22.1 & 86.1 & 88.8 & 715 & 772 & 610 & 623 & 1.68 & \underline{1.81} & 376 & 410 & 0.0 & 0.0 & 0.0 \\
PDN~\cite{riegler2016deep} & 95 & 128 & 90.5 & 79.8 & 84.0 & 85.7 & 635 & \textbf{701} & 523 & 589 & 1.66 & 2.18 & 364 & 457 & 0.0 & 6.6 & 6.6 \\
DG~\cite{gu2017learning} & 39 & 82 & \underline{3.4} & 11.7 & 81.6 & 83.6 & 696 & \underline{719} & 602 & 617 & 1.75 & 1.99 & \underline{328} & \underline{383} & \underline{2.9} & \textbf{65.4} & \underline{68.4} \\
DIP~\cite{Ulyanov_2018_CVPR} & \underline{29} & \underline{45} & 3.9 & 9.3 & 91.0 & 91.8 & 871 & 923 & 576 & 605 & 2.95 & 3.31 & 434 & 500 & 1.5 & 5.9 & 7.4 \\
MSG~\cite{hui2016depth} & 106 & 1182 & 5.8 & 11.9 & 82.8 & 89.6 & 741 & 869 & 624 & 661 & 2.60 & 3.01 & 485 & 550 & 0.7 & 0.0 & 0.7 \\
	\midrule
DIP-v & \textbf{20} & \textbf{34} & \textbf{1.5} & \textbf{4.2} & 78.9 & 85.0 & \textbf{575} & 735 & \textbf{388} & \textbf{485} & \underline{1.56} & 1.86 & \textbf{273} & \textbf{332} & \textbf{94.9} & 3.7 & \textbf{98.5} \\
MSG-v & 51 & 76 & 3.9 & \underline{7.9} & \underline{73.9} & \underline{82.1} & \underline{603} & 737 & 520 & \underline{564} & 1.66 & 2.02 & 368 & 473 & 0.0 & 7.4 & 7.4 \\
    \bottomrule
\end{tabular}
\caption{Quantitative evaluation on samples with small number of missing measurements from Middlebury dataset.
The best result is in bold, the second best is underlined.}
\label{tab:Middlebury_Simple}
\end{table*}

\begin{table*}
\footnotesize
\centering
\begin{tabular}{l|cc|cc|cc|cc|cc|cc|cc|c|c|c}
    \toprule
    \multicolumn{1}{c}{} & \multicolumn{17}{c}{Umbrella} \\
               \cmidrule{2-18}
    \multicolumn{1}{c}{} & \multicolumn{2}{c}{$\mathrm{RMSE_d}$} & \multicolumn{2}{c}{$\mathrm{BadPix_d(5cm)}$} & \multicolumn{2}{c}{$\mathrm{BadPix_v(5)}$} & \multicolumn{2}{c}{$\mathrm{DSSIM_v}$} & \multicolumn{2}{c}{$\mathrm{LPIPS_v}$} & \multicolumn{2}{c}{$\mathrm{Bumpiness_d}$} & \multicolumn{2}{c}{$\ourmetric$} & \multicolumn{1}{c}{User, 1st} & \multicolumn{1}{c}{User, 2nd} & \multicolumn{1}{c}{Top 2} \\
    \multicolumn{1}{c}{} & x4 & x8 & x4 & x8 & x4 & x8 & x4 & x8 & x4 & x8 & x4 & x8 & x4 & x8 & x4 & x4 & x4 \\
    \midrule
    Bicubic & 1013 & 1507 & 6.9 & 12.1 & \textbf{77.7} & \textbf{80.9} & \textbf{749} & \underline{837} & 747 & 886 & \textbf{0.60} & \textbf{0.60} & \underline{323} & \underline{380} & \underline{5.9} & \textbf{35.3} & \underline{41.2} \\
SRfS~\cite{haefner2018fight} & 148 & 217 & 19.4 & 35.5 & 87.5 & 90.8 & 843 & 853 & 797 & 831 & 0.71 & \underline{0.78} & 397 & 443 & 0.0 & 0.0 & 0.0 \\
PDN~\cite{riegler2016deep} & 220 & 287 & 94.9 & 89.1 & 86.6 & 88.1 & 799 & \textbf{828} & 847 & 882 & 0.79 & 1.13 & 367 & 452 & 3.7 & \underline{22.8} & 26.5 \\
DG~\cite{gu2017learning} & 365 & 507 & 9.1 & 20.3 & 84.6 & 87.3 & 846 & 878 & 781 & 856 & 0.92 & 1.36 & 399 & 457 & 0.0 & 0.7 & 0.7 \\
DIP~\cite{Ulyanov_2018_CVPR} & 138 & \underline{145} & 48.5 & 21.6 & 90.5 & 93.2 & 915 & 953 & 737 & \underline{722} & 1.19 & 1.65 & 467 & 528 & 2.9 & 16.2 & 19.1 \\
MSG~\cite{hui2016depth} & 292 & 555 & 7.4 & 12.4 & 84.3 & 88.1 & 834 & 896 & \underline{678} & 787 & 1.27 & 1.47 & 442 & 496 & 0.0 & 0.7 & 0.7 \\
	\midrule
DIP-v & \textbf{91} & \textbf{129} & \textbf{3.4} & \textbf{5.7} & 83.4 & 85.3 & 796 & 854 & \textbf{604} & \textbf{598} & \underline{0.67} & 0.79 & \textbf{318} & \textbf{352} & \textbf{82.4} & 8.1 & \textbf{90.4} \\
MSG-v & \underline{129} & 218 & \underline{5.2} & \underline{9.7} & \underline{79.1} & \underline{82.3} & \underline{778} & 842 & 800 & 890 & 0.72 & 0.89 & 348 & 427 & 5.1 & 16.2 & 21.3 \\
    \midrule
    \midrule
    \multicolumn{1}{c}{} & \multicolumn{17}{c}{Classroom1} \\
    \midrule
    Bicubic & 966 & 1371 & 6.7 & \underline{9.0} & \textbf{75.8} & \textbf{78.3} & \textbf{636} & \textbf{728} & \underline{581} & 784 & \textbf{0.41} & \textbf{0.30} & \underline{268} & \textbf{295} & 12.5 & \textbf{37.5} & \underline{50.0} \\
SRfS~\cite{haefner2018fight} & 135 & 202 & 18.5 & 28.5 & 82.6 & 85.7 & 761 & 781 & 718 & 756 & 0.62 & \underline{0.62} & 332 & 363 & 0.0 & 0.0 & 0.0 \\
PDN~\cite{riegler2016deep} & 239 & 324 & 96.0 & 91.0 & 81.5 & 82.9 & 739 & 759 & 751 & 807 & 0.62 & 0.76 & 279 & 342 & \underline{16.9} & \underline{26.5} & 43.4 \\
DG~\cite{gu2017learning} & 307 & 503 & 8.8 & 16.8 & 82.0 & 82.7 & 743 & 762 & 766 & 812 & 0.74 & 0.87 & 313 & 337 & 0.0 & 1.5 & 1.5 \\
DIP~\cite{Ulyanov_2018_CVPR} & \underline{96} & \underline{145} & 17.0 & 22.4 & 94.4 & 94.6 & 956 & 952 & 789 & 751 & 1.94 & 2.12 & 540 & 557 & 0.0 & 1.5 & 1.5 \\
MSG~\cite{hui2016depth} & 297 & 408 & 7.3 & 10.0 & 81.2 & 83.8 & 723 & 810 & 626 & \underline{604} & 0.90 & 1.01 & 351 & 391 & 0.0 & 0.7 & 0.7 \\
	\midrule
DIP-v & \textbf{69} & \textbf{117} & \textbf{4.1} & 9.3 & 81.0 & 86.0 & 700 & 789 & \textbf{516} & \textbf{537} & 0.64 & 0.86 & \textbf{266} & \underline{327} & \textbf{64.0} & 18.4 & \textbf{82.4} \\
MSG-v & 127 & 203 & \underline{5.4} & \textbf{8.4} & \underline{76.9} & \underline{79.4} & \underline{678} & \underline{735} & 739 & 803 & \underline{0.60} & 0.64 & 283 & 330 & 2.2 & 11.8 & 14.0 \\
    \bottomrule
\end{tabular}
\caption{Quantitative evaluation on samples with small number of missing measurements from Middlebury dataset.
The best result is in bold, the second best is underlined.}
\label{tab:Middlebury_Simple}
\end{table*}

\begin{table*}
\footnotesize
\centering
\begin{tabular}{l|cc|cc|cc|cc|cc|cc|cc|c|c|c}
    \toprule
    \multicolumn{1}{c}{} & \multicolumn{17}{c}{Playroom} \\
               \cmidrule{2-18}
    \multicolumn{1}{c}{} & \multicolumn{2}{c}{$\mathrm{RMSE_d}$} & \multicolumn{2}{c}{$\mathrm{BadPix_d(5cm)}$} & \multicolumn{2}{c}{$\mathrm{BadPix_v(5)}$} & \multicolumn{2}{c}{$\mathrm{DSSIM_v}$} & \multicolumn{2}{c}{$\mathrm{LPIPS_v}$} & \multicolumn{2}{c}{$\mathrm{Bumpiness_d}$} & \multicolumn{2}{c}{$\ourmetric$} & \multicolumn{1}{c}{User, 1st} & \multicolumn{1}{c}{User, 2nd} & \multicolumn{1}{c}{Top 2} \\
    \multicolumn{1}{c}{} & x4 & x8 & x4 & x8 & x4 & x8 & x4 & x8 & x4 & x8 & x4 & x8 & x4 & x8 & x4 & x4 & x4 \\
    \midrule
    Bicubic & 1263 & 1744 & 14.4 & 20.4 & \textbf{72.0} & \textbf{76.9} & \textbf{684} & \textbf{783} & \underline{509} & 675 & \textbf{0.80} & \textbf{0.52} & \underline{386} & \underline{441} & 0.0 & 2.2 & 2.2 \\
SRfS~\cite{haefner2018fight} & 97 & 151 & 26.9 & 42.1 & 88.1 & 91.2 & 802 & 829 & 663 & 715 & \underline{1.24} & \underline{1.08} & 493 & 540 & 0.0 & 0.0 & 0.0 \\
PDN~\cite{riegler2016deep} & 181 & 253 & 85.0 & 69.7 & 86.3 & 89.2 & 820 & 862 & 583 & 656 & 1.54 & 1.88 & 472 & 543 & \underline{25.7} & \textbf{61.8} & \underline{87.5} \\
DG~\cite{gu2017learning} & 425 & 133 & 22.4 & 25.0 & 85.4 & 86.0 & 845 & 826 & 779 & 691 & 1.96 & 1.63 & 519 & 469 & 1.5 & 2.2 & 3.7 \\
DIP~\cite{Ulyanov_2018_CVPR} & \underline{58} & \underline{91} & 18.4 & 20.0 & 93.0 & 93.2 & 941 & 937 & 647 & \underline{612} & 3.09 & 2.86 & 602 & 592 & 1.5 & 2.9 & 4.4 \\
MSG~\cite{hui2016depth} & 433 & 349 & 16.1 & 22.3 & 85.8 & 89.9 & 855 & 911 & 685 & 705 & 2.51 & 2.74 & 576 & 616 & 0.0 & 0.0 & 0.0 \\
	\midrule
DIP-v & \textbf{49} & \textbf{83} & \textbf{5.4} & \textbf{12.2} & 83.8 & 88.5 & 728 & 847 & \textbf{459} & \textbf{530} & 1.29 & 1.52 & \textbf{357} & \textbf{433} & \textbf{70.6} & \underline{27.2} & \textbf{97.8} \\
MSG-v & 112 & 166 & \underline{9.4} & \underline{15.5} & \underline{75.2} & \underline{80.1} & \underline{721} & \underline{810} & 565 & 615 & 1.46 & 1.67 & 453 & 510 & 0.0 & 3.7 & 3.7 \\
    \midrule
    \midrule
    \multicolumn{1}{c}{} & \multicolumn{17}{c}{Backpack} \\
    \midrule
    Bicubic & 985 & 1078 & 14.3 & 11.5 & \textbf{62.7} & \textbf{69.4} & \textbf{639} & \textbf{730} & \underline{564} & 692 & \textbf{0.60} & \textbf{0.45} & \textbf{392} & \textbf{424} & 2.2 & \underline{34.6} & 36.8 \\
SRfS~\cite{haefner2018fight} & 69 & 83 & 18.9 & 25.5 & 89.9 & 89.9 & 831 & 847 & 630 & 651 & 1.37 & 1.26 & 500 & 505 & 0.0 & 0.0 & 0.0 \\
PDN~\cite{riegler2016deep} & 173 & 207 & 81.6 & 65.2 & 80.4 & 85.4 & 770 & 820 & 609 & 719 & 1.59 & 1.96 & 519 & 553 & \underline{3.7} & \textbf{37.5} & \underline{41.2} \\
DG~\cite{gu2017learning} & 325 & 465 & 26.5 & 39.9 & 77.0 & 82.0 & 765 & 808 & 650 & 696 & 1.62 & 2.07 & 529 & 545 & 0.7 & 2.9 & 3.7 \\
DIP~\cite{Ulyanov_2018_CVPR} & \underline{41} & \underline{67} & \underline{8.2} & 17.9 & 93.2 & 94.5 & 943 & 984 & 766 & 692 & 3.36 & 2.95 & 639 & 645 & 1.5 & 11.8 & 13.2 \\
MSG~\cite{hui2016depth} & 211 & 170 & 15.1 & \underline{10.4} & 76.5 & 86.9 & 762 & 856 & 671 & 723 & 2.11 & 2.31 & 577 & 609 & 0.7 & 0.0 & 0.7 \\
	\midrule
DIP-v & \textbf{38} & \textbf{62} & \textbf{6.2} & 12.9 & 82.5 & 88.7 & 677 & 768 & \textbf{457} & \textbf{496} & 1.31 & 1.43 & \underline{409} & \underline{448} & \textbf{90.4} & 5.1 & \textbf{95.6} \\
MSG-v & 113 & 89 & 10.8 & \textbf{5.9} & \underline{65.3} & \underline{72.5} & \underline{663} & \underline{752} & 577 & \underline{635} & \underline{1.07} & \underline{1.15} & 462 & 480 & 0.0 & 5.9 & 5.9 \\
    \midrule
    \midrule
    \multicolumn{1}{c}{} & \multicolumn{17}{c}{Jadeplant} \\
    \midrule
    Bicubic & 1017 & 1297 & 19.3 & 18.4 & \textbf{68.8} & \textbf{75.7} & \underline{695} & \underline{788} & \underline{545} & 696 & \textbf{0.97} & \textbf{0.62} & \textbf{449} & \textbf{464} & 2.3 & \underline{27.7} & 30.0 \\
SRfS~\cite{haefner2018fight} & 105 & 143 & 39.5 & 48.9 & 87.2 & 92.7 & 787 & 839 & 637 & 719 & 1.96 & 1.70 & 551 & 583 & 0.0 & 0.0 & 0.0 \\
PDN~\cite{riegler2016deep} & 161 & 205 & 81.8 & 62.0 & 82.4 & 88.0 & 778 & 849 & 551 & \underline{625} & 1.95 & 2.20 & 512 & 572 & \underline{19.1} & \textbf{41.4} & \underline{60.5} \\
DG~\cite{gu2017learning} & 326 & 512 & 27.8 & 47.6 & 82.6 & 86.8 & 791 & 823 & 718 & 670 & 2.28 & 2.66 & 567 & 601 & 0.0 & 0.5 & 0.5 \\
DIP~\cite{Ulyanov_2018_CVPR} & \textbf{70} & \underline{121} & \underline{16.7} & 32.8 & 91.2 & 92.3 & 913 & 911 & 735 & 764 & 3.19 & 3.21 & 615 & 638 & 0.0 & 0.5 & 0.5 \\
MSG~\cite{hui2016depth} & 216 & 263 & 21.1 & \underline{17.8} & 81.5 & 87.6 & 796 & 880 & 751 & 783 & 2.73 & 2.90 & 614 & 649 & 0.0 & 0.0 & 0.0 \\
	\midrule
DIP-v & \underline{84} & \underline{121} & 21.8 & 24.0 & 86.6 & 89.5 & 820 & 870 & \textbf{542} & 654 & 1.92 & 1.99 & \underline{503} & 535 & \textbf{78.2} & 20.5 & \textbf{98.6} \\
MSG-v & 109 & \textbf{117} & \textbf{13.9} & \textbf{11.2} & \underline{71.0} & \underline{79.0} & \textbf{688} & \textbf{781} & 605 & \textbf{622} & \underline{1.61} & \underline{1.66} & 507 & \underline{529} & 0.5 & 8.2 & 8.6 \\
    \bottomrule
\end{tabular}
\caption{Quantitative evaluation on samples with large number of missing measurements from Middlebury dataset.
The best result is in bold, the second best is underlined.}
\label{tab:Middlebury_Complex}
\end{table*}
\endgroup

    {\clearpage\newpage\small\bibliographystyle{ieee_fullname}\bibliography{src/bib}}

\begin{thebibliography}{10}\itemsep=-1pt

\bibitem{agresti2017deep}
Gianluca Agresti, Ludovico Minto, Giulio Marin, and Pietro Zanuttigh.
\newblock Deep learning for confidence information in stereo and tof data
  fusion.
\newblock In {\em Proceedings of the IEEE Conference on Computer Vision and
  Pattern Recognition}, pages 697--705, 2017.

\bibitem{pmlr-v80-bojanowski18a}
Piotr Bojanowski, Armand Joulin, David Lopez-Pas, and Arthur Szlam.
\newblock Optimizing the latent space of generative networks.
\newblock In Jennifer Dy and Andreas Krause, editors, {\em Proceedings of the
  35th International Conference on Machine Learning}, volume~80 of {\em
  Proceedings of Machine Learning Research}, pages 600--609, Stockholmsmässan,
  Stockholm Sweden, 10--15 Jul 2018. PMLR.

\bibitem{Butler:ECCV:2012}
Daniel~J. Butler, Jonas Wulff, Garrett~B. Stanley, and Michael~J. Black.
\newblock A naturalistic open source movie for optical flow evaluation.
\newblock In {A. Fitzgibbon et al. (Eds.)}, editor, {\em European Conf. on
  Computer Vision (ECCV)}, Part IV, LNCS 7577, pages 611--625. Springer-Verlag,
  Oct. 2012.

\bibitem{chen2018single}
Baoliang Chen and Cheolkon Jung.
\newblock Single depth image super-resolution using convolutional neural
  networks.
\newblock In {\em 2018 IEEE International Conference on Acoustics, Speech and
  Signal Processing (ICASSP)}, pages 1473--1477. IEEE, 2018.

\bibitem{chen2018estimating}
Zhao Chen, Vijay Badrinarayanan, Gilad Drozdov, and Andrew Rabinovich.
\newblock Estimating depth from rgb and sparse sensing.
\newblock {\em CoRR}, abs/1804.02771, 2018.

\bibitem{cheng2018depth}
Xinjing Cheng, Peng Wang, and Ruigang Yang.
\newblock Depth estimation via affinity learned with convolutional spatial
  propagation network.
\newblock In {\em European Conference on Computer Vision}, pages 108--125.
  Springer, Cham, 2018.

\bibitem{cheon2019generative}
Manri Cheon, Jun-Hyuk Kim, Jun-Ho Choi, and Jong-Seok Lee.
\newblock Generative adversarial network-based image super-resolution using
  perceptual content losses.
\newblock In Laura Leal-Taix{\'e} and Stefan Roth, editors, {\em ECCV
  Workshops}, pages 51--62, Cham, 2019. Springer International Publishing.

\bibitem{chodosh2018deep}
Nathaniel Chodosh, Chaoyang Wang, and Simon Lucey.
\newblock Deep convolutional compressed sensing for lidar depth completion.
\newblock {\em arXiv preprint arXiv:1803.08949}, 2018.

\bibitem{eigen2014depth}
David Eigen, Christian Puhrsch, and Rob Fergus.
\newblock Depth map prediction from a single image using a multi-scale deep
  network.
\newblock In {\em Advances in neural information processing systems}, pages
  2366--2374, 2014.

\bibitem{ferstl2013image}
David Ferstl, Christian Reinbacher, Rene Ranftl, Matthias R{\"u}ther, and Horst
  Bischof.
\newblock Image guided depth upsampling using anisotropic total generalized
  variation.
\newblock In {\em Proceedings of the IEEE International Conference on Computer
  Vision}, pages 993--1000, 2013.

\bibitem{ferstl2015variational}
David Ferstl, Matthias Ruther, and Horst Bischof.
\newblock Variational depth superresolution using example-based edge
  representations.
\newblock In {\em Proceedings of the IEEE International Conference on Computer
  Vision}, pages 513--521, 2015.

\bibitem{gondal2018unreasonable}
Muhammad~Waleed Gondal, Bernhard Sch{\"o}lkopf, and Michael Hirsch.
\newblock The unreasonable effectiveness of texture transfer for single image
  super-resolution.
\newblock In {\em ECCV}, pages 80--97. Springer, 2018.

\bibitem{gu2017learning}
Shuhang Gu, Wangmeng Zuo, Shi Guo, Yunjin Chen, Chongyu Chen, and Lei Zhang.
\newblock Learning dynamic guidance for depth image enhancement.
\newblock In {\em Proceedings of the IEEE Conference on Computer Vision and
  Pattern Recognition}, pages 3769--3778, 2017.

\bibitem{haefner2018fight}
Bjoern Haefner, Yvain Qu{\'e}au, Thomas M{\"o}llenhoff, and Daniel Cremers.
\newblock Fight ill-posedness with ill-posedness: Single-shot variational depth
  super-resolution from shading.
\newblock In {\em Proceedings of the IEEE Conference on Computer Vision and
  Pattern Recognition}, pages 164--174, 2018.

\bibitem{ham2018robust}
Bumsub Ham, Minsu Cho, and Jean Ponce.
\newblock Robust guided image filtering using nonconvex potentials.
\newblock {\em IEEE transactions on pattern analysis and machine intelligence},
  40(1):192--207, 2018.

\bibitem{han2018image}
Wei Han, Shiyu Chang, Ding Liu, Mo Yu, Michael Witbrock, and Thomas~S Huang.
\newblock Image super-resolution via dual-state recurrent networks.
\newblock In {\em Proc. CVPR}, 2018.

\bibitem{handa:etal:ICRA2014}
Ankur Handa, Thomas Whelan, John McDonald, and Andrew~J. Davison.
\newblock A benchmark for {RGB-D} visual odometry, {3D} reconstruction and
  {SLAM}.
\newblock In {\em IEEE Intl. Conf. on Robotics and Automation, ICRA}, Hong
  Kong, China, May 2014.

\bibitem{haris2018deep}
Muhammad Haris, Greg Shakhnarovich, and Norimichi Ukita.
\newblock Deep back-projection networks for super-resolution.
\newblock In {\em Proc. CVPR}, 2018.

\bibitem{honauer2016dataset}
Katrin Honauer, Ole Johannsen, Daniel Kondermann, and Bastian Goldluecke.
\newblock A dataset and evaluation methodology for depth estimation on 4d light
  fields.
\newblock In {\em Asian Conference on Computer Vision}, pages 19--34. Springer,
  2016.

\bibitem{honauer2015hci}
Katrin Honauer, Lena Maier-Hein, and Daniel Kondermann.
\newblock The hci stereo metrics: Geometry-aware performance analysis of stereo
  algorithms.
\newblock In {\em Proceedings of the IEEE International Conference on Computer
  Vision}, pages 2120--2128, 2015.

\bibitem{hua2018normalized}
Jiashen Hua and Xiaojin Gong.
\newblock A normalized convolutional neural network for guided sparse depth
  upsampling.
\newblock In {\em IJCAI}, pages 2283--2290, 2018.

\bibitem{hui2016depth}
Tak-Wai Hui, Chen~Change Loy, and Xiaoou Tang.
\newblock Depth map super-resolution by deep multi-scale guidance.
\newblock In {\em European Conference on Computer Vision}, pages 353--369.
  Springer, 2016.

\bibitem{i2016squeezenet}
Forrest~N. Iandola, Song Han, Matthew~W. Moskewicz, Khalid Ashraf, William~J.
  Dally, and Kurt Keutzer.
\newblock Squeezenet: Alexnet-level accuracy with 50x fewer parameters and
  <0.5mb model size.
\newblock In {\em The IEEE Conference on Computer Vision and Pattern
  Recognition (CVPR)}, 2016.

\bibitem{jiang2018depth}
Zhongyu Jiang, Yonghong Hou, Huanjing Yue, Jingyu Yang, and Chunping Hou.
\newblock Depth super-resolution from rgb-d pairs with transform and spatial
  domain regularization.
\newblock {\em IEEE Transactions on Image Processing}, 27(5):2587--2602, 2018.

\bibitem{johnson2016perceptual}
Justin Johnson, Alexandre Alahi, and Li Fei-Fei.
\newblock Perceptual losses for real-time style transfer and super-resolution.
\newblock In {\em ECCV}, 2016.

\bibitem{kim2018deformable}
Beomjun Kim, Jean Ponce, and Bumsub Ham.
\newblock {Deformable Kernel Networks for Joint Image Filtering}.
\newblock working paper or preprint, Oct. 2018.

\bibitem{krizhevsky2012imagenet}
Alex Krizhevsky, Ilya Sutskever, and Geoffrey~E Hinton.
\newblock Imagenet classification with deep convolutional neural networks.
\newblock In {\em Advances in neural information processing systems}, pages
  1097--1105, 2012.

\bibitem{ledig2016photo}
Christian {Ledig}, Lucas {Theis}, Ferenc {Huszár}, Jose {Caballero}, Andrew
  {Cunningham}, Alejandro {Acosta}, Andrew {Aitken}, Alykhan {Tejani}, Johannes
  {Totz}, Zehan {Wang}, and Wenzhe {Shi}.
\newblock Photo-realistic single image super-resolution using a generative
  adversarial network.
\newblock In {\em Proc. CVPR}, pages 105--114, 2017.

\bibitem{li2018depth}
Beichen Li, Yuan Zhou, Yeda Zhang, and Aihua Wang.
\newblock Depth image super-resolution based on joint sparse coding.
\newblock {\em Pattern Recognition Letters}, 2018.

\bibitem{li2016deep}
Yijun Li, Jia-Bin Huang, Narendra Ahuja, and Ming-Hsuan Yang.
\newblock Deep joint image filtering.
\newblock In {\em European Conference on Computer Vision}, pages 154--169.
  Springer, 2016.

\bibitem{luo2019bigan}
Xiaotong Luo, Rong Chen, Yuan Xie, Yanyun Qu, and Cuihua Li.
\newblock Bi-gans-st for perceptual image super-resolution.
\newblock In Laura Leal-Taix{\'e} and Stefan Roth, editors, {\em ECCV
  Workshops}, pages 20--34, Cham, 2019. Springer International Publishing.

\bibitem{ma2017learning}
Chao Ma, Chih-Yuan Yang, Xiaokang Yang, and Ming-Hsuan Yang.
\newblock Learning a no-reference quality metric for single-image
  super-resolution.
\newblock {\em Computer Vision and Image Understanding}, 158:1 -- 16, 2017.

\bibitem{ma2018self}
Fangchang Ma, Guilherme~Venturelli Cavalheiro, and Sertac Karaman.
\newblock Self-supervised sparse-to-dense: Self-supervised depth completion
  from lidar and monocular camera.
\newblock {\em arXiv preprint arXiv:1807.00275}, 2018.

\bibitem{mal2018sparse}
Fangchang Mal and Sertac Karaman.
\newblock Sparse-to-dense: Depth prediction from sparse depth samples and a
  single image.
\newblock In {\em 2018 IEEE International Conference on Robotics and Automation
  (ICRA)}, pages 1--8. IEEE, 2018.

\bibitem{mantiuk2011hdr}
Rafat Mantiuk, Kil~Joong Kim, Allan~G Rempel, and Wolfgang Heidrich.
\newblock Hdr-vdp-2: a calibrated visual metric for visibility and quality
  predictions in all luminance conditions.
\newblock In {\em ACM Transactions on graphics (TOG)}, volume~30, page~40. ACM,
  2011.

\bibitem{mechrez2018learning}
Roey Mechrez, Itamar Talmi, Firas Shama, and Lihi Zelnik-Manor.
\newblock Learning to maintain natural image statistics.
\newblock {\em arXiv preprint arXiv:1803.04626}, 2018.

\bibitem{paszke2017automatic}
Adam Paszke, Sam Gross, Soumith Chintala, Gregory Chanan, Edward Yang, Zachary
  DeVito, Zeming Lin, Alban Desmaison, Luca Antiga, and Adam Lerer.
\newblock Automatic differentiation in pytorch.
\newblock In {\em NIPS-W}, 2017.

\bibitem{peng2017depth}
Songyou Peng, Bjoern Haefner, Yvain Queau, and Daniel Cremers.
\newblock Depth super-resolution meets uncalibrated photometric stereo.
\newblock In {\em Proceedings of the IEEE International Conference on Computer
  Vision}, pages 2961--2968, 2017.

\bibitem{riegler2016deep}
David Riegler, Gernot aand~Ferstl, Matthias R{\"u}ther, and Horst Bischof.
\newblock A deep primal-dual network for guided depth super-resolution.
\newblock In {\em British Machine Vision Conference}. The British Machine
  Vision Association, 2016.

\bibitem{scharstein2014high}
Daniel Scharstein, Heiko Hirschm{\"u}ller, York Kitajima, Greg Krathwohl, Nera
  Ne{\v{s}}i{\'c}, Xi Wang, and Porter Westling.
\newblock High-resolution stereo datasets with subpixel-accurate ground truth.
\newblock In {\em German Conference on Pattern Recognition}, pages 31--42.
  Springer, 2014.

\bibitem{Simonyan14c}
Karen Simonyan and Andrew Zisserman.
\newblock Very deep convolutional networks for large-scale image recognition.
\newblock {\em CoRR}, abs/1409.1556, 2014.

\bibitem{song2015sun}
Shuran Song, Samuel~P Lichtenberg, and Jianxiong Xiao.
\newblock Sun rgb-d: A rgb-d scene understanding benchmark suite.
\newblock In {\em Proceedings of the IEEE conference on computer vision and
  pattern recognition}, pages 567--576, 2015.

\bibitem{song2016deep}
Xibin Song, Yuchao Dai, and Xueying Qin.
\newblock Deep depth super-resolution: Learning depth super-resolution using
  deep convolutional neural network.
\newblock In {\em Asian Conference on Computer Vision}, pages 360--376.
  Springer, 2016.

\bibitem{song2018deeply}
Xibin Song, Yuchao Dai, and Xueying Qin.
\newblock Deeply supervised depth map super-resolution as novel view synthesis.
\newblock {\em IEEE Transactions on Circuits and Systems for Video Technology},
  2018.

\bibitem{tsuchiya2017depth}
Atsuhiko Tsuchiya, Daisuko Sugimura, and Takayuki Hamamoto.
\newblock Depth upsampling by depth prediction.
\newblock In {\em 2017 IEEE International Conference on Image Processing
  (ICIP)}, pages 1662--1666, Sept 2017.

\bibitem{uhrig2017sparsity}
Jonas Uhrig, Nick Schneider, Lukas Schneider, Uwe Franke, Thomas Brox, and
  Andreas Geiger.
\newblock Sparsity invariant cnns.
\newblock In {\em IEEE International Conference on 3D Vision (3DV)}, 2017.

\bibitem{Ulyanov_2018_CVPR}
Dmitry Ulyanov, Andrea Vedaldi, and Victor Lempitsky.
\newblock Deep image prior.
\newblock In {\em The IEEE Conference on Computer Vision and Pattern
  Recognition (CVPR)}, June 2018.

\bibitem{van2014scikit}
Stefan Van~der Walt, Johannes~L Sch{\"o}nberger, Juan Nunez-Iglesias,
  Fran{\c{c}}ois Boulogne, Joshua~D Warner, Neil Yager, Emmanuelle Gouillart,
  and Tony Yu.
\newblock scikit-image: image processing in python.
\newblock {\em PeerJ}, 2:e453, 2014.

\bibitem{vu2019perception}
Thang Vu, Tung~M. Luu, and Chang~D. Yoo.
\newblock Perception-enhanced image super-resolution via relativistic
  generative adversarial networks.
\newblock In Laura Leal-Taix{\'e} and Stefan Roth, editors, {\em ECCV 2018
  Workshops}, pages 98--113, Cham, 2019. Springer International Publishing.

\bibitem{wang2018recovering}
Xintao Wang, Ke Yu, Chao Dong, and Chen~Change Loy.
\newblock Recovering realistic texture in image super-resolution by deep
  spatial feature transform.
\newblock In {\em Proc. CVPR}, June 2018.

\bibitem{wang2018fully}
Yifan Wang, Federico Perazzi, Brian McWilliams, Alexander Sorkine-Hornung, Olga
  Sorkine-Hornung, and Christopher Schroers.
\newblock A fully progressive approach to single-image super-resolution.
\newblock In {\em CVPR Workshops}, June 2018.

\bibitem{wang2004image}
Zhou Wang, Alan~C Bovik, Hamid~R Sheikh, and Eero~P Simoncelli.
\newblock Image quality assessment: from error visibility to structural
  similarity.
\newblock {\em IEEE transactions on image processing}, 13(4):600--612, 2004.

\bibitem{wang2003multiscale}
Zhou Wang, Eero~P Simoncelli, and Alan~C Bovik.
\newblock Multiscale structural similarity for image quality assessment.
\newblock In {\em The Thrity-Seventh Asilomar Conference on Signals, Systems \&
  Computers, 2003}, volume~2, pages 1398--1402. IEEE, 2003.

\bibitem{xie2016edge}
Jun Xie, Rogerio~Schmidt Feris, and Ming-Ting Sun.
\newblock Edge-guided single depth image super resolution.
\newblock {\em IEEE Transactions on Image Processing}, 25(1):428--438, 2016.

\bibitem{yan2018ddrnet}
Shi Yan, Chenglei Wu, Lizhen Wang, Feng Xu, Liang An, Kaiwen Guo, and Yebin
  Liu.
\newblock Ddrnet: Depth map denoising and refinement for consumer depth cameras
  using cascaded cnns.
\newblock In {\em Proceedings of the European Conference on Computer Vision
  (ECCV)}, pages 151--167, 2018.

\bibitem{yang2014color}
Jingyu Yang, Xinchen Ye, Kun Li, Chunping Hou, and Yao Wang.
\newblock Color-guided depth recovery from rgb-d data using an adaptive
  autoregressive model.
\newblock {\em IEEE transactions on image processing}, 23(8):3443--3458, 2014.

\bibitem{zhang2011fsim}
Lin Zhang, Lei Zhang, Xuanqin Mou, and David Zhang.
\newblock Fsim: a feature similarity index for image quality assessment.
\newblock {\em IEEE transactions on Image Processing}, 20(8):2378--2386, 2011.

\bibitem{Zhang_2018_CVPR}
Richard Zhang, Phillip Isola, Alexei~A. Efros, Eli Shechtman, and Oliver Wang.
\newblock The unreasonable effectiveness of deep features as a perceptual
  metric.
\newblock In {\em The IEEE Conference on Computer Vision and Pattern
  Recognition (CVPR)}, June 2018.

\bibitem{zhang2018image}
Yulun Zhang, Kunpeng Li, Kai Li, Lichen Wang, Bineng Zhong, and Yun Fu.
\newblock Image super-resolution using very deep residual channel attention
  networks.
\newblock In {\em Proceedings of the European Conference on Computer Vision
  (ECCV)}, pages 286--301, 2018.

\bibitem{zhang2018residual}
Yulun Zhang, Yapeng Tian, Yu Kong, Bineng Zhong, and Yun Fu.
\newblock Residual dense network for image super-resolution.
\newblock In {\em Proc. CVPR}, June 2018.

\bibitem{zhao2017loss}
Hang {Zhao}, Orazio {Gallo}, Iuri {Frosio}, and Jan {Kautz}.
\newblock Loss functions for image restoration with neural networks.
\newblock {\em IEEE Transactions on Computational Imaging}, 3(1):47--57, 2017.

\bibitem{zhao2017simultaneously}
Lijun Zhao, Huihui Bai, Jie Liang, Bing Zeng, Anhong Wang, and Yao Zhao.
\newblock Simultaneously color-depth super-resolution with conditional
  generative adversarial network.
\newblock {\em arXiv preprint arXiv:1708.09105}, 2017.

\bibitem{zuo2018minimum}
Yifan Zuo, Qiang Wu, Jian Zhang, and Ping An.
\newblock Minimum spanning forest with embedded edge inconsistency measurement
  model for guided depth map enhancement.
\newblock {\em IEEE Transactions on Image Processing}, 27(8):4145--4159, 2018.

\end{thebibliography}
\end{document}